%% file: main.tex
\documentclass[11pt]{article}

\usepackage{geometry}  % set the margins to 1in on all sides
\usepackage[english]{babel}
\usepackage[utf8]{inputenc}
\usepackage[T1]{fontenc}

\usepackage{color}

\usepackage[dvipsnames]{xcolor}
\usepackage{paralist}
\usepackage{graphicx}
\usepackage{subcaption}
\usepackage{longtable} 
\usepackage{multirow}
\usepackage{listings}
\usepackage{makecell}
\usepackage{array}
\usepackage{float}
\usepackage{dsfont}
\usepackage{rotating}
\usepackage{booktabs}
\usepackage{enumerate}
\usepackage{tikz}
\usepackage{pgf}
\usepackage{enumitem}
\usepackage{mathtools}
\newfloat{algorithm}{t}{lop}
\mathtoolsset{showonlyrefs}
\usepackage{scalerel}
\usepackage{authblk}
\usepackage{tabulary}
\usepackage{booktabs}
\usepackage{wrapfig}

\usepackage{amsmath,amssymb,amsfonts,amsthm,bbm}
\usepackage{mathtools}
\usepackage{mathrsfs}

\usepackage{algorithm}
\usepackage{algpseudocode}
\usepackage{algorithmicx}

\usepackage{tcolorbox}

\usetikzlibrary{arrows.meta,positioning,shapes.geometric}

\definecolor{uiGray}{HTML}{9AA0A6}
\definecolor{uiRed}{HTML}{E74C3C}
\definecolor{uiBlue}{HTML}{1F77B4}
\definecolor{uiBlack}{HTML}{111111}

\usepackage{natbib}
\usepackage{authblk}
\usepackage{todonotes}
\usepackage{xr-hyper}
\usepackage[
  colorlinks,
  citecolor=blue,
  linkcolor=red,
  anchorcolor=red,
  urlcolor=blue
]{hyperref}

\usepackage{color}
\usepackage{xcolor}
\definecolor{cm}{RGB}{50, 168, 115}
\definecolor{zm}{RGB}{200,125,40}

\definecolor{yg}{RGB}{0,0,0}
\newcommand{\yg}[1]{\textcolor{yg}{#1}}

\input{macro}

\title{IndiSeek learns information-guided disentangled representations}
\date{\today}
\author[1]{Yu Gui}
\affil[1]{Department of Statistics and Data Science, University of Pennsylvania}
\author[2]{Cong Ma}
\affil[2]{Department of Statistics, University of Chicago}
\author[3]{Zongming Ma}
\affil[3]{Department of Statistics and Data Science, Yale University}

\begin{document}

\maketitle

\begin{abstract}
\input{abs}

\end{abstract}

\input{paper}

\section*{Acknowledgements}
C.M.~was partially supported by the National Science Foundation via grant DMS-2311127 and the CAREER
Award DMS-2443867.
Z.M.~was
partially supported by the National Institutes of Health via grant U01CA294514 and the National Science Foundation via grants DMS-2345215 and DMS-2245575.

\bibliography{main}
\bibliographystyle{apalike}

\newpage
\appendix

\input{appendix}

\end{document}

%% file: macro.tex
\usepackage{mdframed}
\usepackage{kantlipsum}

\mdfdefinestyle{myenvs}{%
  hidealllines=true,%
  nobreak=true, % comment this to allow breaking
  leftmargin=0pt,
  rightmargin=0pt,
  innerleftmargin=0pt,
  innerrightmargin=0pt,
}

\theoremstyle{plain}
\newmdtheoremenv[style=myenvs]{theorem}{Theorem}
\newmdtheoremenv[style=myenvs]{proposition}{Proposition}
\newmdtheoremenv[style=myenvs]{lemma}{Lemma}
\newmdtheoremenv[style=myenvs]{example}{Example}
\newmdtheoremenv[style=myenvs]{corollary}{Corollary}
\theoremstyle{definition}
\newmdtheoremenv[style=myenvs]{definition}{Definition}
\newmdtheoremenv[style=myenvs]{assumption}{Assumption}

\theoremstyle{remark}

\usepackage{xspace}

\def\##1\#{\begin{align}#1\end{align}}
\def\$#1\${\begin{align*}#1\end{align*}}

\definecolor{myblue}{rgb}{.8, .8, 1}
\definecolor{mathblue}{rgb}{0.2472, 0.24, 0.6} % mathematica's Color[1, 1--3]
\definecolor{mathred}{rgb}{0.6, 0.24, 0.442893}
\definecolor{mathyellow}{rgb}{0.6, 0.547014, 0.24}

\newcommand{\calL}{{\mathcal{L}}}

\newcommand{\calN}{{\mathcal{N}}}

\newcommand{\bfm}[1]{\ensuremath{\mathbf{#1}}}

  \def\EE{\mathbb{E}}

 \def\bI{\bfm I}

 \def\bO{\bfm O}

 \def\bR{\bfm R} \def\RR{\mathbb{R}}

\newcommand{\bfsym}[1]{\ensuremath{\boldsymbol{#1}}}

\def\bzeta{\bfsym {\zeta}}

\def\hat{\widehat}
\def\tilde{\widetilde}

\DeclareTextFontCommand{\texttt}{\ttfamily\upshape}

\def\ours{{IndiSeek}}
\def\disen{{InfoDisen}}

%% file: abs.tex
Learning disentangled representations is a fundamental task in multi-modal learning.
In modern applications such as single-cell multi-omics, both shared and modality-specific features are critical for characterizing cell states and supporting downstream analyses.
Ideally, modality-specific features should be independent of shared ones while also capturing all complementary information within each modality.
This tradeoff is naturally expressed through information-theoretic criteria, but mutual-information-based objectives are difficult to estimate reliably, and their variational surrogates often underperform in practice. 
In this paper, we introduce \ours, a novel disentangled representation learning approach that addresses this challenge by combining an independence-enforcing objective with a computationally efficient reconstruction loss that bounds conditional mutual information. This formulation explicitly balances independence and completeness, enabling principled extraction of modality-specific features. 
We demonstrate the effectiveness of \ours~on synthetic simulations, a CITE-seq dataset and multiple real-world multi-modal benchmarks.

%% file: paper.tex
\section{Introduction}

The growing availability of multi-modal data has broadened the scope of representation learning. 
One notable example is the development of single-cell multi-omics technologies in genomics \citep{teichmann2020method}. 
% As an example, 
For instance,
CITE-seq
(Cellular Indexing of Transcriptomes and Epitopes by Sequencing) \citep{stoeckius2017simultaneous} enables simultaneous measurement of gene expression via single-cell RNA sequencing and protein abundance via antibody-derived tags (ADTs). 
{Integrative analysis of both modalities at the single-cell resolution has improved tasks such as cell state delineation and enabled
% large-scale efforts in 
cell atlas construction and querying at finer granularity.} 
Multi-modal data is also abundant at the intersection of computer vision, natural language processing, and audio processing, where combining information from different sources for representation learning forms the backbone of many groundbreaking progresses 
\citep{radford2021learning, baltruvsaitis2018multimodal, jia2021scaling, akbari2021vatt, liang2024foundations}.

The central question of multi-modal learning is how to extract informative and interpretable representations from complex multi-modal data.  
This involves two complementary goals: 
% that are both crucial: 
\begin{enumerate}
    \item[(1)] Extracting interdependence (or \emph{shared} information) across modalities;
    \item[(2)] Extracting \emph{modality-specific} information unique to each modality.
\end{enumerate}
Shared features enhance predictive power across modalities by capturing cross-modal dependence. From an information-theoretic perspective, they should contain all information common across modalities.
At the same time, preserving modality-specific diversity is equally important.
In single-cell multi-omics, modality-specific features often correspond to unique biological signals that aid cell type identification and the discovery of novel or rare cell populations~\citep{hao2021integrated,caron2025multimodal}.
After all, if all the information in a certain modality can be inferred from other modalities, there is little reason to measure it in the first place. 
Finally, disentangling shared and modality-specific components improves interpretability: shared features summarize cross-modal information, while modality-specific features capture complementary signals. 
% Across these domains, 
% {From field to field}, these same two principles—capturing shared information and preserving modality-specific signals—are central to building effective representation learning frameworks.

Methodologically, shared information is usually extracted by maximizing cross-modal mutual information~\citep{tosh2021contrastive,sridharan2008information}, with contrastive methods like CLIP~\citep{radford2021learning} widely adopted in practice. 
Under certain conditions, 
CLIP-learned representations have been shown to be not only sufficient (i.e., capturing all shared information), but also minimally sufficient (i.e., containing minimal extra information) \citep{gui2025multi, oko2025statistical,lin2025statistical} under certain conditions. 
% However, the notion of shared and modality-specific features can be ambiguous without rigorous definitions. 

In contrast, identifying modality-specific features, also known as disentangled representation learning, has become an active research area~ \citep{fischer2020conditional,liang2023factorized,liu2023focal,wang2024information,dufumier2024align,wang2024decoupling}.
% \todo{I edited this paragraph. Please check.}
% These features carry information that cannot be recovered from cross-modality correlations and therefore provide complementary insights. In biological applications, for example, RNA-specific patterns may reflect transcriptional activity that is not visible at the protein level, while protein-specific markers can highlight post-translational regulation. More generally, modality-specific features are valuable for distinguishing heterogeneous cell populations, and overlooking them risks losing signals that may be both biologically and statistically important.
% Nevertheless, separating \emph{shared} and \emph{modality-specific} features is not straightforward. 
{
While shared features should be minimally sufficient, modality-specific features must both remain \emph{independent} of (i.e., \emph{disentangled} from) shared features and capture all \emph{complementary} information within each modality~\citep{liang2023factorized, wang2024information, wang2024decoupling}. 
From an information-theoretic perspective, this requires (1) simultaneously minimizing the mutual information between shared and modality-specific features and (2) maximizing the mutual information between data and their union.
Multiple state-of-the-art methods \citep{liang2023factorized,wang2024information} tackle this problem from such a perspective, but differ in how they approximate the underlying mutual information. 
Since such quantities are difficult to estimate with finite samples, these methods often trade off independence against completeness.
As we show in the motivating examples below, this tradeoff could interfere with
their ability to achieve both goals simultaneously.} 

% different SOTA methods vary in their approaches to approximating mutual information quantities.
% However, as our motivating examples below suggest, these SOTA methods have their limitations in simultaneously achieving the goals of independence and complementary information capture.

% \paragraph{Paper organization.} 
% \yg{Section~\ref{sec:problem-setup} formally defines the setting studied in this paper and introduces the information-theoretic objectives in learning both shared and modality-specific features and their variants in the literature. In Section~\ref{sec:simu}, we present simulated examples to demonstrate the challenge faced by existing methods. Following the example, we introduce our method \ours~ in Section~\ref{sec:method}, and its performance is evaluated on both simulated datasets in Section~\ref{sec:simu} and a real-world CITE-seq dataset in Section~\ref{sec:citeseq} as well as MultiBench datasets in Section~\ref{sec:multibench}.}

\subsection{Motivating examples and limitations of SOTA}\label{sec:simu}
To demonstrate the limitation of the state-of-the-art (SOTA) methods in disentangled learning, we consider two simulated settings. 
To isolate the challenge of modality-specific feature extraction, we assume the shared features  $C_1 = f_1(X_1)\in \RR^{d_c}$ are provided by an oracle.
The task is then to learn modality-specific features from a single modality $X_1 \in \mathbb{R}^{d_1}$, with $d_1 =6$ and $d_c = 2$.

% Thus, the goal reduces to learning modality-specific features within a single modality.
% In both settings, let $d_1 = 6$ and $d_c=2$.
% The two settings differ in the structures of the shared features.
\begin{itemize}%[itemsep=0pt, topsep=0pt, leftmargin=*]
    \item Setting 1: Let the observed data be iid copies of $X_1 {\sim} \calN(0, I_{d_1})$. For each $x\in \RR^{d_1}$, define the shared representation map
    \begin{align}
        f_1(x) = 0.5A_f x + 0.2 \sin(A_f x) + 0.2(A_f x)^3 \quad \text{with} \quad A_f = (\bI_{d_c}, \bO) \in \RR^{d_c \times d_1}.
    \end{align}
    Here, the sine and cubic functions are applied entrywise. 
    The ideal modality-specific features are the last four coordinates of $X_1$, which contain all remaining information while being independent of $C_1$.
    
    % Intuitively, the ideal set of modality-specific features that contain all remaining information about $X_1$ while being independent of $C_1$ should be a bijection of $(X_1)_{(d_c+1):d_1} \sim \calN(0,\bI_{d_1-d_c})$.  
    
    \item Setting 2: 
    Let the observed data be iid copies of $X_1$ where $(X_1)_{\{1,2,5,6\}} \sim \calN(0, \bI_4)$, $(X_1)_3 = 0.2\times((X_1)_1 + (X_1)_2)$, and $(X_1)_4 = (X_1)_1 \times (X_1)_2$. 
    Thus, the third and fourth coordinates are deterministic functions of the first two, while the others are independent.
    Let $C_1 = (X_1)_{1:d_c}$. 
    Here, the ideal modality-specific features are the last two coordinates. 
\end{itemize}

For both settings, given $C_1$, our goal is to learn a map $h_1:\RR^{d_1}\to \RR^{p_1}$ with $p_1 =10$ such that $h_1(X_1)$ gives the modality-specific features. 
All neural networks are five-layer ReLU MLPs with width $100$, trained on $10000$ samples.
Ablation studies beyond Gaussian distributions are presented in the appendix.

For any learned map $h_1$, we quantify the importance of each coordinate of $X_1$ in determining its output via gradients based on feature masking\footnote{See Appendix~\ref{sec:app-metric} for details on this feature importance metric.} on an independent test set of size $1000$, following standard model-free importance measures~\citep{robnik2008explaining,zeiler2014visualizing,li2016understanding}
% sampled from the same data-generating process. 

\begin{figure}[t]
    \centering
    \begin{subfigure}[t]{0.3\linewidth}
        \centering
        \includegraphics[width=\linewidth]{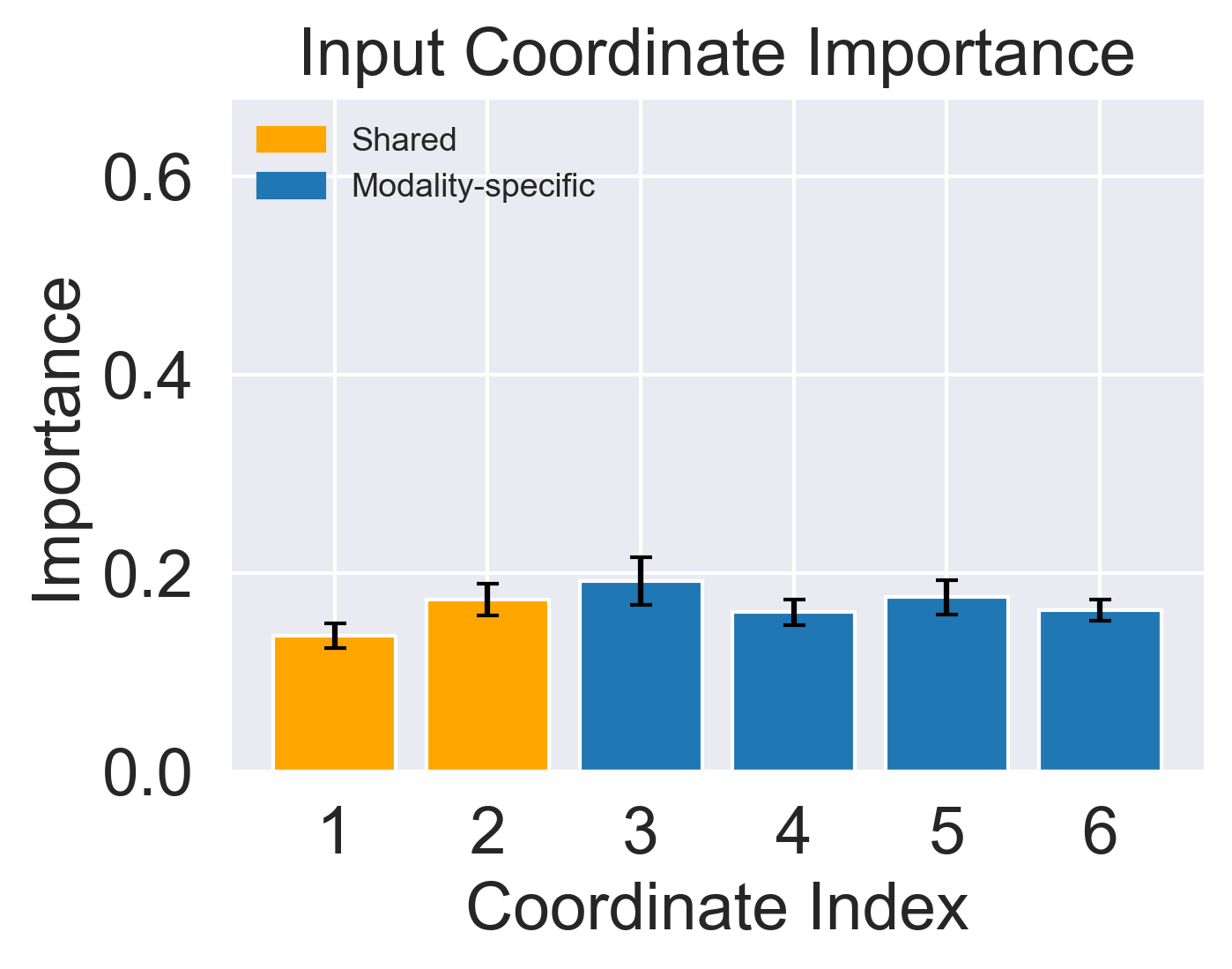}
        \caption{Factorized CL.}
        \label{fig:factcl}
    \end{subfigure}
    \hfill
    \begin{subfigure}[t]{0.3\linewidth}
        \centering
        \includegraphics[width=\linewidth]{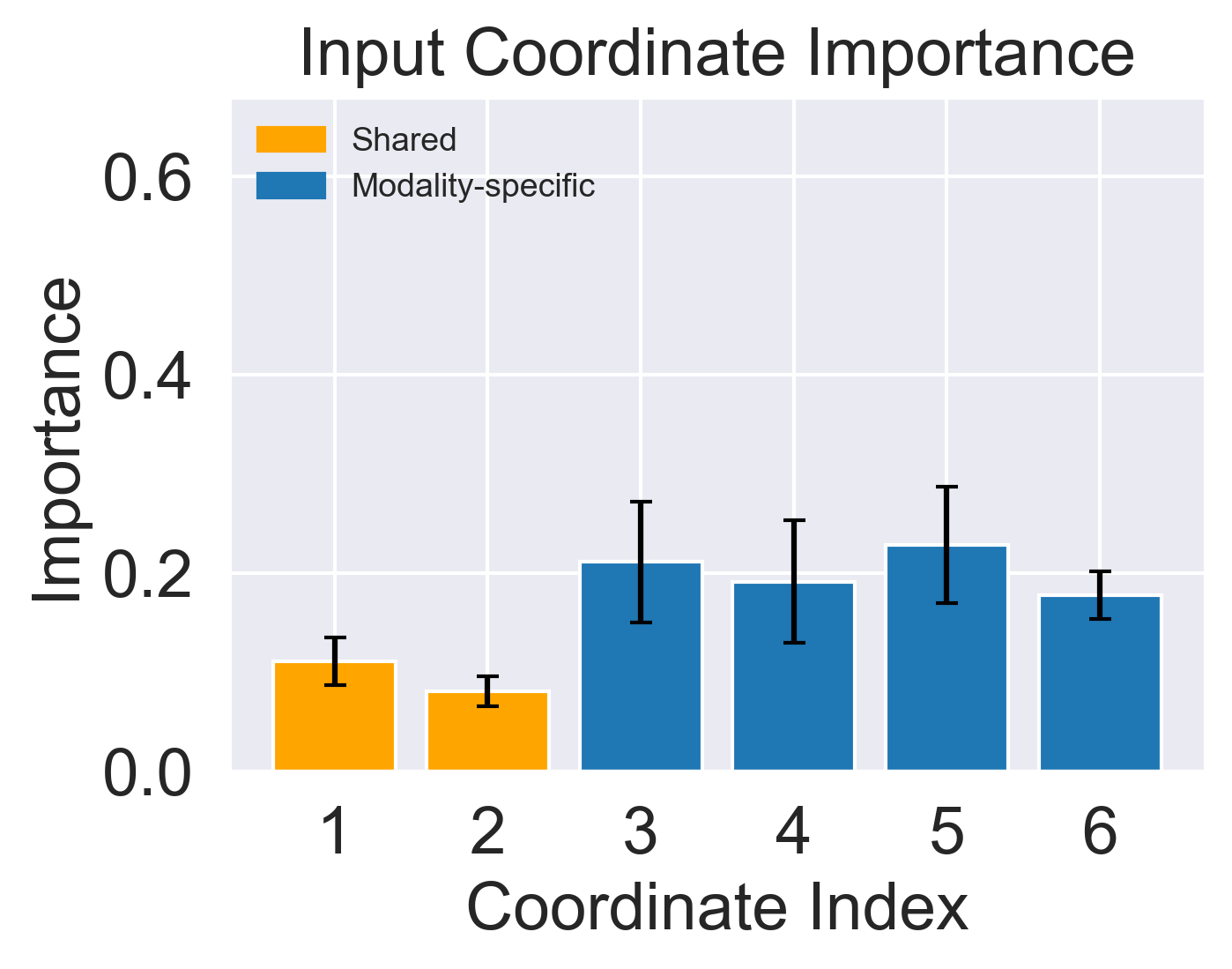}
        \caption{\disen.}
        \label{fig:disenssl}
    \end{subfigure}
    \hfill
    \begin{subfigure}[t]{0.3\linewidth}
        \centering
        \includegraphics[width=\linewidth]{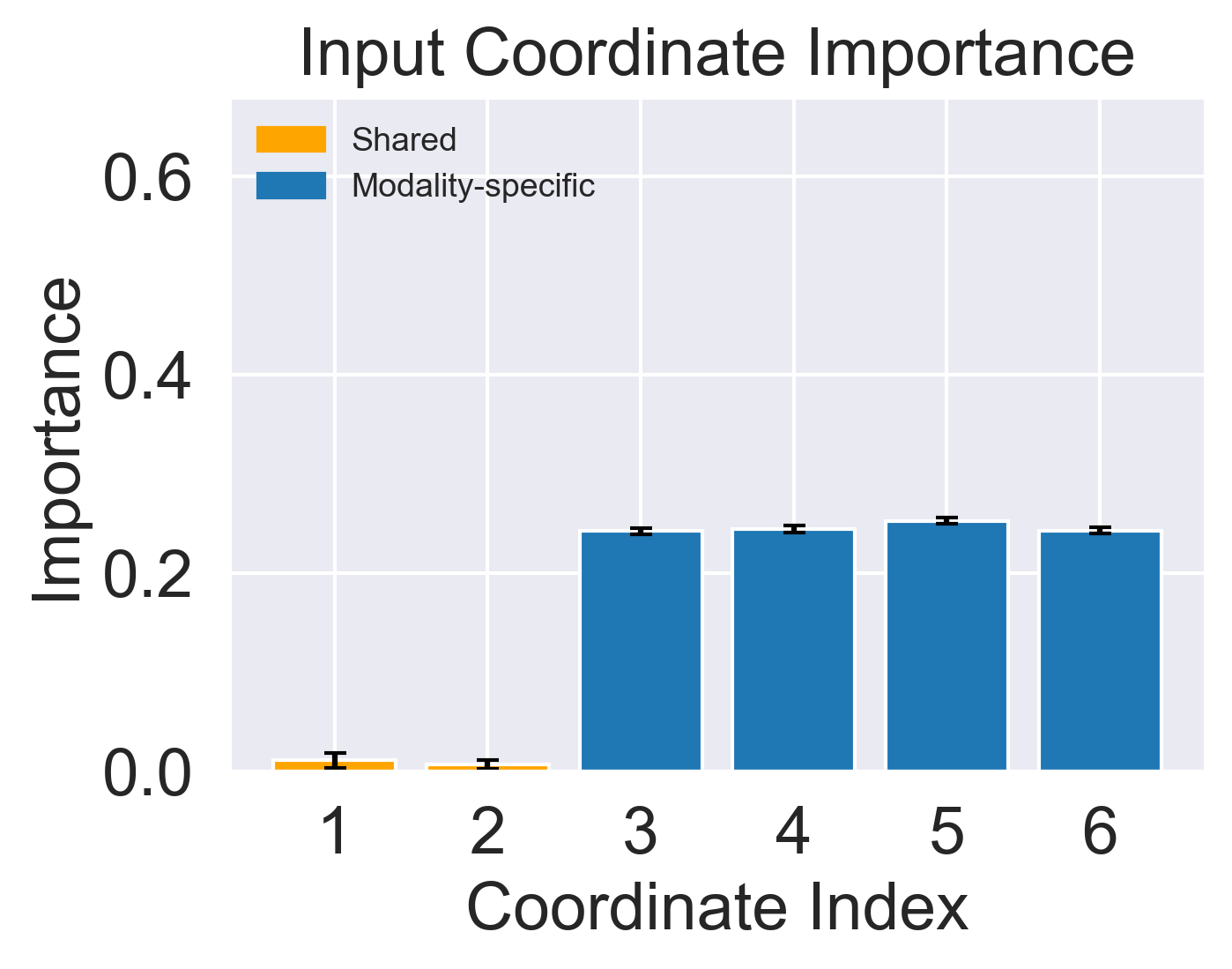}
        \caption{Our method (\ours).}
        \label{fig:recons}
    \end{subfigure}
    \caption{\yg{Importance of learned modality-specific features: Setting 1.}}
    \label{fig:setting1}
\end{figure}

\begin{figure}[t]
    \centering
    \begin{subfigure}[t]{0.3\linewidth}
        \centering
        \includegraphics[width=\linewidth]{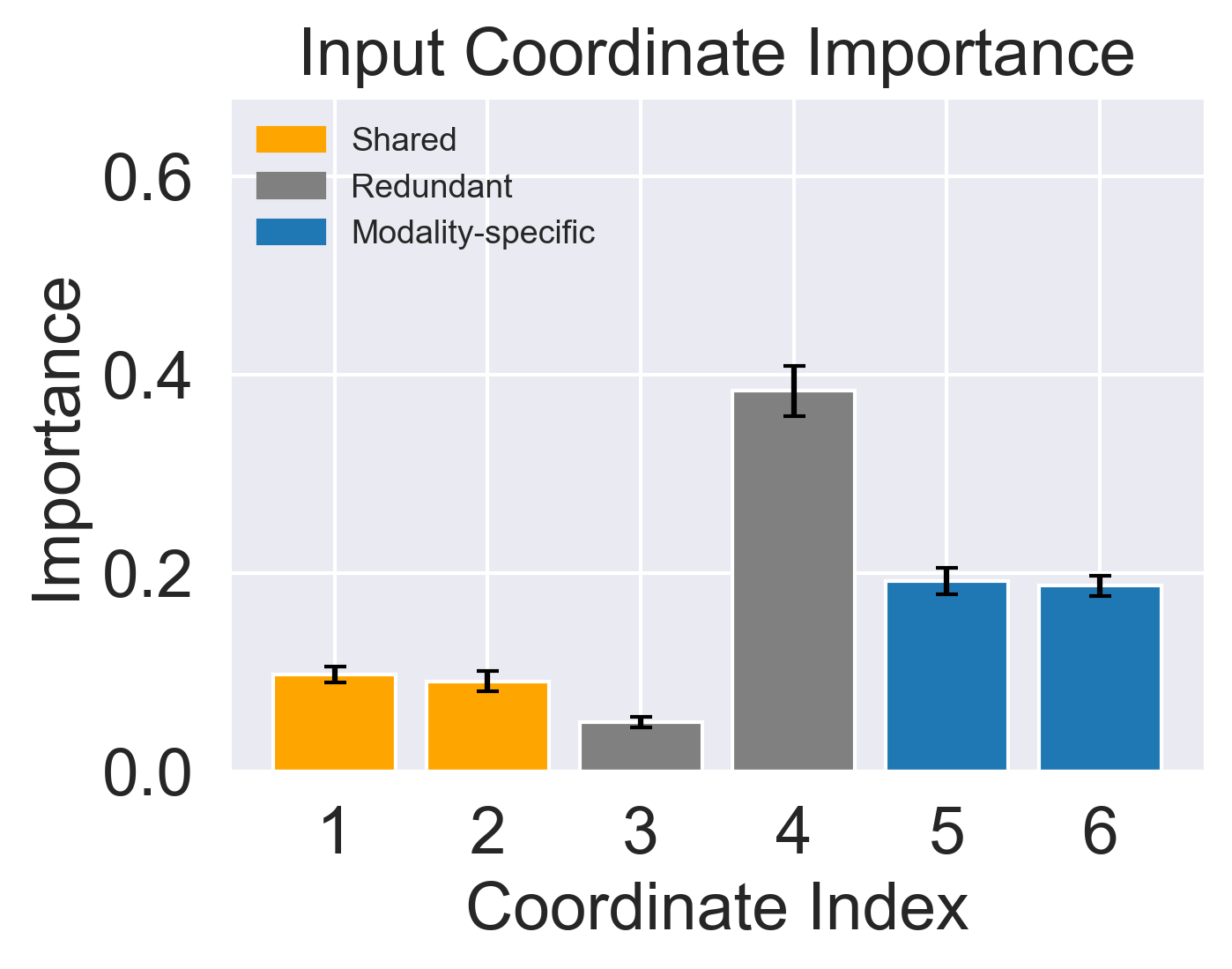}
        \caption{Factorized CL.}
        \label{fig:factcl1}
    \end{subfigure}
    \hfill
    \begin{subfigure}[t]{0.3\linewidth}
        \centering
        \includegraphics[width=\linewidth]{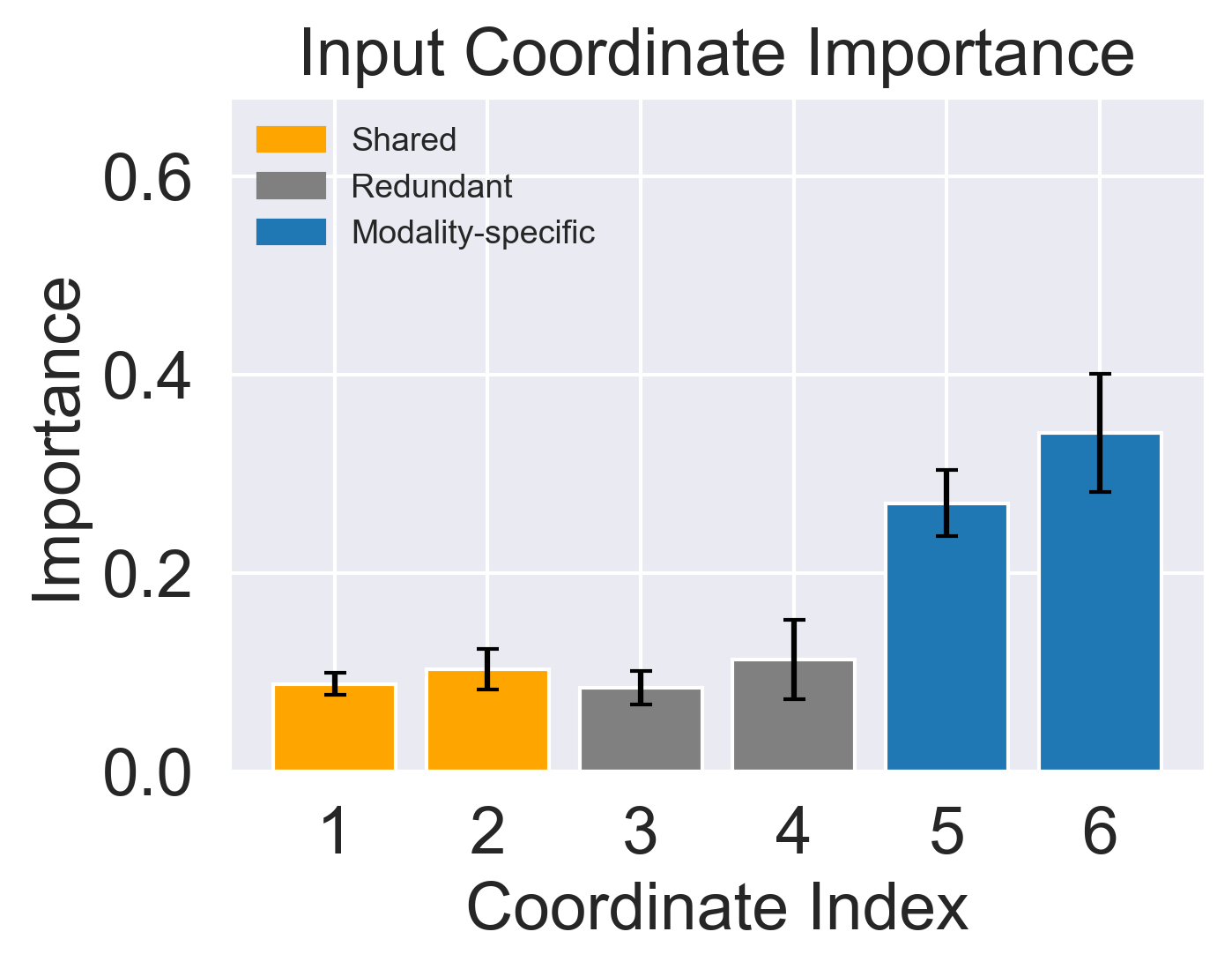}
        \caption{\disen.}
        \label{fig:disenssl1}
    \end{subfigure}
    \hfill
    \begin{subfigure}[t]{0.3\linewidth}
        \centering
        \includegraphics[width=\linewidth]{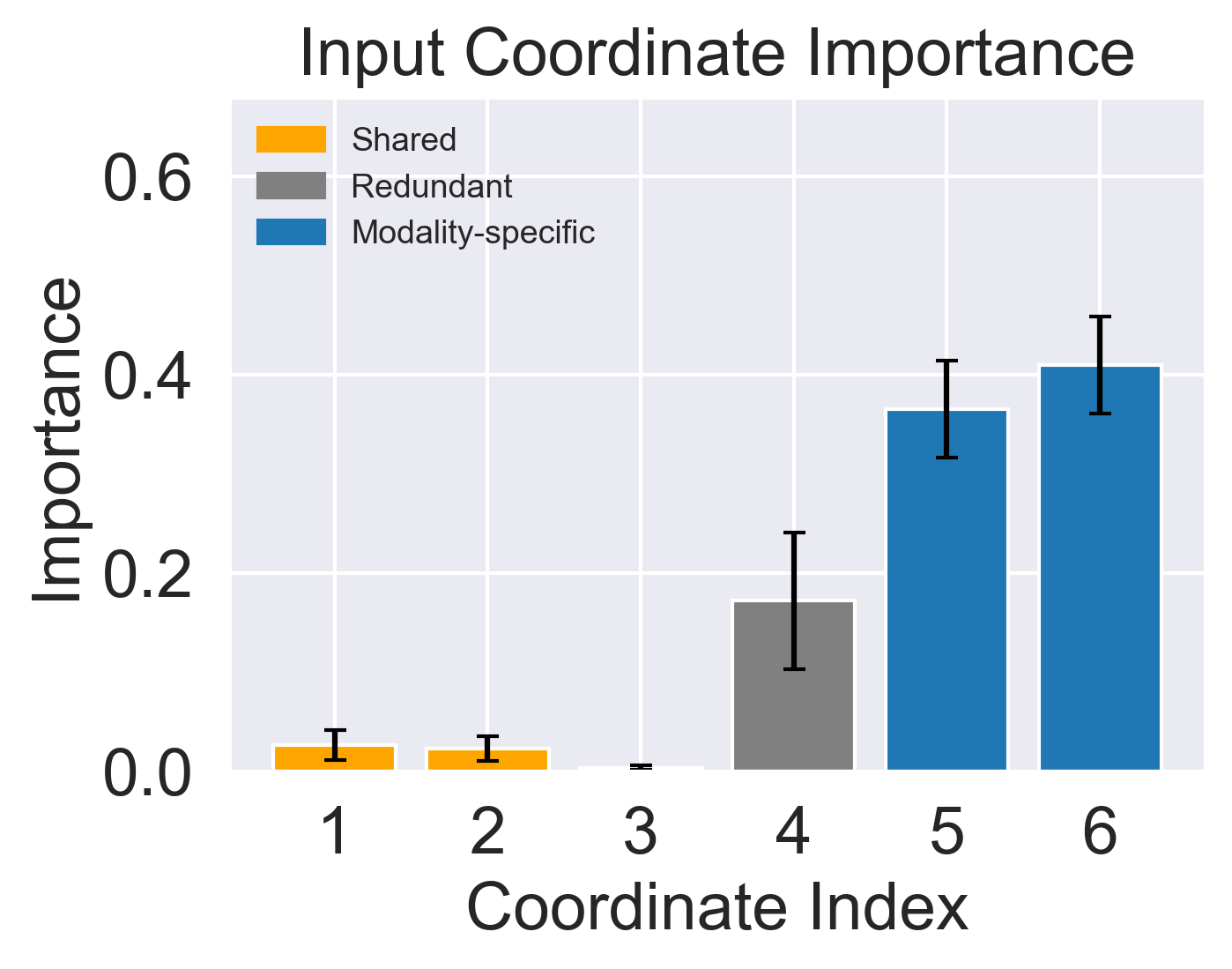}
        \caption{Our method (\ours).}
        \label{fig:recons1}
    \end{subfigure}
    \caption{\yg{Importance of learned modality-specific features: Setting 2.}}
    \label{fig:setting4}
\end{figure}

We present the average coordinate importance metrics in learned modality-specific representation maps over $50$ simulation runs in Figures~\ref{fig:setting1} and \ref{fig:setting4} for three different learning algorithms in the two settings, respectively.
In both figures, the left panels report results of the SOTA approach FactorizedCL in \cite{liang2023factorized} (without self-supervision), the middle panels report results of the SOTA approach {\disen} in \cite{wang2024information}, and the right panels report results of the new approach {\ours} we propose in this paper. 
\yg{Tuning parameters\footnote{See Appendix~\ref{sec:app-simu} for an ablation study on performances with different tuning parameter values.} are set at $0.1$ for {\disen} and {\ours} and at $1.0$ by default for FactorizedCL.} 
% \nb{why 0.01 for IndiSeek, this is different from our default value later}

As we have reasoned when introducing the settings, the ideal modality-specific representations in Setting 1 should depend on and only on the last four coordinates, while in Setting 2, on and only the last two coordinates. 
The left and the middle panels of both figures suggest that neither FactorizedCL nor {\disen} learn an ideal representation in either setting, as irrelevant coordinates play non-trivial roles in determining the learned modality-specific representations. 
Therefore, both SOTA approaches have their limitations in achieving the desired goal of disentangled representation learning.
In contrast, the right panel of Figure \ref{fig:setting1} shows that the new method \ours~we are to propose achieves the ideal property: the learned modality-specific features depend on and only on all coordinates that are independent of those involved in shared features. \yg{In Figure~\ref{fig:setting4}, compared with SOTA methods, modality-specific representations learned by \ours~exhibit much less dependence on both shared and redundant features, which demonstrates its effectiveness in this setting with highly nonlinear dependence and nontrivial redundancy.}

\subsection{Paper organization}
{In view of the foregoing limitations of SOTA approaches, in this paper, we introduce a new computationally efficient method \ours, for extracting both shared and modality-specific features, and with a focus on the latter. 
Based on a reconstruction loss serving as an upper bound for mutual information, \ours~is capable of disentangling shared and modality-specific features, while avoiding information loss in each modality at the same time. 
The effectiveness of {\ours} is demonstrated on both simulated and real-world datasets, including a single-cell multi-omics dataset CITE-seq, as well as multi-modal benchmark datasets.}   

The rest of this paper is organized as follows. 
Section \ref{sec:problem-setup} reviews the properties an ideal ``shared + modality-specific'' feature representation of multi-modal data should have.
Section \ref{sec:method} introduces {\ours} in detail as a practical implementation for seeking such an ideal representation and compares it with SOTA methods in terms of different choices made when approximating information-theoretic quantities. Experiments on a CITE-seq dataset and on MultiBench datasets are presented in Sections \ref{sec:citeseq} and \ref{sec:multibench}, respectively. 
Section \ref{sec:discuss} discusses other related works and potential extensions of the proposed method.
Technical details and additional numerical studies are deferred to Appendices.

\section{Ideal decomposition of multi-modal data}\label{sec:problem-setup}

We consider a \emph{fully unsupervised} setting with observations from two modalities $X_1 \in \RR^{d_1}$ and $X_2 \in \RR^{d_2}$. 
Our goal is to decompose each modality into a shared information component that captures cross-modal information and a modality-specific information component that captures unique variation:
\begin{align}\label{eq:modality-decomposition}
    X_1 \;\mapsto\; (\underbrace{C_1}_{\text{shared}}, \underbrace{Z_1}_{\text{specific}}), \qquad X_2 \;\mapsto\; (\underbrace{C_2}_{\text{shared}}, \underbrace{Z_2}_{\text{specific}}).
\end{align}
Here $C_1$ and $C_2$ are $d_c$-dimensional\footnote{In theory, the dimensions of $C_1$ and $C_2$ need not match. In practice, these shared latent factors are often obtained from mapping both modalities into some co-embedding space. Therefore, without loss of generality, we assume that $C_1$ and $C_2$ have the same dimension.} shared information components (though not necessarily identical) while $Z_1 \in \RR^{p_1}$ and $Z_2 \in \RR^{p_2}$ are modality-specific information components.

We formalize the ideal representation~/~decomposition by requiring the following desiderata: 

\begin{enumerate}[itemsep=0pt, topsep=0pt, leftmargin=*]
    \item \textbf{Minimal sufficiency:} the shared features preserve all information common to both modalities,~i.e., 
    $$I(X_1; X_2) = I(C_1; C_2).$$
    Intuitively, $C_1$ and $C_2$ should be sufficient for cross-modal prediction tasks. In addition, the shared features should not encode redundant details,
    \[I(C_1;X_1) = \min\{I(f(X_1);X_1): I(X_1;X_2) = I(f(X_1);X_2)\},
    \]
    and an analogous identity holds for $C_2$.
    This prevents ``over-capturing'' modality-specific signals inside the shared space.

\item \textbf{Independence:} modality-specific and shared features should be disentangled\footnote{\yg{As $C_1$ is restricted to be a function of $X_1$, it is possible that $C_1$ also contains modality-specific information. Meanwhile, all information in $C_2$ about $X_1$ is strictly shared. Therefore, we require $I(Z_1;C_2) = I(Z_2;C_1) = 0$ instead of $I(Z_1;C_1) = I(Z_2;C_2) = 0$.}},
    \yg{
    \[
    % I(Z_1; C_1) = 
    I(Z_1;C_2) = I(Z_2;C_1) 
    % = I(Z_2; C_2) 
    = 0.
    \]
    }

    \item \textbf{Complementary information capture:} the pair $(C_i, Z_i)$ should retain enough information to recover the $i$th modality:
    \[
    I(C_i, Z_i; X_i) \ \text{is maximized for } i=1,2.
    \]
    This ensures that information not explained by shared features is fully captured by the modality-specific components. 
\end{enumerate}
Together, these criteria define what it means for a decomposition to be both 
\emph{sufficient} and \emph{disentangled}.  

In practice, the extraction of shared features is often carried out with contrastive learning methods such as CLIP, 
which we also adopt in our framework. 
Therefore, in the remainder of this work, we focus primarily on the more challenging problem of learning 
\emph{modality-specific} representations, assuming that the shared representations have already been obtained.

\section{Information-guided disentangled representation seeking (\ours)}\label{sec:method}

Given learned shared features,
the desiderata in Section~\ref{sec:problem-setup} motivate the search for disentangled modality-specific features, 
but leave open the question of \emph{how to realize them}. 
Prior work has proposed an information-theoretic objective:\yg{
\begin{align}\label{eq:obj-specific}
\max_{Z_i}\; I(Z_i, C_j; X_i) 
\qquad \text{subject to} \quad I(Z_i; C_{j'}) \approx 0,
\end{align}
where $j, j' \in \{1,2\}$ have respective specifications in each method, in which cross-modal switching of shared features is adopted when $j, j' \neq i$.}
% \nb{this is now a problem as we now regard setting $I(Z_i; C_i) \approx 0$ as the wrong practice. is FactorizedCL also doing this? need to discuss more carefully here}
This objective reflects the prevailing philosophy: modality-specific features should be maximally 
informative about their own modality, conditional on the shared component, while being independent of it.

\begin{figure}[t]
\centering
\begin{tikzpicture}[
  scale=0.7,
  transform shape,
  >=Latex,
  font=\sffamily,
  node distance=20mm and 28mm,
  enc/.style={draw, very thick, rounded corners=2pt, minimum width=12mm, minimum height=9mm},
  xnode/.style={enc},
  xhat/.style={enc, draw=gray!70},
  cnode/.style={draw=red!80!black, very thick, ellipse, minimum width=16mm, minimum height=12mm},
  znode/.style={draw=blue!80!black, very thick, ellipse, minimum width=16mm, minimum height=12mm},
  bus/.style={draw=gray!70, very thick, rounded corners=6pt},
  lab/.style={inner sep=1pt, font=\large}
]
% ================== Left branch: modality 1 ==================
\node[xnode] (x) at (0,0) {Input $\mathbf{X}_1$};
\node[xhat] (xhat) at (0,2) {Reconstructed $\tilde{\mathbf{X}}_1$};
\node[cnode] (c) at (4,0) {shared $\mathbf{C}_1$};
\node[znode] (z) at (4,-2) {specific $\mathbf{Z}_1$};
% Encoders
\draw[very thick, ->] (x.east) -- node[lab, above, pos=0.5] {$f_1$} (c.west);
\draw[very thick, ->] (x.south) -- (0,-2) -- node[lab, above, pos=0.6] {$h_1$} (z.west);
% Reconstruction bus
\coordinate (r0) at (6,0);     
\coordinate (r1) at (6,-2);    
\coordinate (rtop) at (6,2);   
\draw[bus] (c.east) -- (r0);
\draw[bus] (z.east) -- (r1);
\draw[bus] (r0) -- (r1);       
\draw[bus, ->] (r0) -- (rtop) -- (2.0,2) -- (xhat.east)
  node[lab, above, pos=-2] {$g_1$};
% skip connections
\draw[very thick, draw=gray!60, dashed, <->] (x.north) -- (xhat.south);
% REMOVED: \draw[very thick, draw=blue!70, dashed] (c.south) -- (z.north);

% ================== Right branch: modality 2 ==================
\node[xnode] (x2) at (14,0) {Input $\mathbf{X}_2$};
\node[xhat] (xhat2) at (14,2) {Reconstructed $\tilde{\mathbf{X}}_2$};
\node[cnode] (c2) at (10,0) {shared $\mathbf{C}_2$};
\node[znode] (z2) at (10,-2) {specific $\mathbf{Z}_2$};
% Encoders
\draw[very thick, ->] (x2.west) -- node[lab, above, pos=0.5] {$f_2$} (c2.east);
\draw[very thick, ->] (x2.south) -- (14,-2) -- node[lab, above, pos=0.6] {$h_2$} (z2.east);
% Reconstruction bus
\coordinate (r0b) at (8,0);     
\coordinate (r1b) at (8,-2);    
\coordinate (rtopb) at (8,2);   
\draw[bus] (c2.west) -- (r0b);
\draw[bus] (z2.west) -- (r1b);
\draw[bus] (r0b) -- (r1b);      
\draw[bus, ->] (r0b) -- (rtopb) -- (12.0,2) -- (xhat2.west)
  node[lab, above, pos=-2] {$g_2$};
% skip connections
\draw[very thick, draw=gray!60, dashed, <->] (x2.north) -- (xhat2.south);
% REMOVED: \draw[very thick, draw=blue!70, dashed] (c2.south) -- (z2.north);

% ================== Connection between shared nodes ==================
\draw[very thick, red!70, dashed,<->] (c.east) -- (c2.west);

% ================== NEW: Cross-connections C1-Z2 and C2-Z1 ==================
\draw[very thick, draw=blue!70, dashed, <->] (c.east) -- (z2.west);   % C1 to Z2
\draw[very thick, draw=blue!70, dashed, <->] (c2.west) -- (z.east);   % C2 to Z1

\end{tikzpicture}
\caption{\ours: Information-guided Disentangled Representation Seeking.}
\label{fig:arch}
\end{figure}
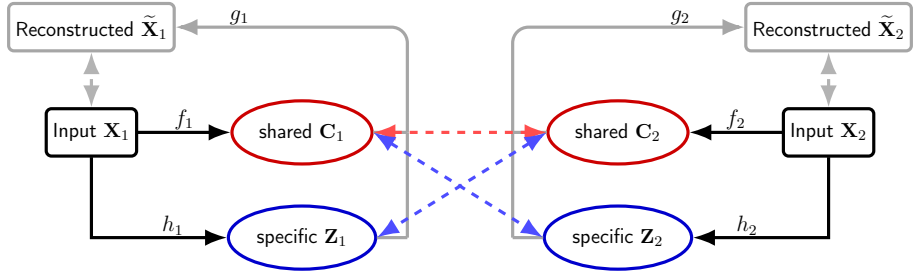

% Our position is not to redefine this conceptual objective \nb{need to modify, we are redefining...}, but to propose {\bf an effective way to implement it in practice}.
\yg{Our position is not only to refine this conceptual objective to relieve the issue with redundancy in shared features, but also to propose an effective and tractable way to implement it in practice.}
In the following subsections, we describe a two-stage strategy to this end:  
(1) learning shared features via CLIP, and  
(2) extracting modality-specific features with a reconstruction-guided objective conditional on learned shared features in step (1). See Figure~\ref{fig:arch} for an overview of \ours.

\subsection{Step 1: Extracting Shared Features via CLIP}
As noted at the end of Section~\ref{sec:problem-setup}, shared feature extraction is commonly handled by 
contrastive learning methods such as CLIP~\citep{radford2021learning}, which we also adopt. 
Concretely, encoders $(f_1, f_2)$ are trained with the InfoNCE loss \citep{oord2018representation}, which provides a lower bound on the mutual information between $X_1$ and $X_2$.  
This encourages \yg{$C_1 = f_1(X_1)$} and \yg{$C_2 = f_2(X_2)$} to capture all cross-modal dependence while avoiding modality-specific redundancy.

\subsection{Step 2: Extracting Modality-Specific Features}

Given the shared features \yg{$(C_1, C_2)$,} we aim to extract the modality-specific components $(Z_1, Z_2)$ 
by operationalizing the constrained problem in Eq.~\eqref{eq:obj-specific}.
\yg{Here, we note that, if the learned shared representations are redundant, that is, $C_1$ contains more information than the underlying shared features, learning the unique feature $Z_1$ by encouraging the independence between $Z_1$ and $C_1$ is lossy: the learned $Z_1$ may lose modality-specific information. To address this, we propose the \emph{cross-modal disentanglement}:  
\begin{align}\label{eq:lagrange}
    \min_{Z_i}\; \underbrace{I(Z_i; C_{3-i})}_{\text{disentanglement-enforcing term}} - \quad \lambda \cdot \underbrace{ I(Z_i, C_i; X_i)}_{\text{complementary information capture term}}.
\end{align}
The key is in the disentanglement term $I(Z_1, C_2)$ (or $I(Z_2, C_1)$). 
Even when the learned ``shared '' information is redundant, i.e., when $C_1$ and $C_2$ both contain the underlying shared features (say, $C^*$), the redundancy is only a function of modality-specific features in $X_2$ that is independent of $X_1$; thus, given that $I(C^*; Z_1) = I(C_2;Z_1)$, we ought to encourage the independence between $Z_1$ and $C_2$ instead.}

Here, the minimization over $Z_i$ is in fact minimizing over functions $h_i(\cdot)$ on $X_i$.  
This form highlights the tradeoff: reducing the dependence of $Z_i$ on the shared representation 
while maximizing the collective informativeness of $(Z_i, C_i)$ about $X_i$.
% \yg{We also note that, in the reconstruction term, \cite{wang2024information} adopts the shared feature from the other modality, i.e., the actual reconstruction in implementation becomes $I(Z_i, C_{3-i}; X_i)$ instead of $I(Z_i, C_i; X_i)$.}
However, both terms involve mutual information, which is intractable to optimize directly.  

To make this objective implementable, we replace  each term with a bound that aligns with its optimization direction:
\begin{itemize}%[itemsep=0pt, topsep=0pt, leftmargin=*]
    \item For the disentanglement enforcing term \yg{$I(Z_i; C_{3-i})$, which we aim to minimize, we replace it by an \textbf{upper bound}  
    given by the NCE-CLUB loss \citep{cheng2020club,liang2023factorized}:
    \begin{align}\label{eq:nce-club}
        \calL_{\rm NCE-CLUB}(Z_1; C_2) = \EE_{(Z_1,C_2)}\left[\log p(Z_1 \mid C_2)\right] - \EE_{Z_1, \tilde C_2}\left[\log p(Z_1 \mid \tilde C_2)\right],
    \end{align}
    {where $p(z_1 \mid c_2)$ is the underlying conditional density of $Z_1 \mid C_2$ and $\tilde C_2$ is an independent copy of $C_2$ from the same marginal distribution.\footnote{In implementation, following \cite{cheng2020club}, we replace conditional distributions by their estimates via multivariate Gaussian distributions whose moments are fitted from data in each epoch.}}  
    }
    \item For the complementary information capture term $I(Z_i, C_i; X_i)$, which we aim to maximize, we replace it by a \textbf{lower bound} 
    realized through a reconstruction loss that serves as a surrogate for conditional mutual information.  
\end{itemize}
This yields the practical {\ours} objective that we aim to minimize:
\yg{
\begin{align}\label{eq:ours-loss}
\mathcal{L}_{\text{\ours}}(Z_i; C_1, C_2, \lambda) \;=\; 
\mathcal{L}_{\text{NCE-CLUB}}(Z_i; C_{3-i}) \;+\;
\frac{\lambda}{2 \EE\|X_i\|^2} \min_{g_i} \EE\|g_i(Z_i, C_i) - X_i\|^2.
\end{align}
}
This formulation encourages $Z_i$ to be independent of the shared features 
while still retaining sufficient information measured via the optimal reconstruction error. 
Because the objectives are separable across modalities, 
the extraction procedure can be parallelized.

% Given the shared features, we extract modality-specific components by relaxing the ideal objective \eqref{eq:obj-specific}.  
% Direct mutual information objectives are intractable, so $\ours$ combines two tractable surrogates: {Edit later. }
% \begin{itemize}
%     \item the \textbf{NCE-CLUB loss}, a variational upper bound on $I(Z_i; C_i)$, and {Move the definition of the loss}
%     \item the \textbf{reconstruction error} $\|g_i(Z_i, C_i) - X_i\|^2$ as a surrogate for conditional mutual information.
% \end{itemize}

% This yields the $\ours$ objective: {need to add a bit of explanations on how to reach here. Shouldn't be long}
% \begin{align}
% \mathcal{L}_{\ours}(Z_i; C_i, \lambda) \;=\; 
% \mathcal{L}_{\text{NCE-CLUB}}(Z_i; C_i) \;+\;
% \frac{\lambda}{2} \min_{g_i} \|g_i(Z_i, C_i) - X_i\|^2.
% \end{align}

% This formulation guarantees that $Z_i$ is both independent from shared features 
% and sufficient for reconstructing the modality. 
% The procedure is parallelizable across modalities.  

\paragraph{Justification via reconstruction as MI bound.}  
{When $C_1$ is fixed,}
maximizing ${I(Z_1, C_1;X_1)} = I(Z_1; X_1 \mid C_1) {+ I(X_1;C_1)}$ is equivalent to {maximizing $I(Z_1;X_1\mid C_1) = H(X_1\mid C_1) - H(X_1\mid Z_1, C_1)$, which is further equivalent to} minimizing the conditional entropy 
$H(X_1 \mid Z_1, C_1)$. 
% Since $H(X_1 \mid C_1)$ is fixed once $C_1$ is given, the task reduces to minimizing 
% $H(X_1 \mid Z_1, C_1)$. 
By Fano's inequality \citep[Theorem 8.6.6]{cover1999elements}, 
the reconstruction error upper bounds this conditional entropy for any $g_1$:
\[
H(X_1 \mid Z_1, C_1) \;\leq\; \tfrac{1}{2} \log \!\Big(2\pi e \cdot \EE\|X_1 - g_1(Z_1, C_1)\|^2\Big),
\]
where equality holds if and only if the conditional distribution of $X_1 \mid (Z_1, C_1)$ is Gaussian.  
This shows why quadratic reconstruction loss is a principled surrogate for conditional mutual information.

\subsection{Comparison with two SOTA Methods}
Recent approaches to disentangled multi-modal representation learning differ mainly in how they approximate the two terms in Eq.~\eqref{eq:obj-specific}: \yg{the disentanglement enforcing term $I(Z_i, C_j)$, and the complementary information capture term $I(Z_i, C_{j'}; X_i)$, where $j, j' \in \{1,2\}$ will be specified in each method.
}

\paragraph{Factorized Contrastive Learning (Factorized CL, \citealp{liang2023factorized}).} 

Like \ours, this method uses the NCE-CLUB loss to upper bound the entanglement enforcing term. However, it replaces the complementary information capture term with an InfoNCE lower bound. For comparison, we consider its task-agnostic, two-step variant:
\yg{
\begin{align}\label{eq:fac}
    \calL_{\rm FactorizedCL}(Z_1; C_1, C_2, \tau, \lambda) = \calL_{\rm NCE-CLUB}(Z_1;C_1) + \frac{\lambda}{2} \calL_{\rm infoNCE}(h(C_1,Z_1),X_1,\tau).
\end{align}
We note that here $j, j'$ in \eqref{eq:obj-specific} both equal $i$, i.e., no cross-modal switching is implemented.
}
\cite{liang2021multibench} also proposes a self-supervised variant of Factorized CL by leveraging task-relevant augmentations, more discussions on which are deferred to Section~\ref{sec:discuss} and Appendix~\ref{sec:app-metric}. In this paper, we adapt the unsupervised version of Factorized CL, i.e., we use \texttt{CLUBInfoNCECritic} between corresponding features in \cite{liang2023factorizedgithub} as the disentanglement objective, and adopt the implementation of CLUB loss in \cite{cheng2020club} in both \ours\; and Factorized CL for fair comparison. (more details are presented in Appendix~\ref{sec:app-metric}).
% This method extends CLIP-style contrastive training by adding orthogonality constraints between 
% shared and modality-specific features. While simple and effective in some cases, 
% its orthogonality penalty is heuristic and does not provide information-theoretic guarantees of disentanglement.  

\paragraph{Disentangled Self-Supervised Learning (\disen~\citealp{wang2024information}).} 
This method also relies on InfoNCE for the complementary information capture term, but replaces NCE-CLUB with an orthogonal loss based on von Mises–Fisher assumptions: 
\yg{
\begin{align}\label{eq:disen-info}
    \calL_{\rm InfoDisen}(Z_1;C_1,C_2,\tau,\lambda) = \EE_{(Z_1,C_1)} \left[\langle \mu(Z_1), \mu(C_1) \rangle\right]+ \frac{\lambda}{2} \calL_{\rm infoNCE}(h(C_2,Z_1),X_1,\tau).
\end{align} 
We note that, in the reconstruction term, \cite{wang2024information} adopts the shared feature from the other modality, i.e., the actual reconstruction of interest becomes $I(Z_i, C_{3-i}; X_i)$ instead of $I(Z_i, C_i; X_i)$ by replacing $j$ with $3-i$ in \eqref{eq:obj-specific}.
}

% \begin{align}
%     \calL_{\rm otho}(Z_1,C_1) = \EE_{(Z_1,C_1)} \left[\langle \mu(Z_1), \mu(C_1) \rangle\right].
% \end{align}

% We also note that the orthogonal is also related to the asymptotics of infoNCE loss with an infinite temperature. To see this, as $\tau \rightarrow +\infty$, we have
% \begin{align}
%     \calL_{\rm infoNCE}(Z_1,C_1;\tau) = -\frac{2}{\tau}\Tr\left[\cov(Z_1,C_1)\right] + O(\tau^{-2}).
% \end{align}

Both methods inherit limitations: InfoNCE is sensitive to temperature tuning $\tau$ and may underperform as a surrogate for capturing complementary information, while the orthogonal loss assumes linearity and fails to capture nonlinear dependencies.
Additionally, without cross-modal disentanglement, \disen\; and FactorizedCL may fail to fully capture modality-specific features due to redundant shared features.
\ours~avoids these issues by pairing NCE-CLUB with a reconstruction-based mutual information bound.

% We note that the infoNCE loss depends on the choice/optimization of the temperature $\tau$, which introduces additional complexity and uncertainty in loss minimization \citep{geiping2023cookbook,gui2023unraveling,wang2020understanding,wang2021understanding}. In addition, the orthogonal loss heavily relies on the von Mises-Fisher family and, in general, may not be able to capture non-linear dependence between shared and modality-specific features.

\section{Experiments with a CITE-seq dataset}\label{sec:citeseq}

In this section, we evaluate the performance of {\ours} on applications in single-cell biology.
We consider the bone marrow CITE-seq dataset from \cite{stuart2019comprehensive} which consists of measurements in two modalities on $30672$ individual cells: transcriptome (RNA) and $25$ cell-surface proteins (ADT). 
% There are a total of  scRNA-seq profiles, and 
We randomly split the dataset into a training set of size $15000$ and a test set of size $15672$.
% As a benchmark in 
To leverage both RNA and ADT data for annotating cell types, \cite{hao2021integrated} proposed to cluster cells according to a weighted-nearest-neighbor (WNN) graph with cells as nodes. In its construction, the similarity between a cell $i$ and every other cell $j$ is measured by a weighted average of their respective similarities in the two modalities. 
The weights used by cell $i$ are in turn determined by the cross-modal predictive powers of its individual modalities after local smoothing.
Thus, the RNA weight of each cell can be viewed as a quantification of the importance RNA plays in determining its cell state. The larger, the more important.
Based on clustering nodes of the WNN graph, \cite{hao2021integrated} annotated cells at two granularity levels with $5$ (level-1) and $27$ (level-2) cell types, respectively.

\begin{figure}[ht]
    \centering
    \includegraphics[width=\linewidth]{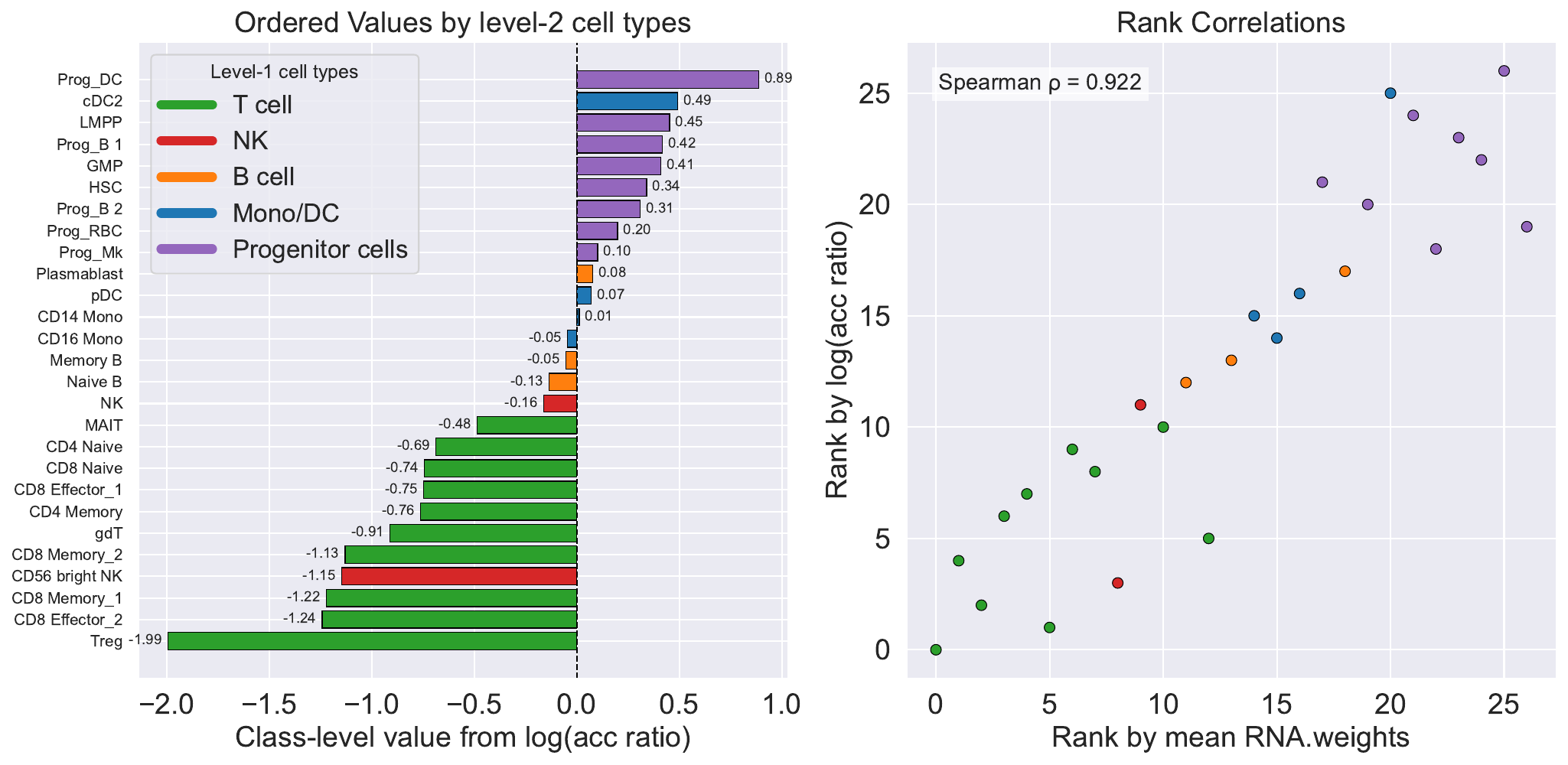}
    \caption{\yg{Performance of {\ours} in CITE-seq dataset ($\lambda=10.0$).}}
    \label{fig:cite-seq-recons}
\end{figure}

In this experiment, we first apply {\ours} to train neural nets\footnote{In this example, representation maps are trained within the class of $5$-layer ReLU neural networks with middle layers of width $50$.} 
that map RNA and ADT data to their disentangled representations, $(C_{\rm RNA}, C_{\rm ADT})$ and $(Z_{\rm RNA}$, $Z_{\rm ADT})$, using the training set.
We then apply the trained networks to infer disentangled representations of cells in the test set. 
The ideal disentangled representations should reflect distinctive levels of informativeness that individual modalities have in determining each cell's annotation.
To this end, for each cell $c$, we find its 10 nearest neighbors measured by Euclidean distance in $Z_{\rm RNA}$ and $Z_{\rm ADT}$, respectively, and compare the proportions of neighbors with the same level-2 annotations as $i$ in the two modalities, denoted by $\beta_{\rm RNA}(c)$ and $\beta_{\rm ADT}(c)$, respectively. 
If the two proportions differ sizably, the modality-specific information in the higher-proportion modality plays a more important role in determining the cell's annotation, which motivates us to summarize the comparison with a score $\theta(c) = \log(\beta_{\rm RNA}(c) / \beta_{\rm ADT}(c))$ that is monotone increasing with respect to the importance of RNA-specific information.

\begin{figure}[t]
    \centering
    \includegraphics[width=0.6\linewidth]{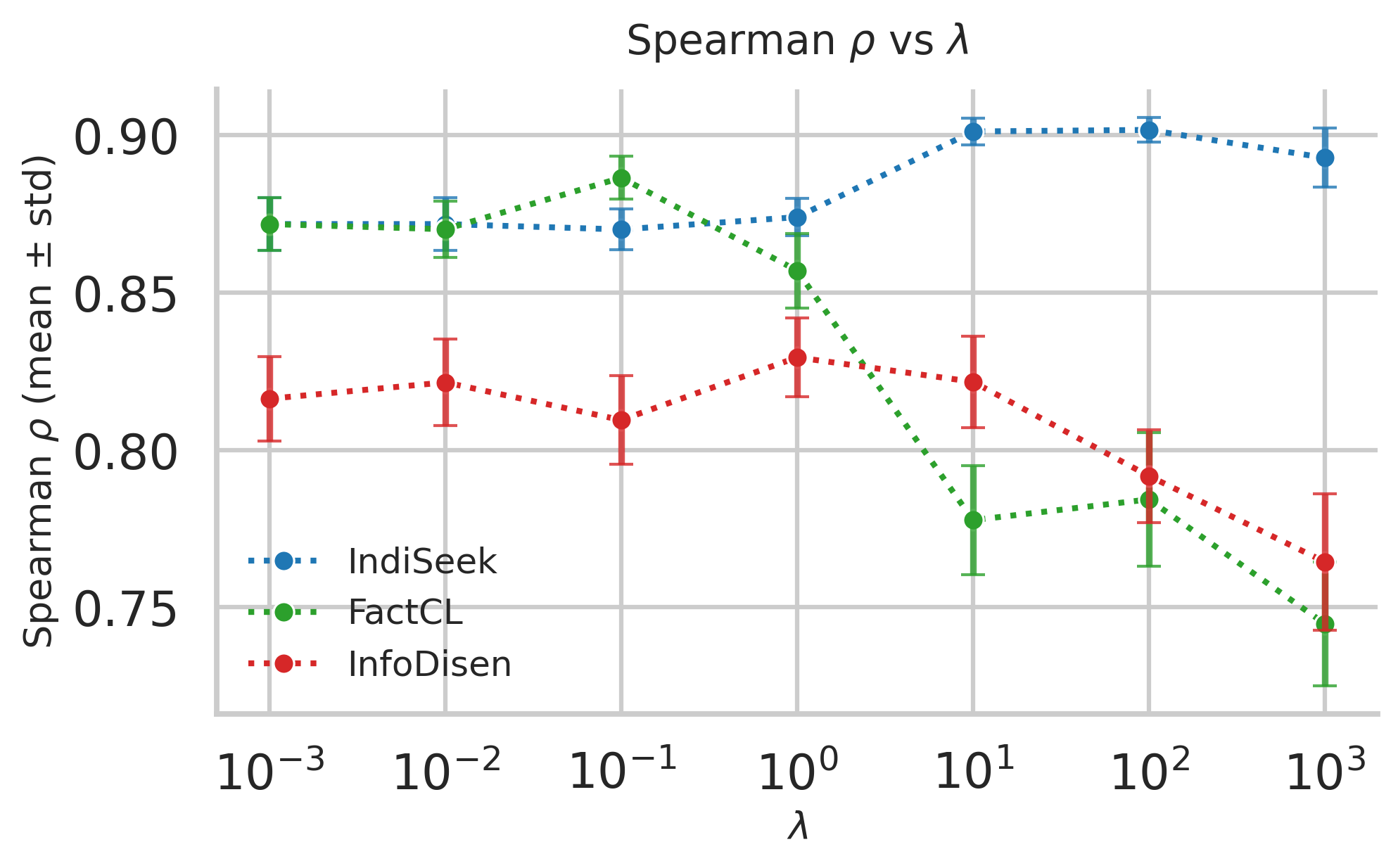}
    \caption{\yg{Comparison of rank correlation metrics across three methods on the CITE-seq dataset.}}
    \label{fig:rank-corr-comparison}
\end{figure}

In the left panel of Figure~\ref{fig:cite-seq-recons}, $27$ level-2 cell types are ranked according to average $\theta(c)$ scores of cells in the test set with respective cell type annotations,
in which a higher rank indicates a higher impact of the RNA-specific information in determining the cell type.
Each bar is colored according to the coarser level-1 cell type that each level-2 cell type belongs to for easier visual inspection.
Different cell types within the T cell family have negative scores, indicating RNA-specific information is less informative compared to ADT-specific counterpart for differentiating different T cell subpopulations, which is in line with the observation that clustering transcriptomics data typically does not delineate T cell subpopulations \citep{szabo2019single, zheng2021pan} and gating based on surface protein markers is often needed \citep{kotliar2025reproducible}. In contrast, RNA-specific information plays a more important role in distinguishing different progenitor cells than ADT-specific information.
To further demonstrate the quality of learned modality-specific representations, we plot the ranks of average $\theta(c)$ values
against the benchmark ranks of average RNA weights in \cite{hao2021integrated} of the 27 level-2 cell types in the right panel in Figure~\ref{fig:cite-seq-recons}, which exhibit a strong correlation that is confirmed by Spearman's rank correlation \citep{spearman1961proof} of $0.910$.
\yg{In Appendix~\ref{sec:app-citeseq}, we present UMAPs of IndiSeek-learned representations \yg{in Figures~\ref{fig:umap-citeseq-indiseek} and \ref{fig:umap-citeseq-indiseek-concat}}, which exhibit clear clusters with respect to two levels of cell types that are also confirmed with metrics such as ARI and NMI (details are given in Appendix~\ref{sec:app-citeseq}).}
As a comparison, we repeated the foregoing experiments across $10$ random seeds while replacing {\ours} with {\disen} or FactorizedCL when training the same neural net architecture and the resulting alignment metrics with respect to the results in \cite{hao2021integrated} are presented in Figure~\ref{fig:rank-corr-comparison}, \yg{in which we also report the error bars for results over $10$ random seeds}. 
Both alternative methods yielded inferior alignment with the relative importance of the RNA modality in \cite{hao2021integrated} compared to {\ours}.
To ablate the choice of $\lambda$, in all experiments with real-world data, we report results with varying $\lambda$ on grid values $10^j$ for $j=-3,-2,-1,0,1,2,3$.
\yg{Additionally, UMAPs of representations learned by {\disen} and FactorizedCL are also presented in Appendix~\ref{sec:app-citeseq} Figures~\ref{fig:umap-citeseq-fact}, \ref{fig:umap-citeseq-disen}, \ref{fig:umap-citeseq-fact-concat}, and \ref{fig:umap-citeseq-disen-concat}.}
% \nb{add pointer to UMAPs for InfoDisen and FactorizedCL in appendix.}

\section{Experiments with MultiBench datasets}\label{sec:multibench}
In addition to the application in single-cell multi-omics, we also evaluate the performance of {\ours} on MultiBench datasets \citep{liang2021multibench}, which include 4 video datasets (MOSI \citep{zadeh2016mosi}, MOSEI \citep{zadeh2018multimodal}, UR-FUNNY \citep{hasan2019ur}, MUStARD \citep{castro2019towards}) and Medical Information Mart for Intensive Care III (MIMIC) dataset \citep{johnson2016mimic}.
We follow the processing pipeline in \cite{liang2023factorized} and adopt exactly the same data splitting and feature pre-extraction, in which we focus on the task of binary classification on whether the patient fits any ICD-9 code in group 7 for the MIMIC dataset. We use a smaller Transformer architecture with $2$ heads, $2$ layers, and the intermediate layers with a width $128$. For all methods in comparison, we train the architecture for $2000$ epochs with the same training parameters. 
We note that our focus here is to benchmark the qualities of representations learned with task-agnostic objective functions. Therefore, for fair comparison along this direction, we do not involve task-related within- or cross-modality data augmentations and hence do not involve self-supervised learning terms in objective functions of all methods in comparison. 
In contrast, to achieve the optimal performance in a specific task on a particular benchmark dataset, appropriate data augmentation informed by the task and domain knowledge is often helpful.

Following \cite{liang2023factorized} and \cite{wang2024information}, we focus on the linear probing accuracy for each method with learned representations $(C_1,Z_1,C_2,Z_2)$ as features, i.e., we concatenate shared and modality-specific features for classification. Here we also compare with the CLIP baseline, where only the shared representations $(C_1,C_2)$ are used for linear probing accuracy evaluation.
{Following the same training procedure as for the CITE-seq dataset, we report the results with three methods in Table~\ref{tab:accuracy_summary}. For each method, we vary $\lambda$ in $\{10^j: j=-3,-2,-1,0,1,2,3\}$ and for each $\lambda$, we run each method with $10$ random seeds. Table~\ref{tab:accuracy_summary} presents the averaged accuracy for the best choice of $\lambda$. More details on the performance with varying $\lambda$'s can be found in Appendix~\ref{sec:app-vary-lam}.}

From Table~\ref{tab:accuracy_summary}, we can see that \ours~outperms baseline and two SOTA methods across all datasets, and the performances of three disentangled learning methods are comparable in datasets MUStARD and MIMIC, among which, MUStARD ($690$ samples) has a small scale. Moreover, all three disentangled learning methods outperform the CLIP baseline in most of the datasets, which demonstrates the necessity of involving modality-specific features in downstream tasks. 
In video datasets, such as MOSI ($2199$ samples), MOSEI ($22777$ samples), and UR-FUNNY ($16514$ samples), \ours~exhibits more pronounced performance gain in accuracy under the same training configurations, indicating its efficiency in enforcing disentanglement and preserving sufficiency in representations.  

\yg{
\paragraph{Ablation studies with varying output dimension.}
To investigate the robustness of \ours~ across different output dimension capacities, we conduct experiments on the MOSI dataset with output dimensions $d \in \{ 20, 60, 100, 140, 180\}$. For each output dimension, we perform a sweep over $\lambda$ and report the maximum accuracy achieved across all $\lambda$ values, averaged over 10 random seeds.
Table~\ref{tab:mosi_outdim} presents the results of this analysis. \ours~ consistently achieves the best performance across all output dimensions, demonstrating its robustness to this hyperparameter choice. 
}
% The optimal performance is obtained at $d=60$, suggesting that moderate-dimensional representations provide a good balance between expressiveness and overfitting, especially with a relatively small dataset size of MOSI. 
% Performance of other methods slightly degrades at higher dimensions ($d=60, 100$), likely due to the complexity of loss objectives as well as increased model complexity relative to the small dataset size of MOSI.
% \begin{table}[ht]
% \centering
% \caption{Comparison of accuracy on MultiBench datasets (max over $\lambda$, averaged over 10 seeds).}
% \begin{tabular}{lccccc}
% \toprule
% \textbf{Method} & \textbf{MOSI} & \textbf{MOSEI} & \textbf{UR-FUNNY} & \textbf{MUStARD} & \textbf{MIMIC} \\
% \midrule
% Our method (\ours)   & \textbf{56.81\%} & \textbf{73.87\%} & \textbf{60.43\%} & \textbf{53.91\%} & \textbf{65.80\%} \\
% \disen               & 55.74\% & 73.17\% & 57.67\% & 53.41\% & 65.58\% \\
% FactorizedCL         & 55.92\% & 73.12\% & 57.61\% & 53.70\% & 65.69\% \\
% CLIP (baseline)      & 56.11\% & 72.49\% & 56.81\% & 53.77\% & 64.60\% \\
% \bottomrule
% \end{tabular}
% \label{tab:multibench}
% \end{table}

\begin{table}[ht]
\centering
\caption{\yg{Comparison of accuracy on multimodal datasets (averaged over 10 seeds, standard errors in parentheses, max average over $\lambda$). All values are percentages.}}
\label{tab:accuracy_summary}
\begin{tabular}{lccccc}
\toprule
\textbf{Method} & \textbf{MOSI} & \textbf{MOSEI} & \textbf{UR-FUNNY} & \textbf{MUStARD} & \textbf{MIMIC} \\
\midrule
IndiSeek         & \textbf{70.03}$_{\scriptstyle(1.39)}$ & \textbf{75.47}$_{\scriptstyle(0.13)}$ & \textbf{63.79}$_{\scriptstyle(0.39)}$ & \textbf{57.46}$_{\scriptstyle(1.04)}$ & \textbf{65.99}$_{\scriptstyle(0.31)}$ \\
FactorizedCL     & 67.11$_{\scriptstyle(0.34)}$ & 74.74$_{\scriptstyle(0.04)}$ & 58.36$_{\scriptstyle(0.25)}$ & 56.45$_{\scriptstyle(1.16)}$ & 65.69$_{\scriptstyle(0.11)}$ \\
InfoDisen        & 67.52$_{\scriptstyle(0.62)}$ & 74.73$_{\scriptstyle(0.08)}$ & 58.08$_{\scriptstyle(0.50)}$ & 56.16$_{\scriptstyle(0.92)}$ & 65.47$_{\scriptstyle(0.23)}$ \\
CLIP (baseline)  & 67.61$_{\scriptstyle(0.66)}$ & 74.70$_{\scriptstyle(0.05)}$ & 58.32$_{\scriptstyle(0.68)}$ & 55.36$_{\scriptstyle(1.12)}$ & 64.56$_{\scriptstyle(0.29)}$ \\
\bottomrule
\end{tabular}
\end{table}

% \begin{table}[ht]
% \centering
% \caption{\yg{MOSI results with varying output dimensions (max over $\lambda$, averaged over 10 seeds). All values are percentages.}}
% \label{tab:mosi_outdim}
% \begin{tabular}{lcccc}
% \toprule
% \textbf{Method} & \textbf{outdim=20} & \textbf{outdim=60} & \textbf{outdim=100} & \textbf{outdim=140} \\
% \midrule
% IndiSeek         & \textbf{69.68}$_{\scriptstyle(0.51)}$ & \textbf{70.03}$_{\scriptstyle(0.80)}$ & \textbf{69.56}$_{\scriptstyle(0.57)}$ & \textbf{70.20}$_{\scriptstyle(0.65)}$ \\
% FactorizedCL     & 69.49$_{\scriptstyle(0.26)}$ & 66.91$_{\scriptstyle(1.04)}$ & 67.52$_{\scriptstyle(0.95)}$ & 68.28$_{\scriptstyle(0.67)}$ \\
% InfoDisen        & 68.57$_{\scriptstyle(0.91)}$ & 65.95$_{\scriptstyle(1.59)}$ & 65.17$_{\scriptstyle(0.68)}$ & 66.90$_{\scriptstyle(0.33)}$ \\
% CLIP (baseline)  & 69.46$_{\scriptstyle(0.45)}$ & 67.03$_{\scriptstyle(0.79)}$ & 66.53$_{\scriptstyle(0.68)}$ & 68.54$_{\scriptstyle(0.94)}$ \\
% \bottomrule
% \end{tabular}
% \end{table}

\begin{table}[ht]
\centering
\caption{\yg{MOSI results with varying output dimensions 
(averaged over 10 seeds, standard errors in parentheses, max average over $\lambda$). All values are percentages.}}
\label{tab:mosi_outdim}
\begin{tabular}{lccccc}
\toprule
\textbf{Method} & \textbf{outdim=20} & \textbf{outdim=60} & \textbf{outdim=100} & \textbf{outdim=140} & \textbf{outdim=180} \\
\midrule
IndiSeek         & \textbf{69.68}$_{\scriptstyle(0.51)}$ & \textbf{70.03}$_{\scriptstyle(0.80)}$ & \textbf{69.56}$_{\scriptstyle(0.57)}$ & \textbf{70.20}$_{\scriptstyle(0.65)}$ & \textbf{68.80}$_{\scriptstyle(1.15)}$ \\
FactorizedCL     & 69.49$_{\scriptstyle(0.26)}$ & 66.91$_{\scriptstyle(1.04)}$ & 67.52$_{\scriptstyle(0.95)}$ & 68.28$_{\scriptstyle(0.67)}$ & 67.08$_{\scriptstyle(0.29)}$ \\
InfoDisen        & 68.57$_{\scriptstyle(0.91)}$ & 65.95$_{\scriptstyle(1.59)}$ & 65.17$_{\scriptstyle(0.68)}$ & 66.90$_{\scriptstyle(0.33)}$ & 66.62$_{\scriptstyle(0.68)}$ \\
CLIP (baseline)  & 69.46$_{\scriptstyle(0.45)}$ & 67.03$_{\scriptstyle(0.79)}$ & 66.53$_{\scriptstyle(0.68)}$ & 68.54$_{\scriptstyle(0.94)}$ & 67.20$_{\scriptstyle(1.04)}$ \\
\bottomrule
\end{tabular}
\end{table}

\paragraph{Multi-task performance.}
To further evaluate the learned task-agnostic representations,
we also investigate the multi-task performance of each method on the MIMIC dataset. 
We consider the tasks of predicting ICD-9 codes for multiple groups in the MIMIC dataset in MultiBench \citep{liang2021multibench}. We compare the performance of all three disentangled learning methods together with CLIP baseline in Table~\ref{tab:multi}, in which we calculate the average accuracy across tasks of predicting the last three ICD-9 codes, and present results in a subset of ICD-9 codes. 
Complete results for other tasks are deferred to the appendix.

\begin{table}[ht]
\centering
\caption{\yg{Multi-task performance (averaged over 10 seeds, standard errors in parentheses, max average over $\lambda$) on MIMIC dataset (ICD-9 groups 17–19). All values are percentages.}}
\label{tab:multi}
\begin{tabular}{ccccc}
\toprule
Method & group 17 & group 18 & group 19 & Average (3 tasks) \\
\midrule
IndiSeek  & \textbf{62.58}$_{\scriptstyle(0.17)}$ & \textbf{60.99}$_{\scriptstyle(0.19)}$ & \textbf{69.71}$_{\scriptstyle(0.11)}$ & \textbf{64.43} \\
FactorizedCL    & 61.83$_{\scriptstyle(0.20)}$ & 60.36$_{\scriptstyle(0.14)}$ & 69.30$_{\scriptstyle(0.12)}$ & 63.83 \\
InfoDisen & 61.72$_{\scriptstyle(0.14)}$ & 60.40$_{\scriptstyle(0.34)}$ & 69.46$_{\scriptstyle(0.14)}$ & 63.86 \\
CLIP (baseline)     & 61.50$_{\scriptstyle(0.14)}$ & 60.71$_{\scriptstyle(0.09)}$ & 69.25$_{\scriptstyle(0.16)}$ & 63.82 \\
\bottomrule
\end{tabular}
\end{table}

\section{Extensions and discussion}\label{sec:discuss}

Guided by information-theoretic principles, we proposed {\ours}, a method that extracts shared and modality-specific features that are simultaneously sufficient and disentangled. While our focus has been on task-agnostic disentangled representation learning, {\ours} can be readily extended to task-related settings.  

\paragraph{Task-related extensions.}  
\begin{itemize}%[itemsep=0pt, topsep=0pt, leftmargin=*]
    \item \textbf{With task-specific augmentations.}  
    When a downstream task is known in advance, domain knowledge can be used to design within-modality data augmentations. If the augmentations are \emph{ideal} in the sense that the task-invariant orbit of each observation is fully covered~\citep{liang2023factorized}, one may first perform within-modality contrastive learning. The resulting representations retain only task-relevant information. {\ours} can then be applied to further decompose these task-relevant features into shared and modality-specific components.  

    \item \textbf{With side information $Y$.}  
    When auxiliary task-related information $Y$ is available, {\ours} can treat $Y$ as an additional modality and be applied to $(X_i, Y)$ pairs. The modality-specific features of each $X_i$ relative to $Y$ can then guide the design of effective augmentations, after which the procedure above can be repeated on the task-related representations. In the extreme case where $Y$ is the task label itself, each $X_i$ can first be reduced to its shared features with $Y$, denoted $\widetilde{X}_i$, since $\widetilde{X}_i$ contains all information in $X_i$ about $Y$. {\ours} may then be applied to $\widetilde{X}_1, \widetilde{X}_2$ for further decomposition.  
\end{itemize}

\paragraph{Tuning parameter selection.}
\yg{The selection of the Lagrange multiplier $\lambda$ in \eqref{eq:ours-loss} is important for the implementation of \ours. 
With the rescaled reconstruction loss, both the CLUB upper bound and the reconstruction term are scale-free, and our experimental results have suggested a candidate range of $\lambda$'s between $10^{-1}$ to $10^1$. 
% the optimal $\lambda$ may itself be scale-dependent and vary across datasets. 
Identifying the precise optimal choice of $\lambda$ remains an interesting future direction to enhance empirical performance and theoretical understanding.
}

% We focus on task agnostic rep
% If you have multiple tasks: 
% 1) remove task irrelevant features/noise
% 2) data augmentation
\paragraph{Additional related work.}  
In the literature, there are other alternatives that learn both shared and modality-specific features in an information-theoretic manner. For example, \cite{dufumier2024align} adopts infoNCE as a proxy for mutual information.
However, the framework in \cite{dufumier2024align} cannot ensure disentanglement between shared and modality-specific features, leading to potential redundancy in learned representations. 
In a concurrent work, \cite{shi2025towards} adopts variational bounds for mutual information and entropy regularization as objectives to learn disentangled features. However, similar to \cite{liang2023factorized} and \cite{dufumier2024align}, the objectives are motivated by some task labels $Y$ and hence differ from our goal of learning task-agnostic features.

%% file: appendix.tex
\section{Additional experimental results}

\subsection{Implementation details and metrics}\label{sec:app-metric}

\paragraph{Masking-based feature importance.}
In simulated experiments, to determine which coordinates are extracted as shared or modality-specific features in $X_1$, we consider model-free feature importance based on feature masking related to \cite{robnik2008explaining,zeiler2014visualizing,li2016understanding}. 
Concretely, given a black-box architecture $\psi: \RR^d \rightarrow \RR^p$, to understand which coordinates in each $x \in \RR^d$ are effective in the output $\psi(x)$, we use leave-one-coordinate-out $x_{-j}$ as inputs for $j \in [d]$ and quantify the difference between and original output and output with masked input $\|\psi(x) - \psi(x_{-j})\|^2$. In implementation, for each coordinate $j \in [d]$, we draw $M = 1000$ random vectors $\{x_i\}_{i \in [M]}$ from uniform distributions on $[-10,10]^d$ and calculate the average
\begin{align}
    \zeta_{\psi}(j) = \frac{1}{M} \sum_{i \in [M]} \|\psi(x_i) - \psi(x_{i,-j})\|^2.
\end{align}
In this paper, we normalize the vector $\bzeta_{\psi} = (\zeta_{\psi}(j))_{j \in [p]}$ to the simplex with unit $\ell_1$-norm and use this normalized score $\hat \zeta_{\psi}(j)$ to measure the importance of the $j$th coordinate in $\psi$.

\paragraph{Rank correlation.}

Recall that we obtain disentangled representations for RNA and ADT data, $(C_{\rm RNA}, C_{\rm ADT})$ and $(Z_{\rm RNA}$, $Z_{\rm ADT})$, using the training set.
To this end, for each cell $c$, we find its 10 nearest neighbors measured by Euclidean distance in $Z_{\rm RNA}$ and $Z_{\rm ADT}$, respectively, and compare the proportions of neighbors with the same level-2 annotations as $i$ in the two modalities, denoted by $\beta_{\rm RNA}(c)$ and $\beta_{\rm ADT}(c)$, respectively. 
We summarize the comparison with a score $\theta(c) = \log(\beta_{\rm RNA}(c) / \beta_{\rm ADT}(c))$ that is monotone increasing with respect to the importance of RNA-specific information, and rank $27$ cell types based on this metric. 
We use the RNA weights based on weighted k-NN obtained in \cite{hao2021integrated} as the benchmark and compare our ranks $(R_i)_{i \in [L]}$ for cell types with the RNA-weights-based ranks $(R^*_i)_{i \in [L]}$ with $L=27$. We adopt Spearman's $\rho$ correlation as the metric, where for any two vectors of ranks as permutations of $[L]$,
\begin{align}
    \rho(\bR, \bR^*) = \frac{\sum_{i \in [L]} \left(R_i - \frac{L(L+1)}{2}\right)\left(R^*_i - \frac{L(L+1)}{2}\right)}{\sqrt{\sum_{i \in [L]} \left(R_i - \frac{L(L+1)}{2}\right)^2}\sqrt{\sum_{i \in [L]} \left(R^*_i - \frac{L(L+1)}{2}\right)^2}}. 
\end{align}
Moreover, when there are no ties in both $\bR$ and $\bR^*$, it holds that
\begin{align}
    \rho(\bR, \bR^*) = 1 - \frac{6\sum_{i \in [L]} (R_i - R^*_i)^2}{L(L^2-1)}.
\end{align}

\yg{
\paragraph{Implementation of CLUB loss.}
We follow the implementation of CLUB in \cite{cheng2020club}.
As CLUB loss is a variational upper bound for mutual information, we use an auxiliary conditional density $q_{\theta}(z | c)$ as an approximation for $p(c | z)$, where $q_{\theta}(z | c)$ lies in the Gaussian family, and we optimize the moments of the Gaussian distribution using 5-layer MLPs in each epoch. The inner optimization is run for 5 epochs at each epoch to optimize representations, and the implementation the CLUB loss is the same for both IndiSeek 
and FactorizedCL 
for fair comparison.}

\yg{
% For FactorizedCL, 
\cite{liang2021multibench} also proposes a self-supervised variant of Factorized CL by leveraging task-relevant augmentations, more discussions on which can be found in
% are deferred to 
Section~\ref{sec:discuss}. In this paper, we adapt the unsupervised version of Factorized CL, i.e., we use \texttt{CLUBInfoNCECritic} between corresponding features in \cite{liang2023factorizedgithub} as the disentanglement objective, and adopt the implementation of CLUB loss in \cite{cheng2020club} in both \ours\; and Factorized CL for fair comparison.
}

\subsection{Numerical simulations}\label{sec:app-simu}

In this section, we present additional results for the same setting in the main paper with varying $\lambda$'s. 
Recall the setting with two modalities with $d_1 = 6$ and $d_c=2$:
% and consider the following two settings
\begin{itemize}[itemsep=0pt, topsep=0pt, leftmargin=*]
    \item Setting 1: Let the observed data be iid copies of $X_1 {\sim} \calN(0, I_{d_1})$. For each $x\in \RR^{d_1}$, define the shared representation map
    \begin{align}
        f_1(x) = 0.5A_f x + 0.2 \sin(A_f x) + 0.2(A_f x)^3 \quad \text{with} \quad A_f = (\bI_{d_c}, \bO) \in \RR^{d_c \times d_1}.
    \end{align}
    Here, the sine and cubic functions are applied entrywise. 
    The ideal modality-specific features are the last four coordinates of $X_1$, which contain all remaining information while being independent of $C_1$.
    
    % Intuitively, the ideal set of modality-specific features that contain all remaining information about $X_1$ while being independent of $C_1$ should be a bijection of $(X_1)_{(d_c+1):d_1} \sim \calN(0,\bI_{d_1-d_c})$.  
    
    \item Setting 2: 
    Let the observed data be iid copies of $X_1$ where $(X_1)_{\{1,2,5,6\}} \sim \calN(0, \bI_4)$, $(X_1)_3 = 0.2\times((X_1)_1 + (X_1)_2)$, and $(X_1)_4 = (X_1)_1 \times (X_1)_2$. 
    Thus, the third and fourth coordinates are deterministic functions of the first two, while the others are independent.
    Let $C_1 = (X_1)_{1:d_c}$. 
    Here, the ideal modality-specific features are the last two coordinates. 
\end{itemize}

\begin{figure}[ht]
    \centering
    \begin{subfigure}[t]{0.225\linewidth}
        \centering
        \includegraphics[width=\linewidth]{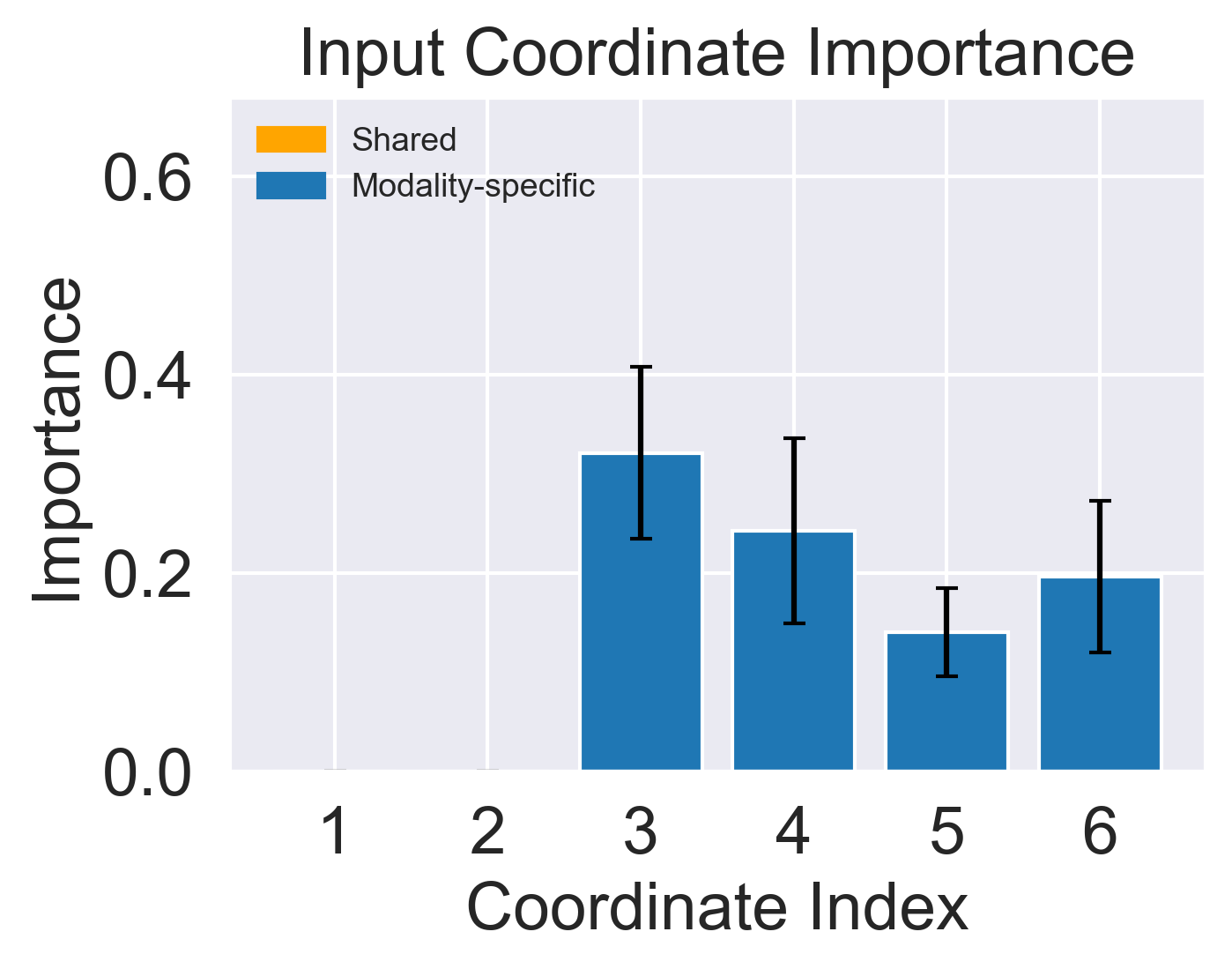}
        \caption{$\lambda=0.01$.}
        \label{fig:rec11}
    \end{subfigure}
    \hfill
    \begin{subfigure}[t]{0.225\linewidth}
        \centering
        \includegraphics[width=\linewidth]{figs/import_recons_setting1_10000_2_6_50_deep_0.1.png}
        \caption{$\lambda=0.1$.}
        \label{fig:rec12}
    \end{subfigure}
    \hfill
    \begin{subfigure}[t]{0.225\linewidth}
        \centering
        \includegraphics[width=\linewidth]{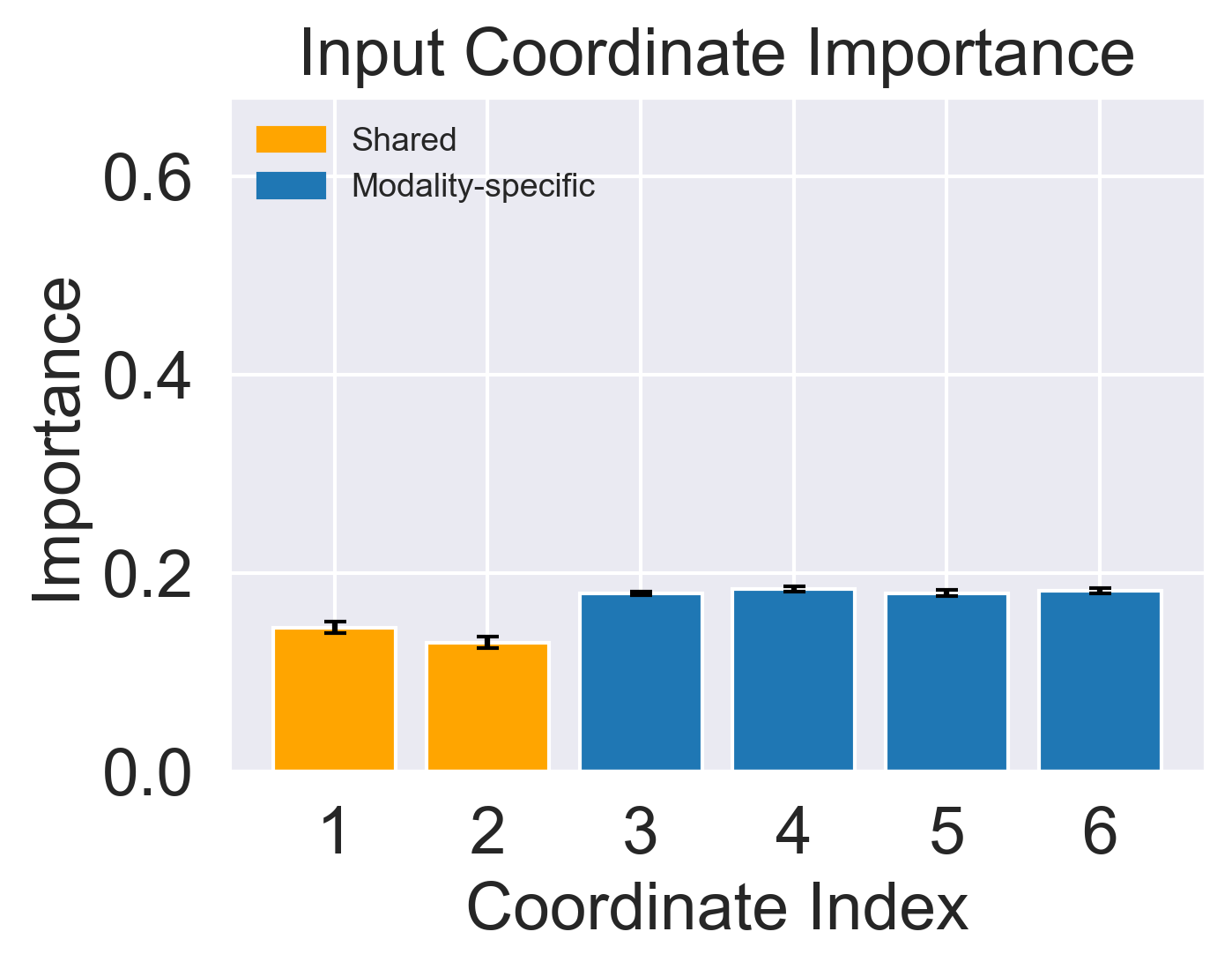}
        \caption{$\lambda=1.0$.}
        \label{fig:rec13}
    \end{subfigure}
    \hfill
    \begin{subfigure}[t]{0.225\linewidth}
        \centering
        \includegraphics[width=\linewidth]{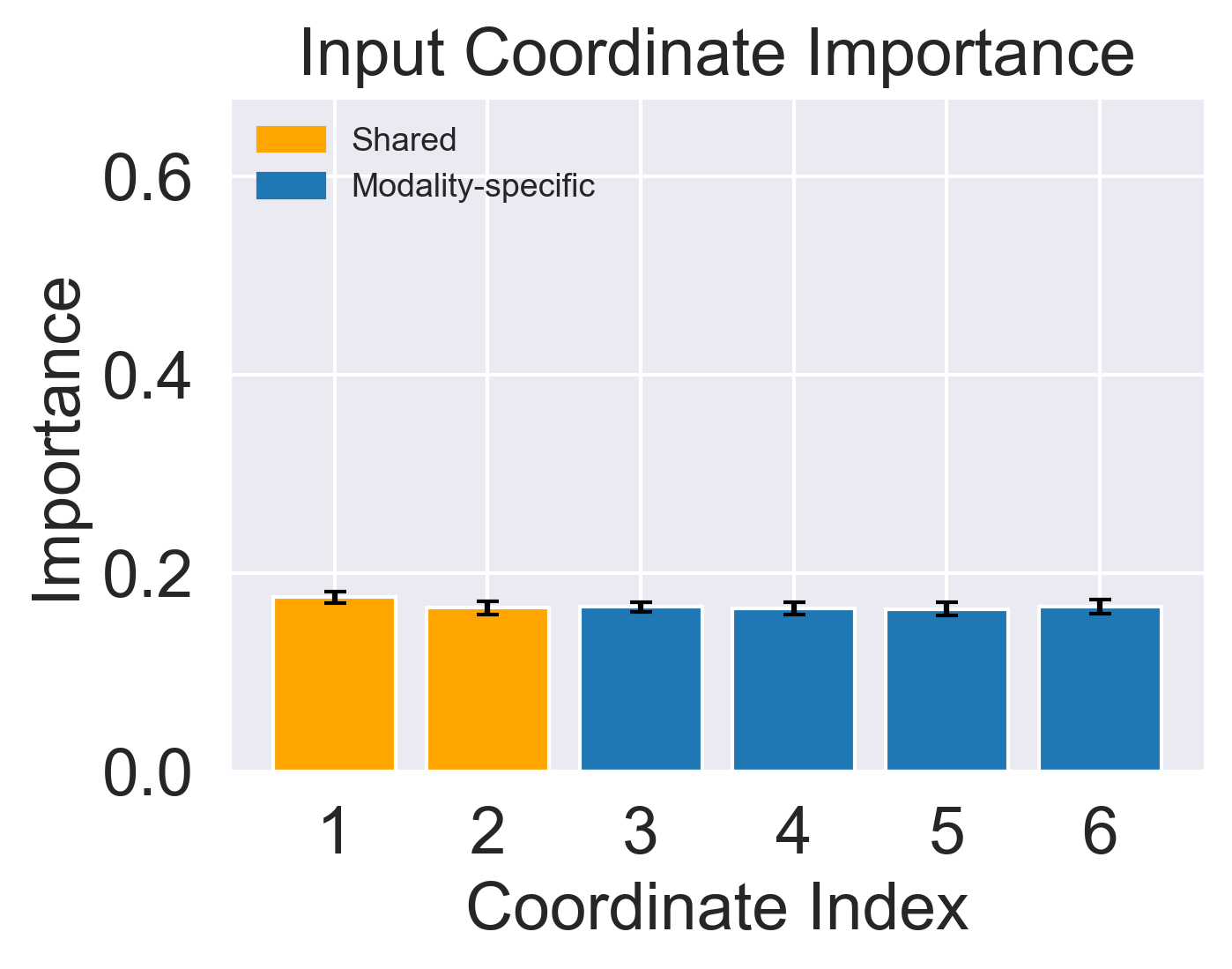}
        \caption{$\lambda=10.0$.}
        \label{fig:rec14}
    \end{subfigure}
    \caption{\yg{\ours~ with varying $\lambda$: feature importance averaged over $50$ simulations (Setting 1).}}
    \label{fig:setting1-recons}
\end{figure}

\begin{figure}[!h]
    \centering
    \begin{subfigure}[t]{0.225\linewidth}
        \centering
        \includegraphics[width=\linewidth]{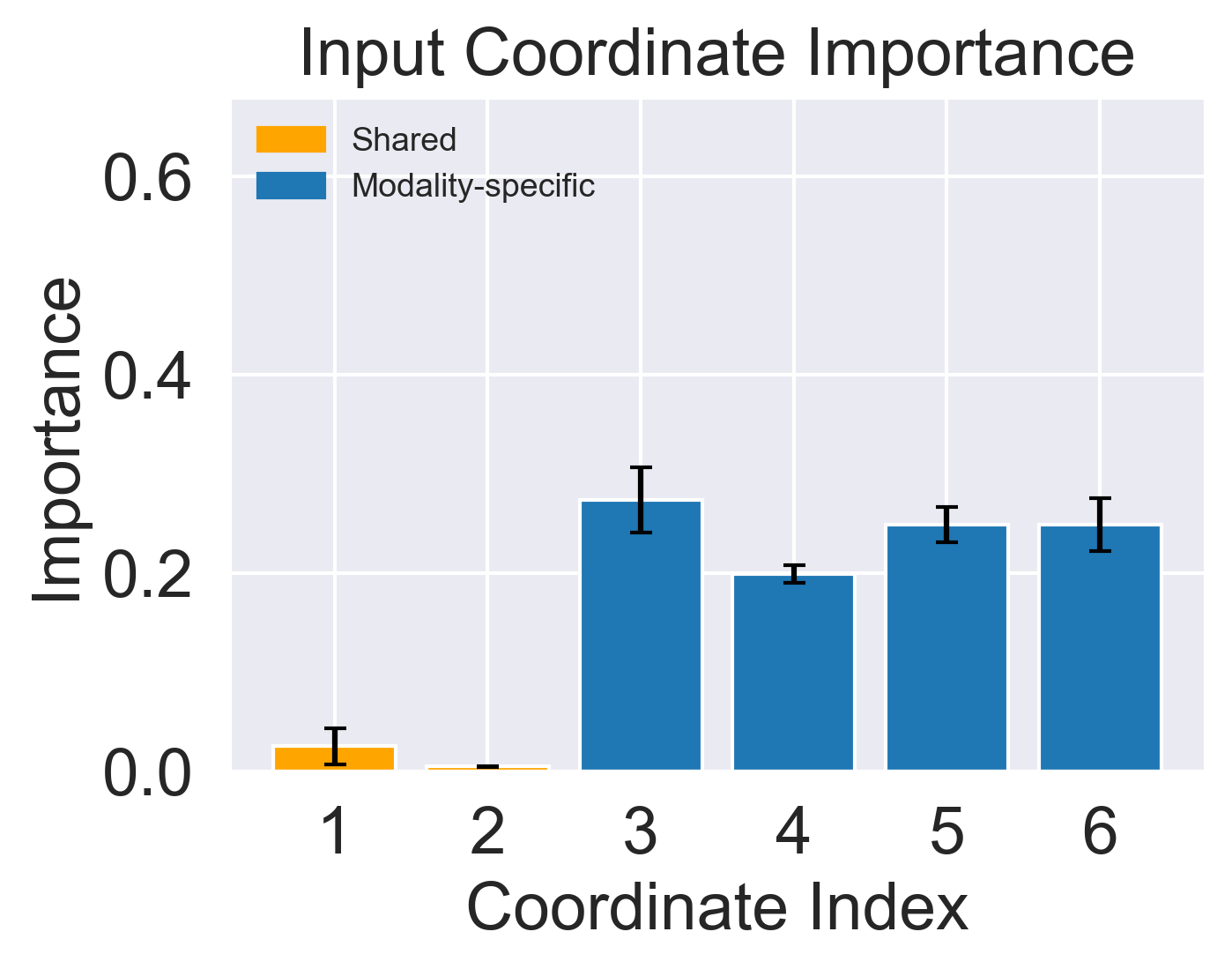}
        \caption{$\lambda=0.01$.}
        \label{fig:fact11}
    \end{subfigure}
    \hfill
    \begin{subfigure}[t]{0.225\linewidth}
        \centering
        \includegraphics[width=\linewidth]{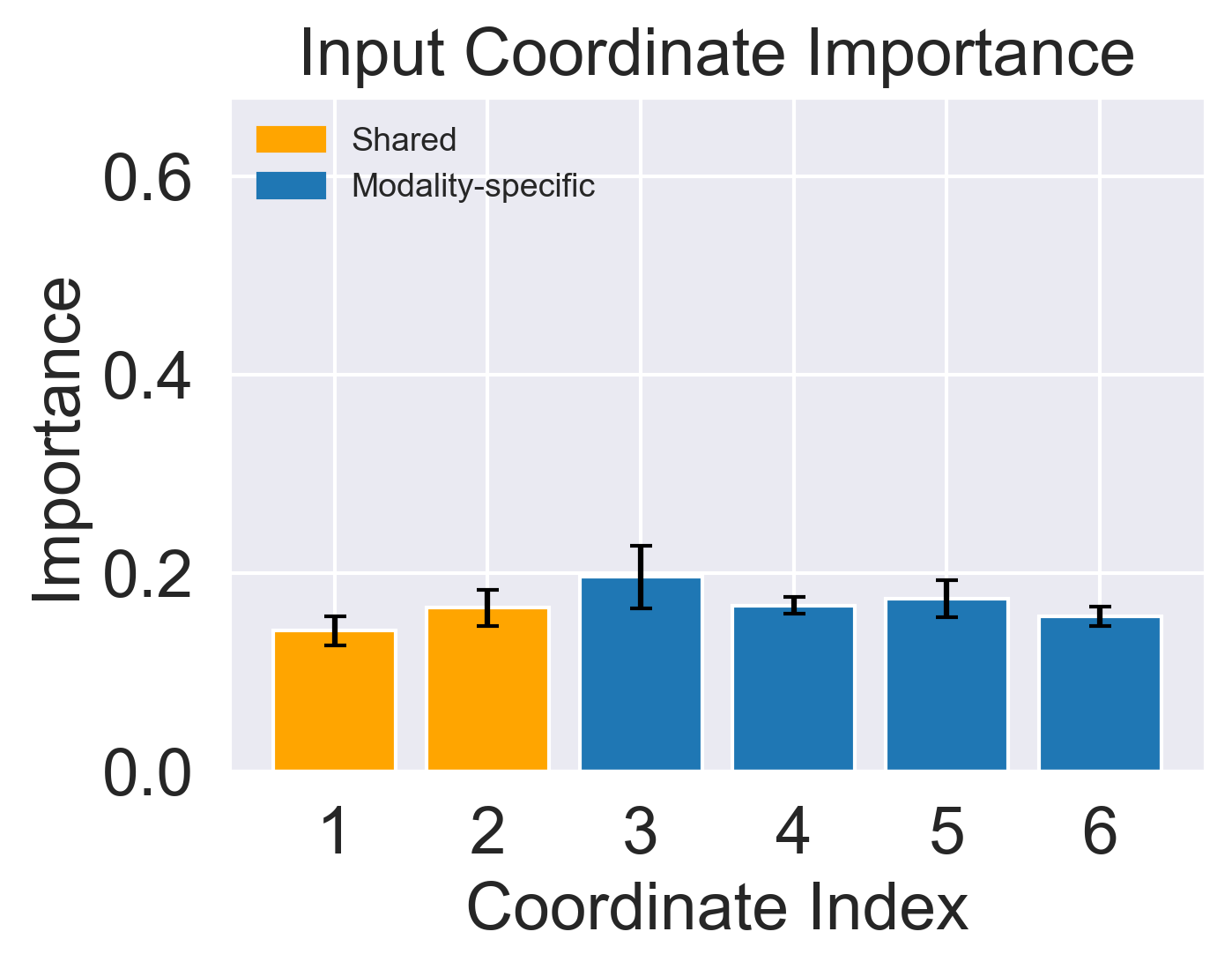}
        \caption{$\lambda=0.1$.}
        \label{fig:fact12}
    \end{subfigure}
    \hfill
    \begin{subfigure}[t]{0.225\linewidth}
        \centering
        \includegraphics[width=\linewidth]{figs/import_fact_setting1_10000_2_6_50_deep_1.0.png}
        \caption{$\lambda=1.0$.}
        \label{fig:fact13}
    \end{subfigure}
    \hfill
    \begin{subfigure}[t]{0.225\linewidth}
        \centering
        \includegraphics[width=\linewidth]{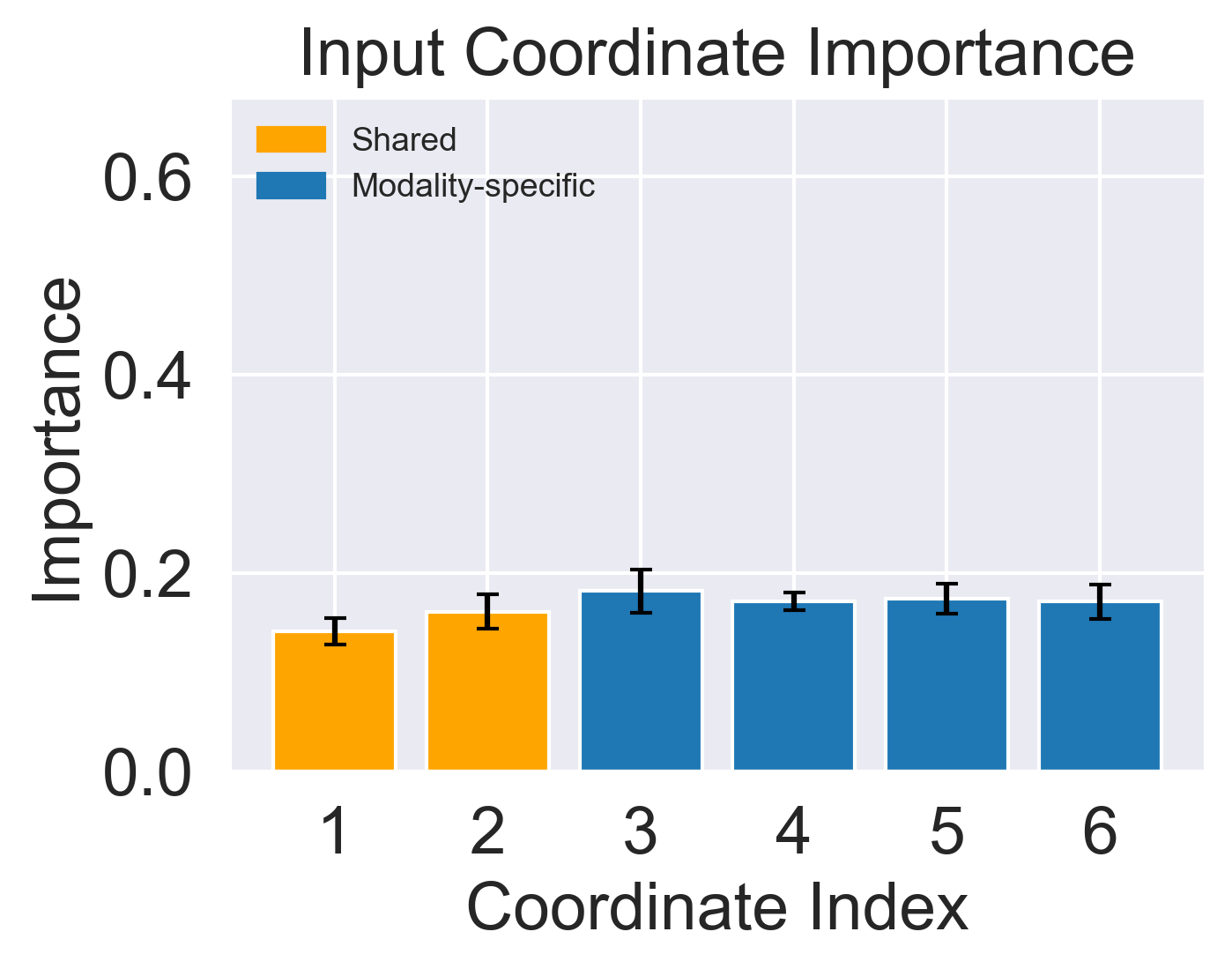}
        \caption{$\lambda=10.0$.}
        \label{fig:fact14}
    \end{subfigure}
    \caption{\yg{Factorized CL with varying $\lambda$: feature importance averaged over $50$ simulations (Setting 1).}}
    \label{fig:setting1-fact}
\end{figure}

\begin{figure}[!h]
    \centering
    \begin{subfigure}[t]{0.225\linewidth}
        \centering
        \includegraphics[width=\linewidth]{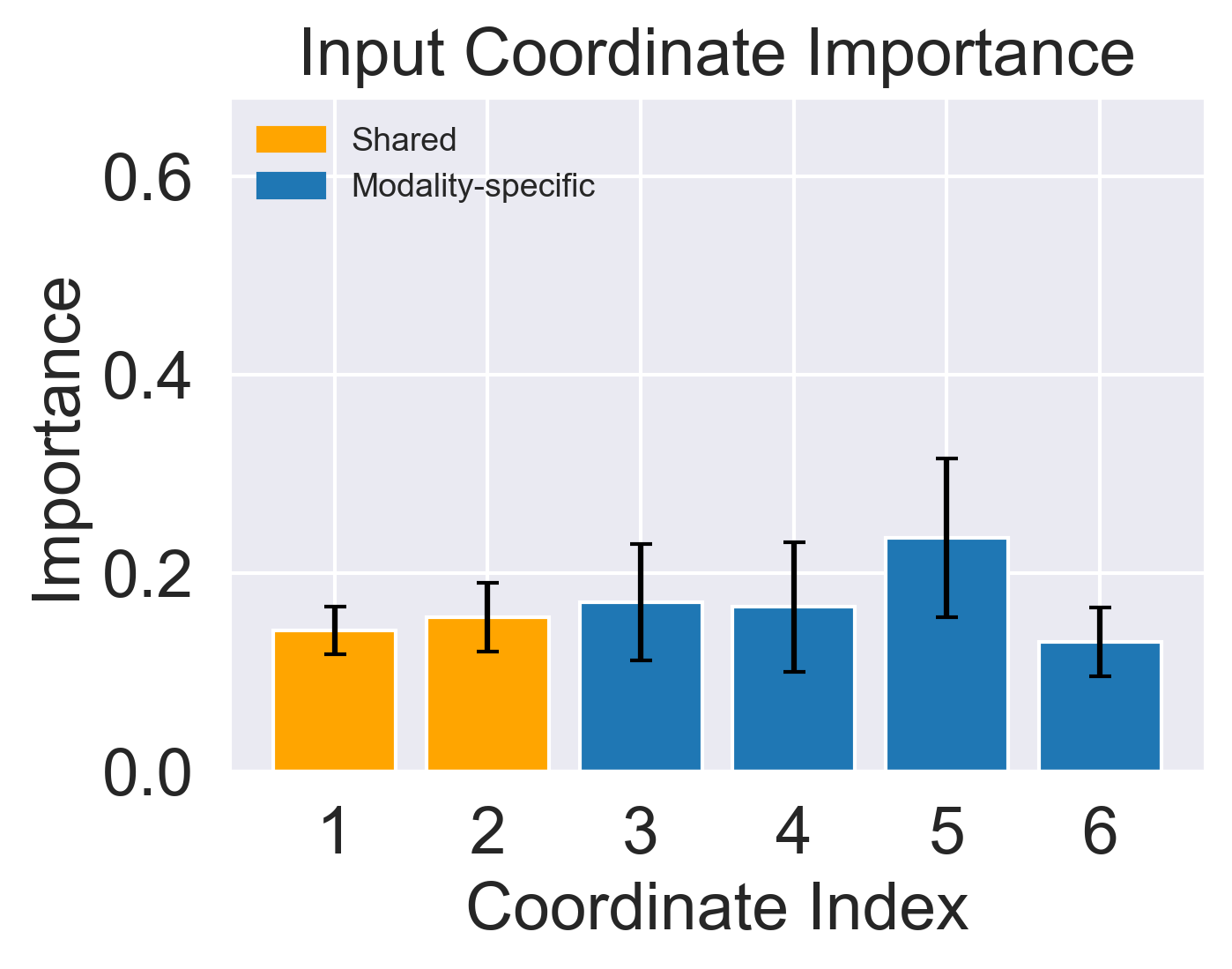}
        \caption{$\lambda=0.01$.}
        \label{fig:disen11}
    \end{subfigure}
    \hfill
    \begin{subfigure}[t]{0.225\linewidth}
        \centering
        \includegraphics[width=\linewidth]{figs/import_disen_setting1_10000_2_6_50_deep_0.1.png}
        \caption{$\lambda=0.1$.}
        \label{fig:disen12}
    \end{subfigure}
    \hfill
    \begin{subfigure}[t]{0.225\linewidth}
        \centering
        \includegraphics[width=\linewidth]{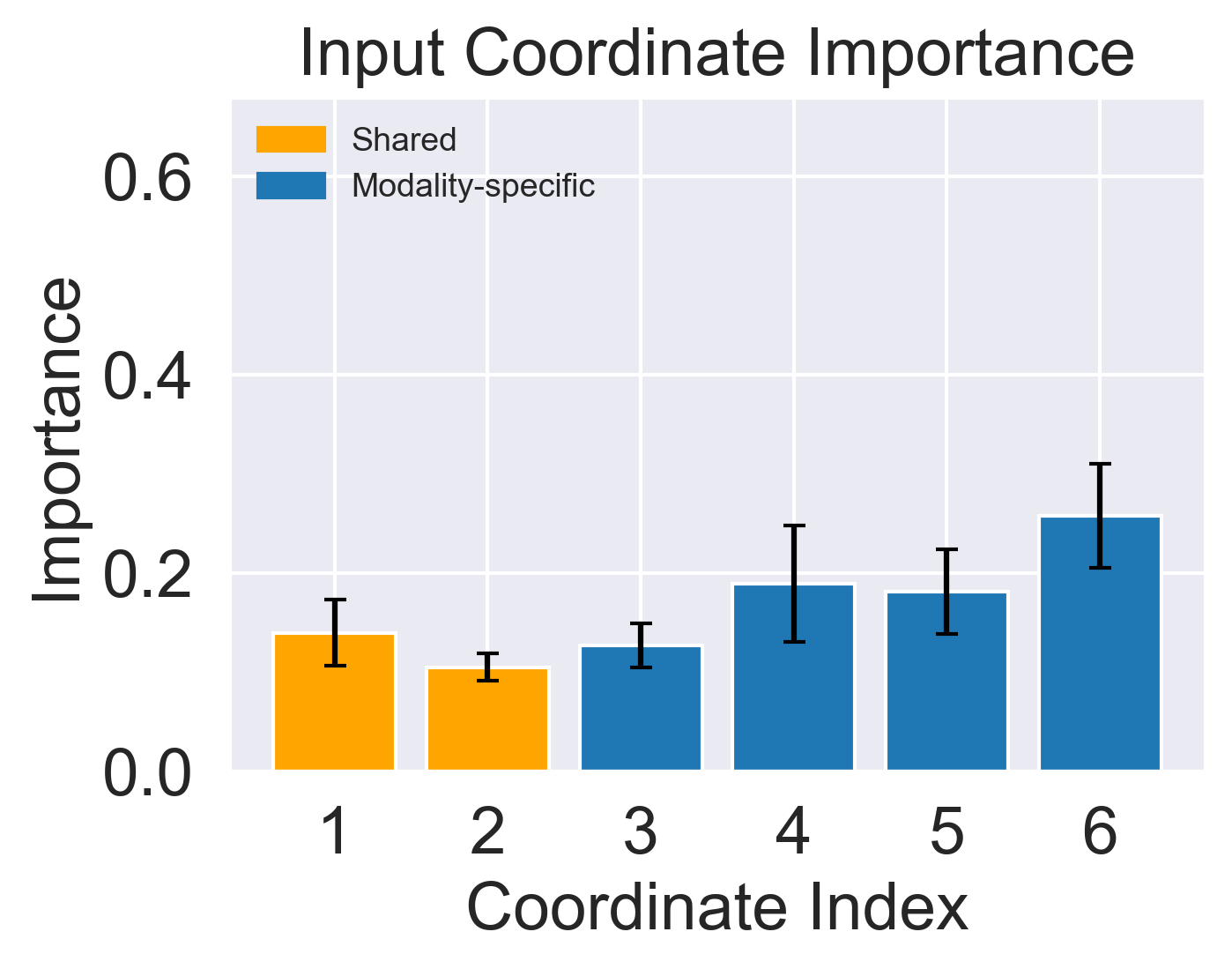}
        \caption{$\lambda=1.0$.}
        \label{fig:disen13}
    \end{subfigure}
    \hfill
    \begin{subfigure}[t]{0.225\linewidth}
        \centering
        \includegraphics[width=\linewidth]{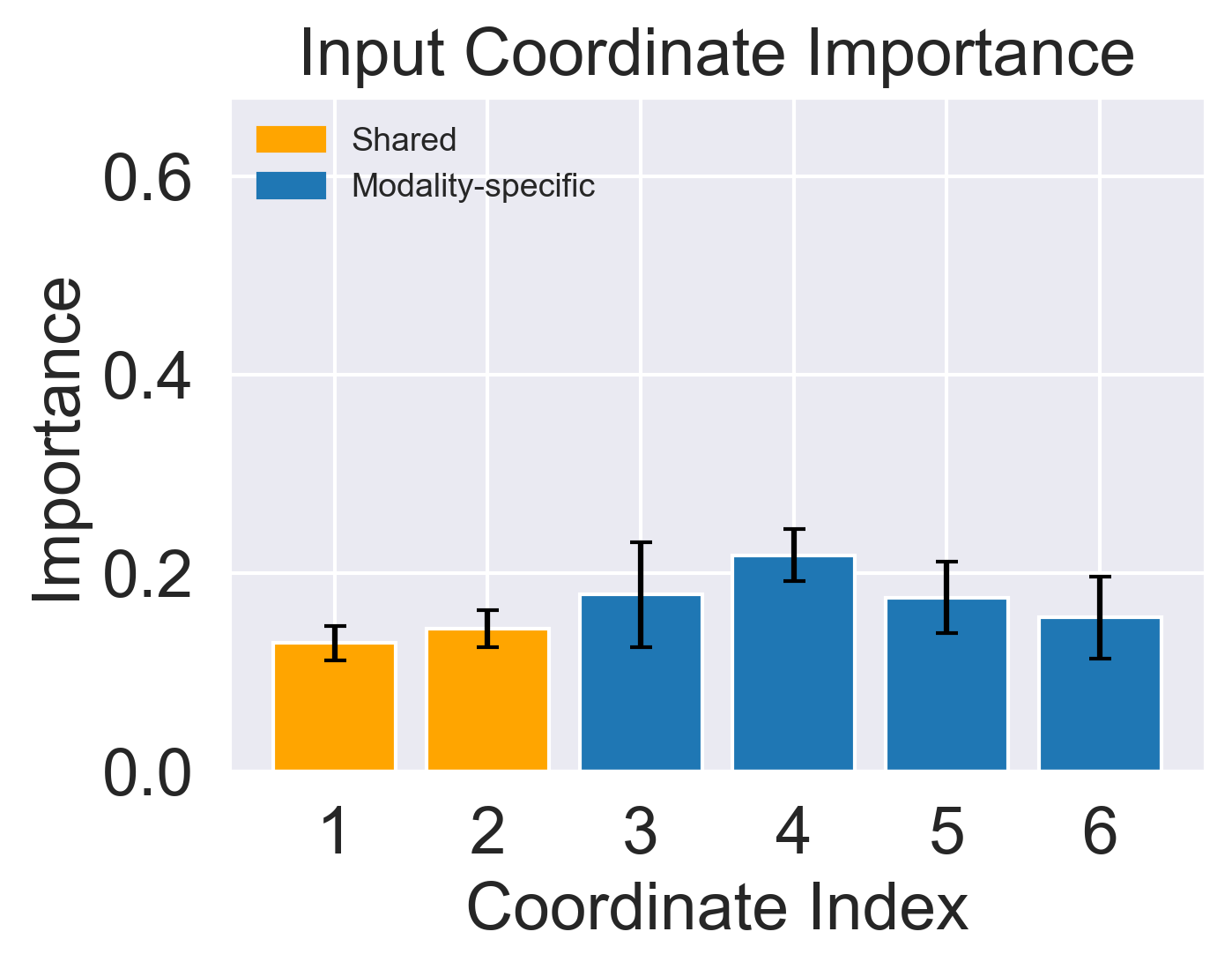}
        \caption{$\lambda=10.0$.}
        \label{fig:disen14}
    \end{subfigure}
    \caption{\yg{\disen~ with varying $\lambda$: feature importance averaged over $50$ simulations (Setting 1).}}
    \label{fig:setting1-disen}
\end{figure}

For Setting 1, given shared features are generated by sine and cubic functions, we can see from Figure~\ref{fig:setting1-recons} and \ref{fig:setting1-fact} that both Factorized CL and \ours~are capable of extracting modality-specific features that are independent of shared ones when $\lambda=0.01$, while \ours~tend to be more robust to the choice of $\lambda$ and preserves the desired performance when $\lambda=0.1$. However,  \disen, due to the limitation of orthogonal loss in handling nonlinear dependence, fails to disentangle modality-specific features from shared ones regardless of the value of $\lambda$.

\begin{figure}[ht]
    \centering
    \begin{subfigure}[t]{0.225\linewidth}
        \centering
        \includegraphics[width=\linewidth]{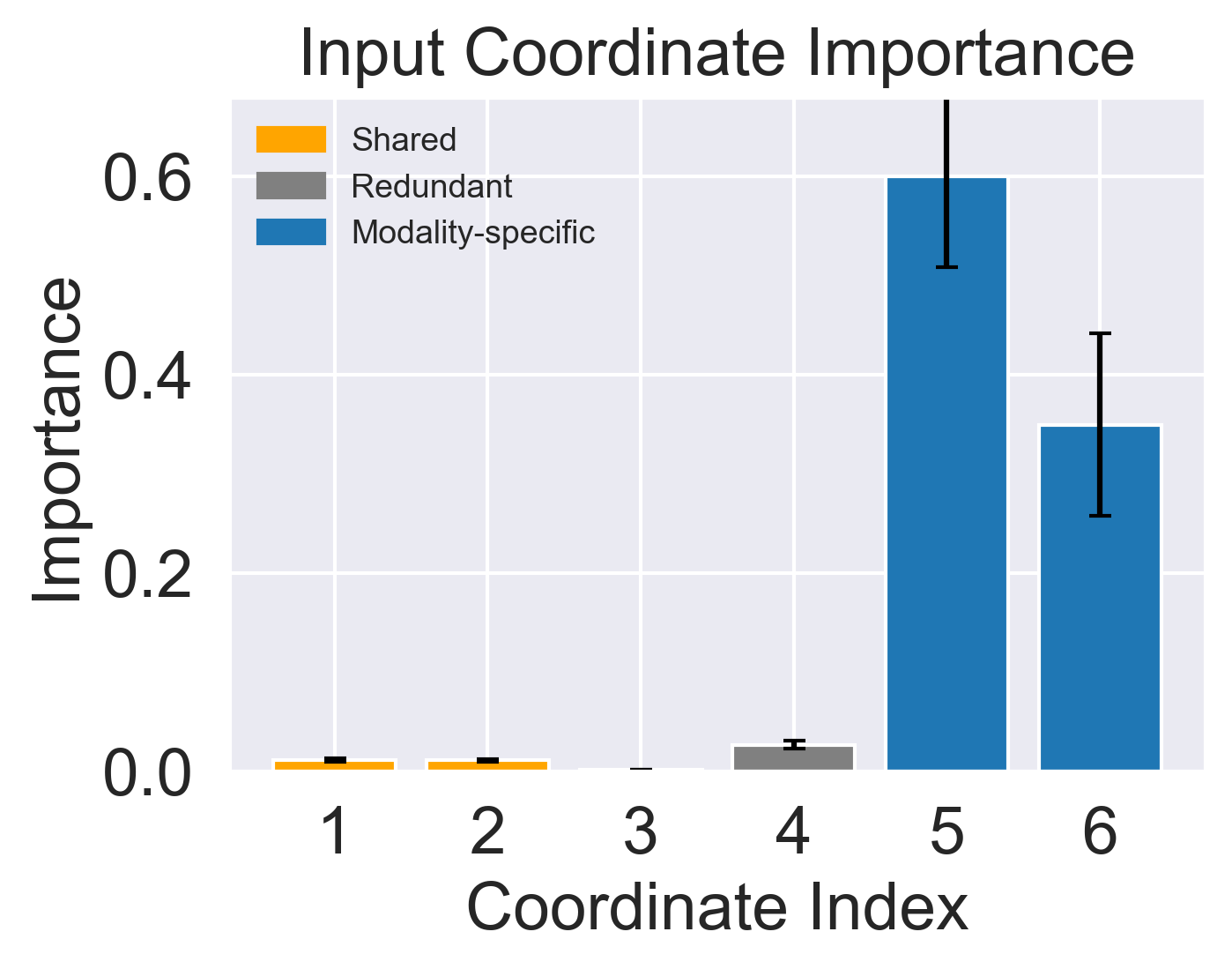}
        \caption{$\lambda=0.01$.}
        \label{fig:rec21}
    \end{subfigure}
    \hfill
    \begin{subfigure}[t]{0.225\linewidth}
        \centering
        \includegraphics[width=\linewidth]{figs/import_recons_setting4_10000_2_6_50_deep_0.1.png}
        \caption{$\lambda=0.1$.}
        \label{fig:rec22}
    \end{subfigure}
    \hfill
    \begin{subfigure}[t]{0.225\linewidth}
        \centering
        \includegraphics[width=\linewidth]{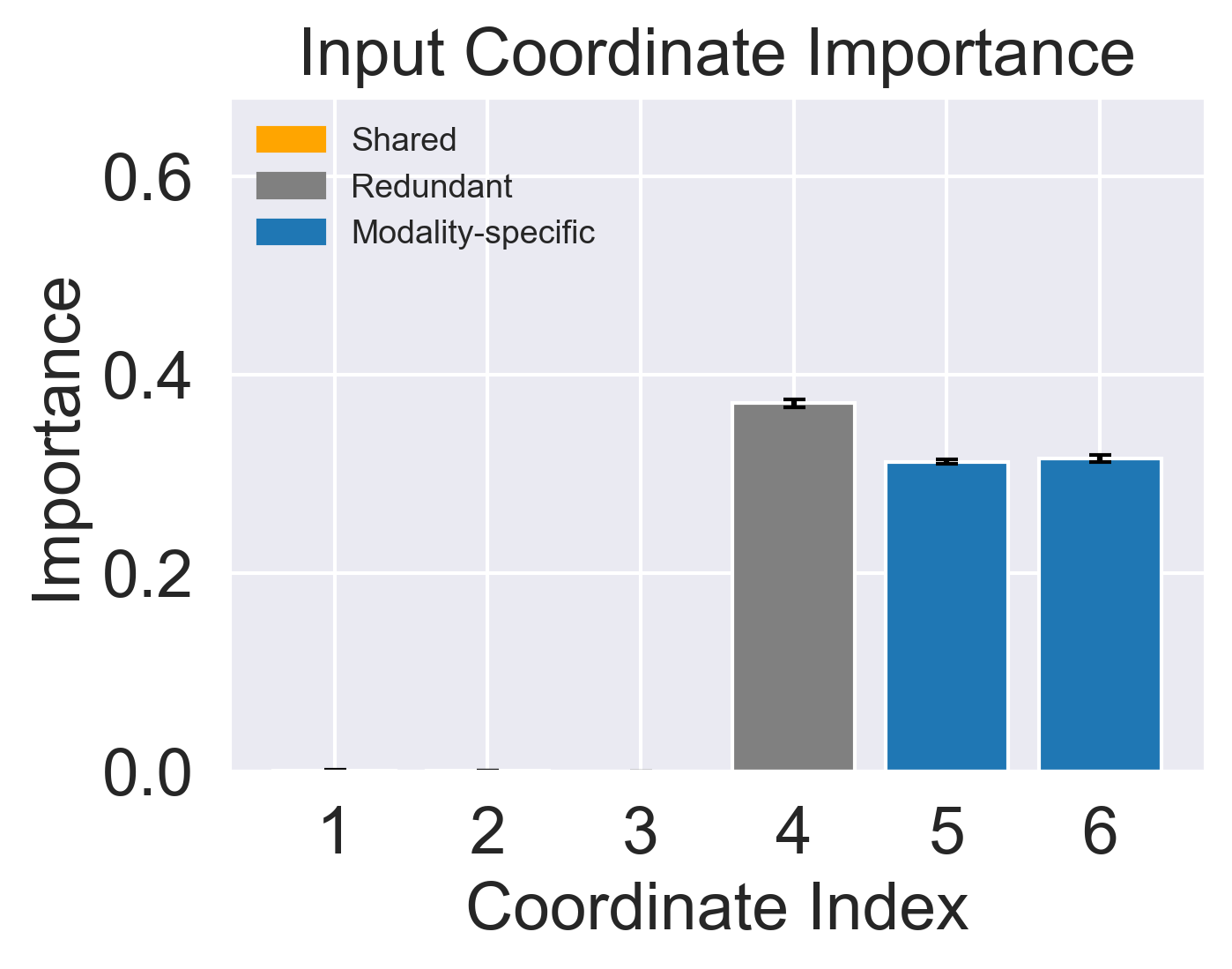}
        \caption{$\lambda=1.0$.}
        \label{fig:rec23}
    \end{subfigure}
    \hfill
    \begin{subfigure}[t]{0.225\linewidth}
        \centering
        \includegraphics[width=\linewidth]{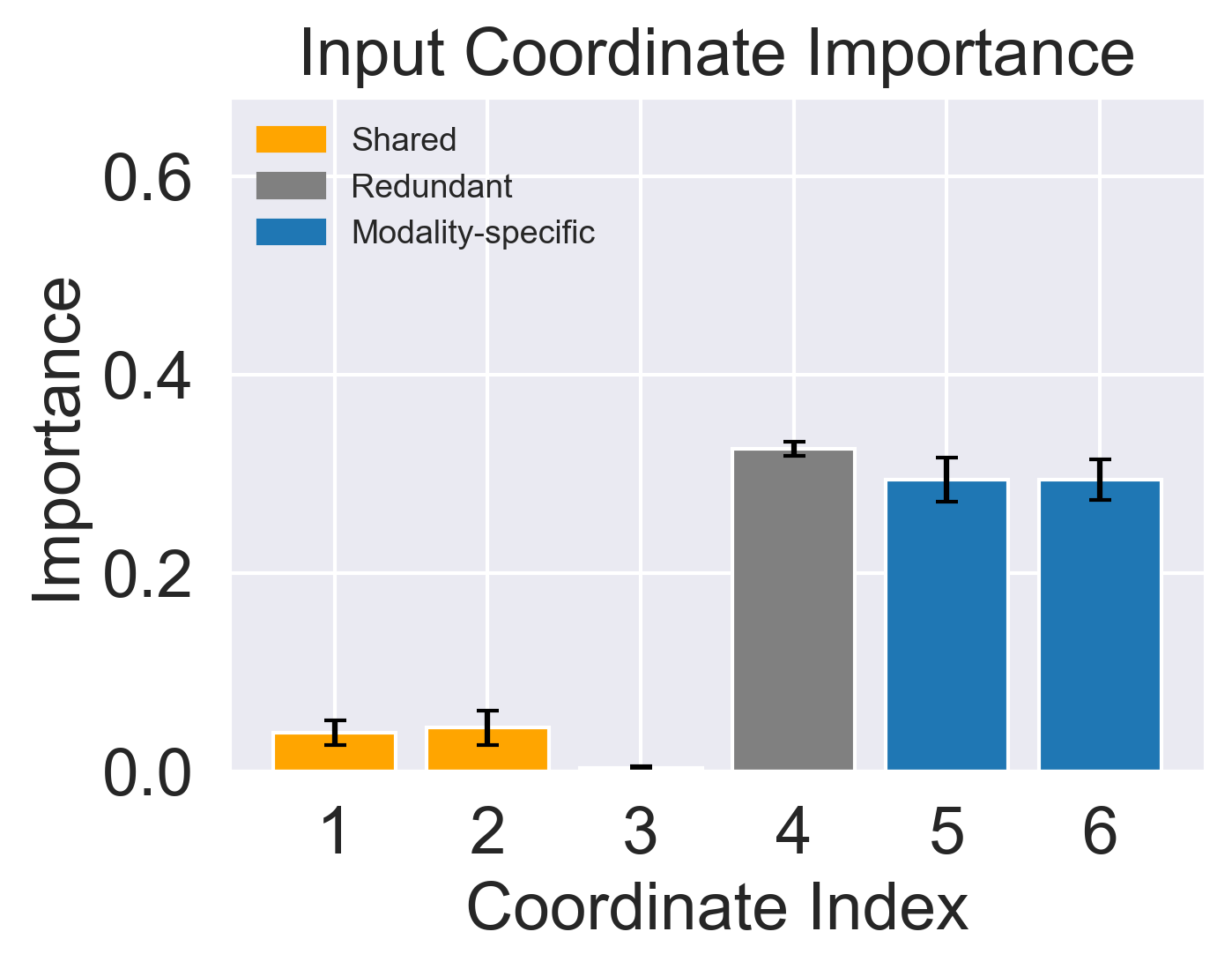}
        \caption{$\lambda=10.0$.}
        \label{fig:rec24}
    \end{subfigure}
    \caption{\yg{\ours~ with varying $\lambda$: feature importance averaged over $50$ simulations (Setting 2).}}
    \label{fig:setting4-recons}
\end{figure}

\begin{figure}[ht]
    \centering
    \begin{subfigure}[t]{0.225\linewidth}
        \centering
        \includegraphics[width=\linewidth]{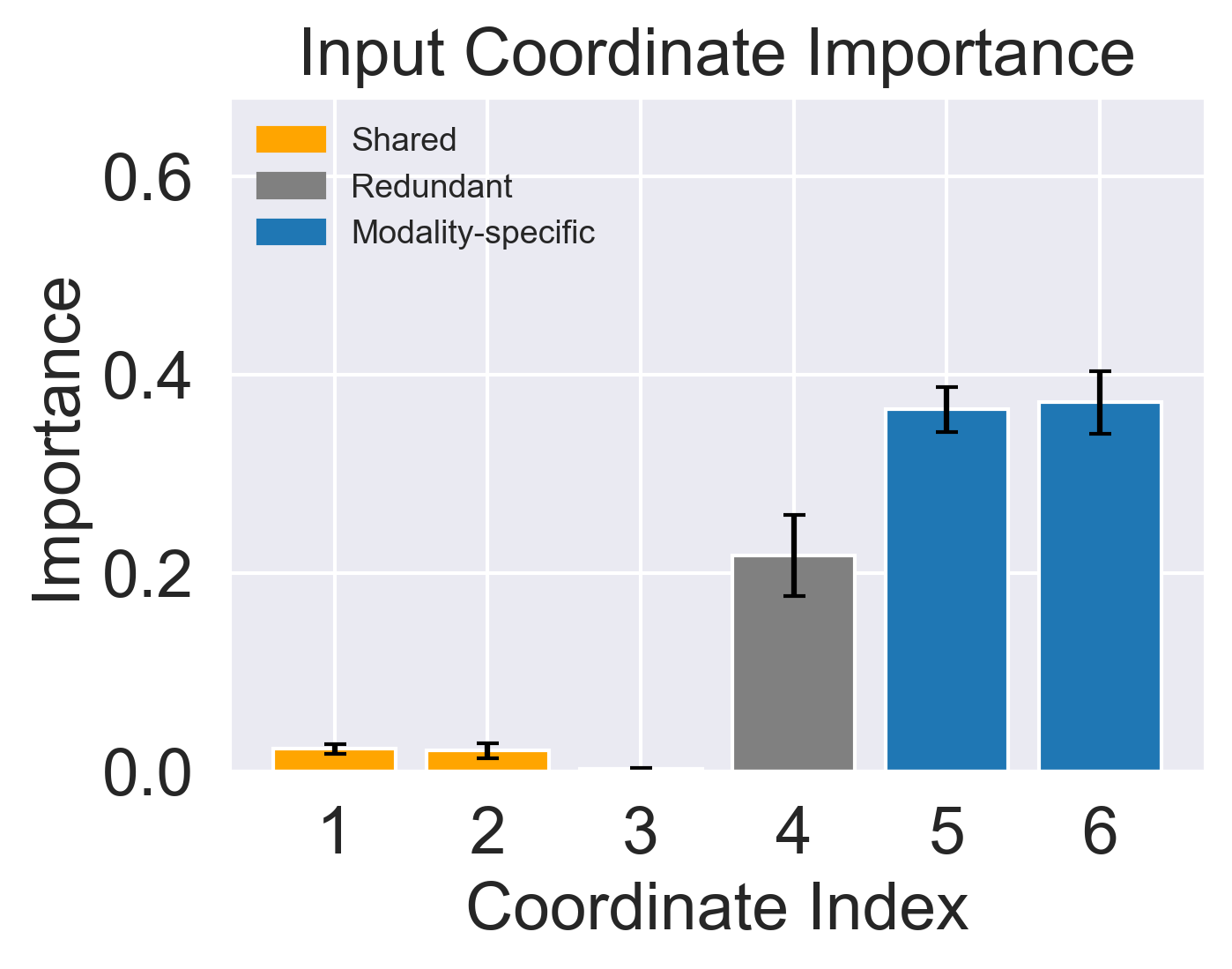}
        \caption{$\lambda=0.01$.}
        \label{fig:fact21}
    \end{subfigure}
    \hfill
    \begin{subfigure}[t]{0.225\linewidth}
        \centering
        \includegraphics[width=\linewidth]{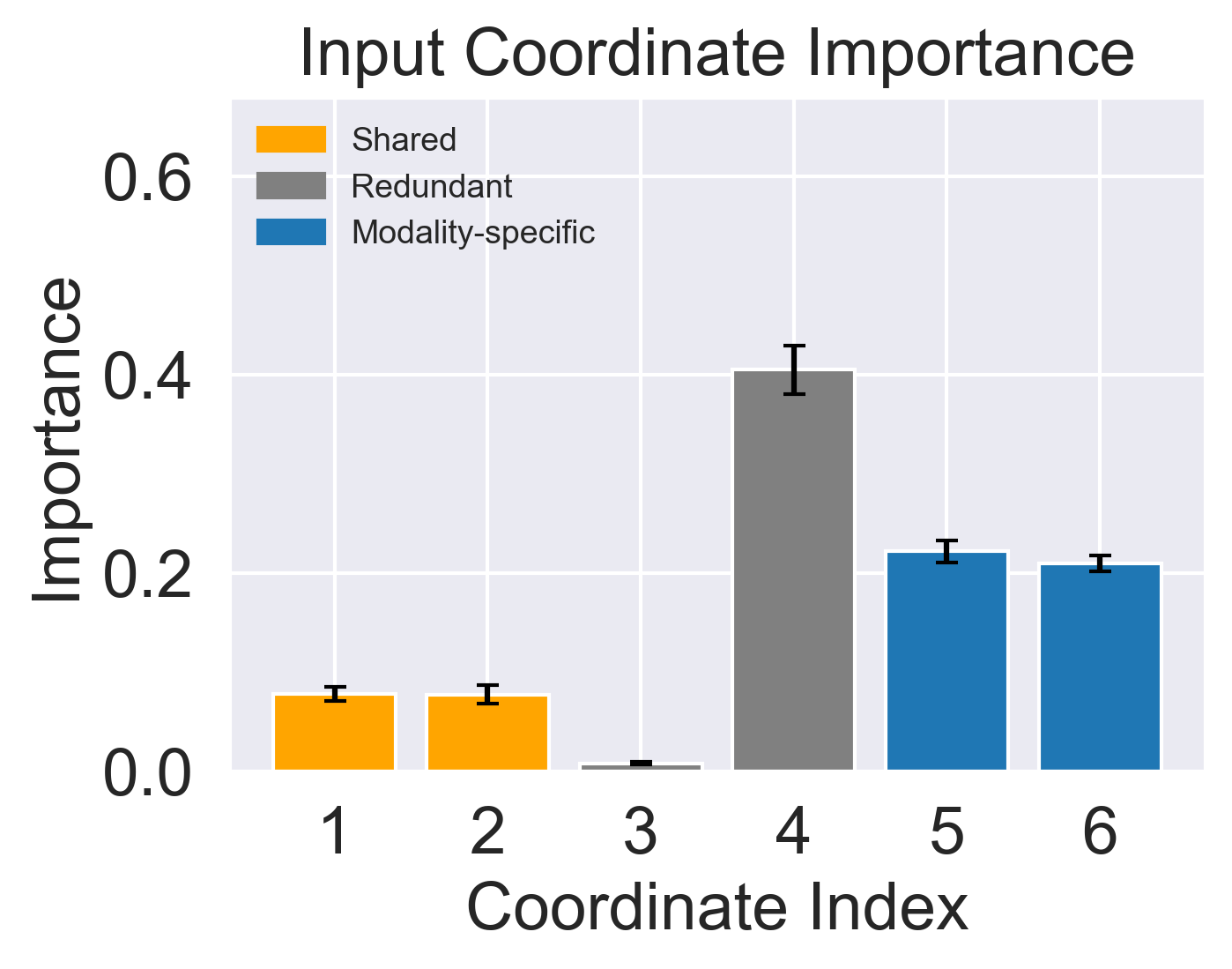}
        \caption{$\lambda=0.1$.}
        \label{fig:fact22}
    \end{subfigure}
    \hfill
    \begin{subfigure}[t]{0.225\linewidth}
        \centering
        \includegraphics[width=\linewidth]{figs/import_fact_setting4_10000_2_6_50_deep_1.0.png}
        \caption{$\lambda=1.0$.}
        \label{fig:fact23}
    \end{subfigure}
    \hfill
    \begin{subfigure}[t]{0.225\linewidth}
        \centering
        \includegraphics[width=\linewidth]{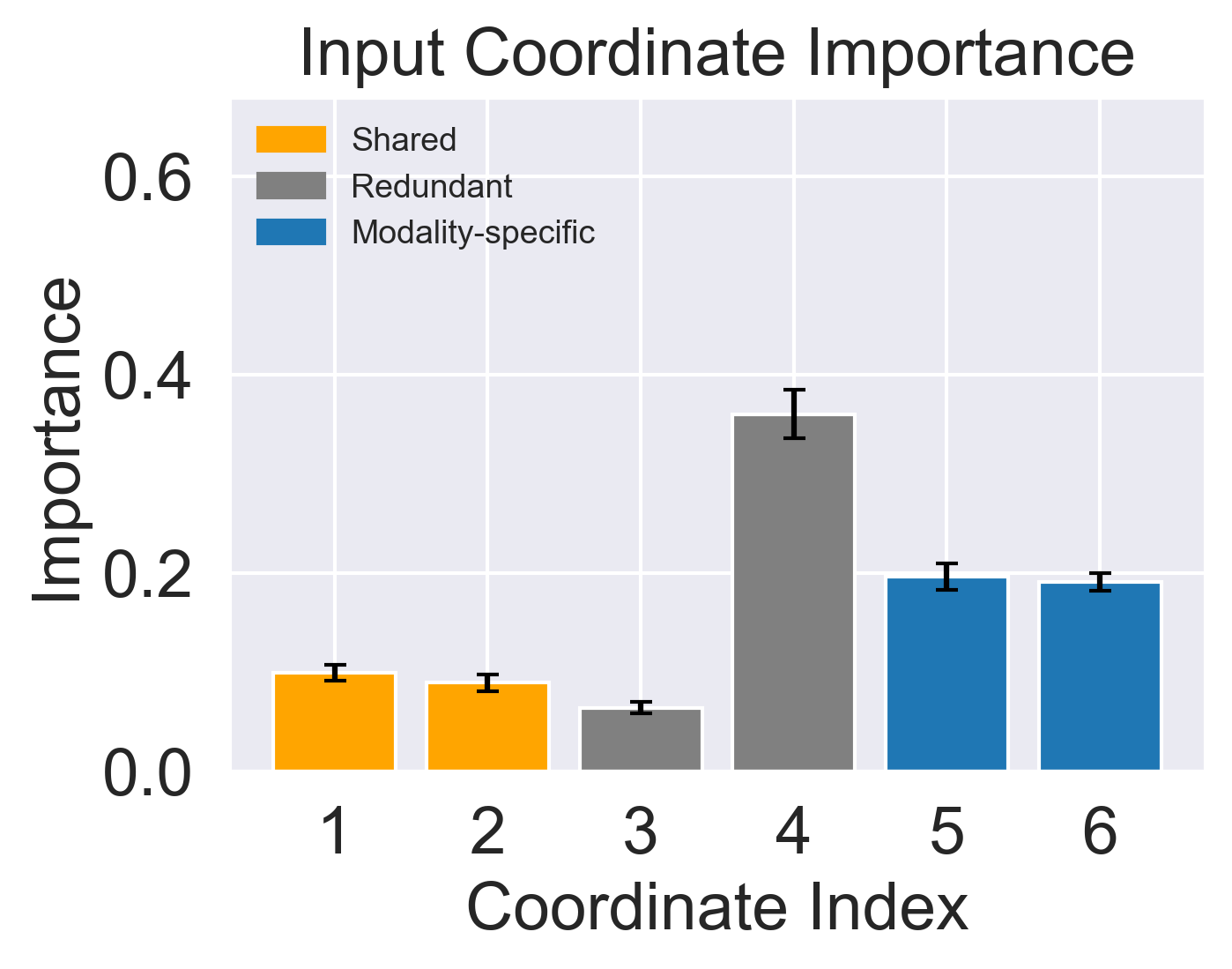}
        \caption{$\lambda=10.0$.}
        \label{fig:fact24}
    \end{subfigure}
    \caption{\yg{Factorized CL with varying $\lambda$: feature importance averaged over $50$ simulations (Setting 2).}}
    \label{fig:setting4-fact}
\end{figure}

\begin{figure}[!h]
    \centering
    \begin{subfigure}[t]{0.225\linewidth}
        \centering
        \includegraphics[width=\linewidth]{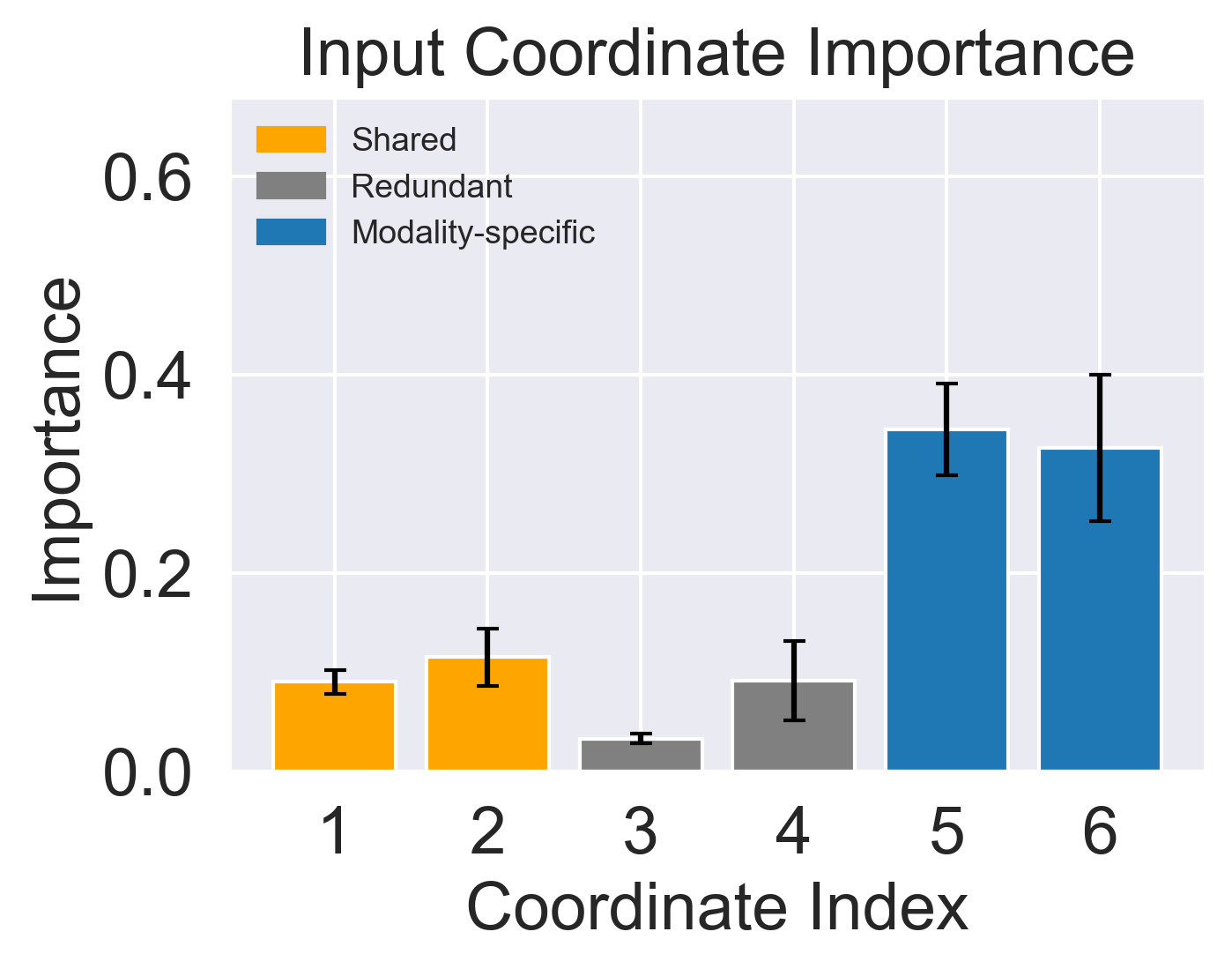}
        \caption{$\lambda=0.01$.}
        \label{fig:disen21}
    \end{subfigure}
    \hfill
    \begin{subfigure}[t]{0.225\linewidth}
        \centering
        \includegraphics[width=\linewidth]{figs/import_disen_setting4_10000_2_6_50_deep_0.1.png}
        \caption{$\lambda=0.1$.}
        \label{fig:disen22}
    \end{subfigure}
    \hfill
    \begin{subfigure}[t]{0.225\linewidth}
        \centering
        \includegraphics[width=\linewidth]{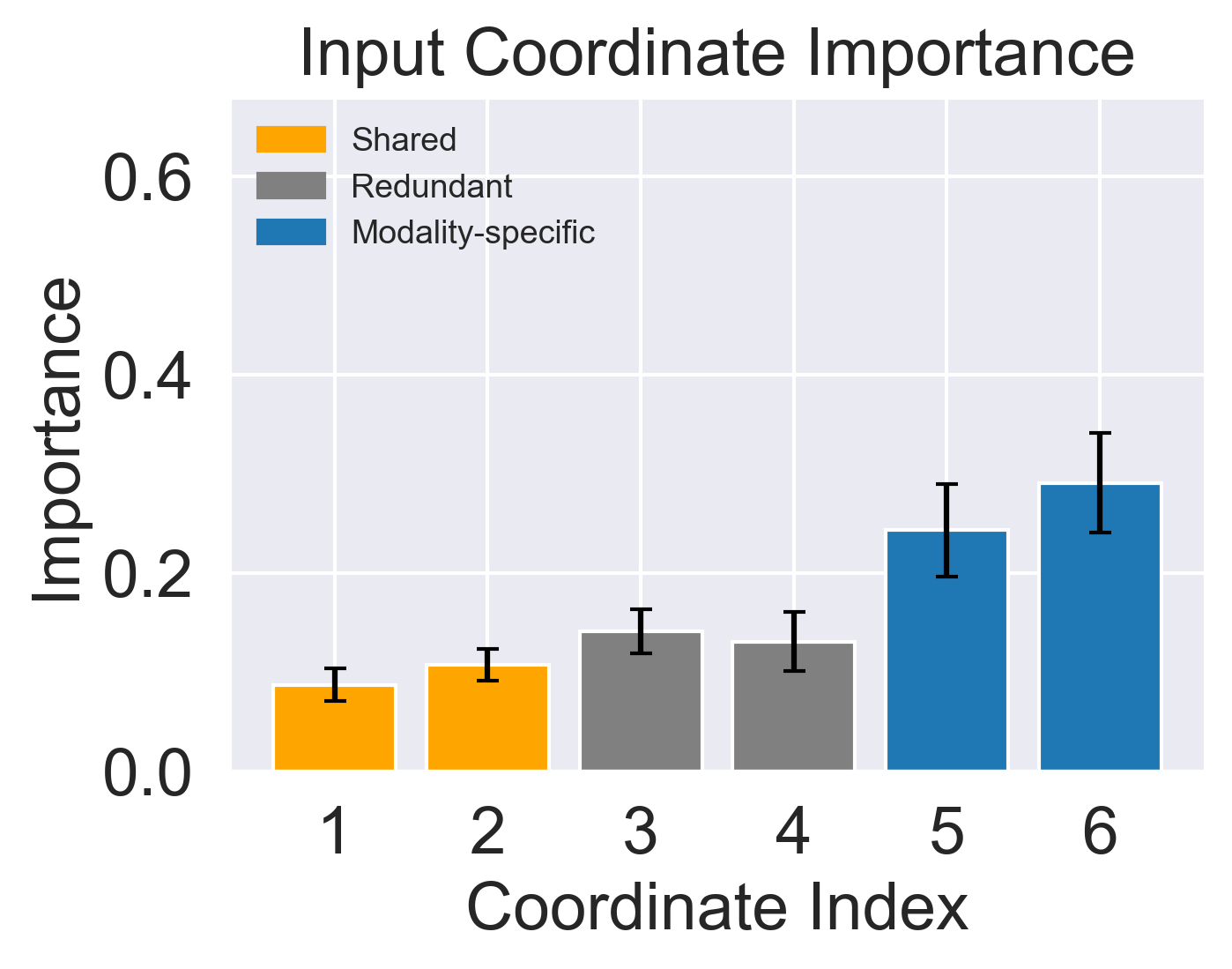}
        \caption{$\lambda=1.0$.}
        \label{fig:disen23}
    \end{subfigure}
    \hfill
    \begin{subfigure}[t]{0.225\linewidth}
        \centering
        \includegraphics[width=\linewidth]{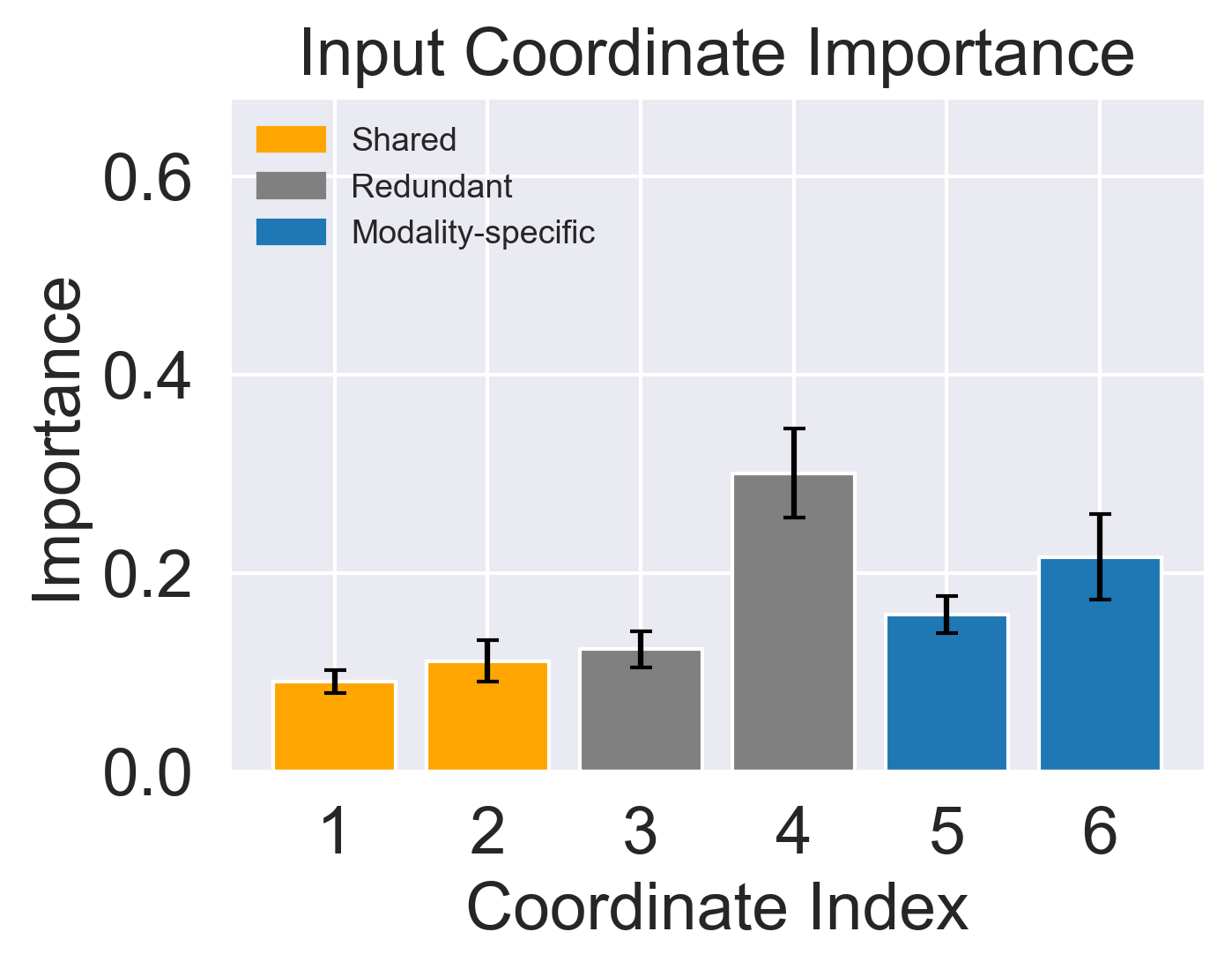}
        \caption{$\lambda=10.0$.}
        \label{fig:disen24}
    \end{subfigure}
    \caption{\yg{\disen~ with varying $\lambda$: feature importance averaged over $50$ simulations (Setting 2).}}
    \label{fig:setting4-disen}
\end{figure}

% \nb{this paragraph no longer describes the updated plots...}
\yg{
From Figures~\ref{fig:setting4-recons}, \ref{fig:setting4-fact}, and \ref{fig:setting4-disen}, we can see that similar to the results presented in Section~\ref{sec:simu}, with $\lambda$ in the range of $\{10^{j}: j = -2,-1,0,1\}$, {\disen} fails to enforce independence between shared and modality-specific features; FactorizedCL tends to exclude shared features with a small $\lambda$ but still captures at least one of redundant features that is correlated with shared ones. 
In comparison, with $\lambda=0.01$, \ours~ can exactly capture the remaining two modality-specific features.   
Moreover, these figures also reveal the tradeoff between the disentanglement and sufficiency, determined by the value of $\lambda$. 
With $\lambda=0.01$, more weight is put on the disentanglement between shared and modality-specific features; thus, with a strong regularization on dependence, \ours~ tends to exclude all shared and redundant features. On the other hand, with $\lambda=10$, the loss puts more weight on reconstruction/sufficiency, and as a result, disentangled methods may extract features that are dependent on the shared ones.
}

\subsubsection{Ablation studies beyond Gaussianity}

\begin{figure}[ht]
    \centering
    \begin{subfigure}[t]{0.225\linewidth}
        \centering
        \includegraphics[width=\linewidth]{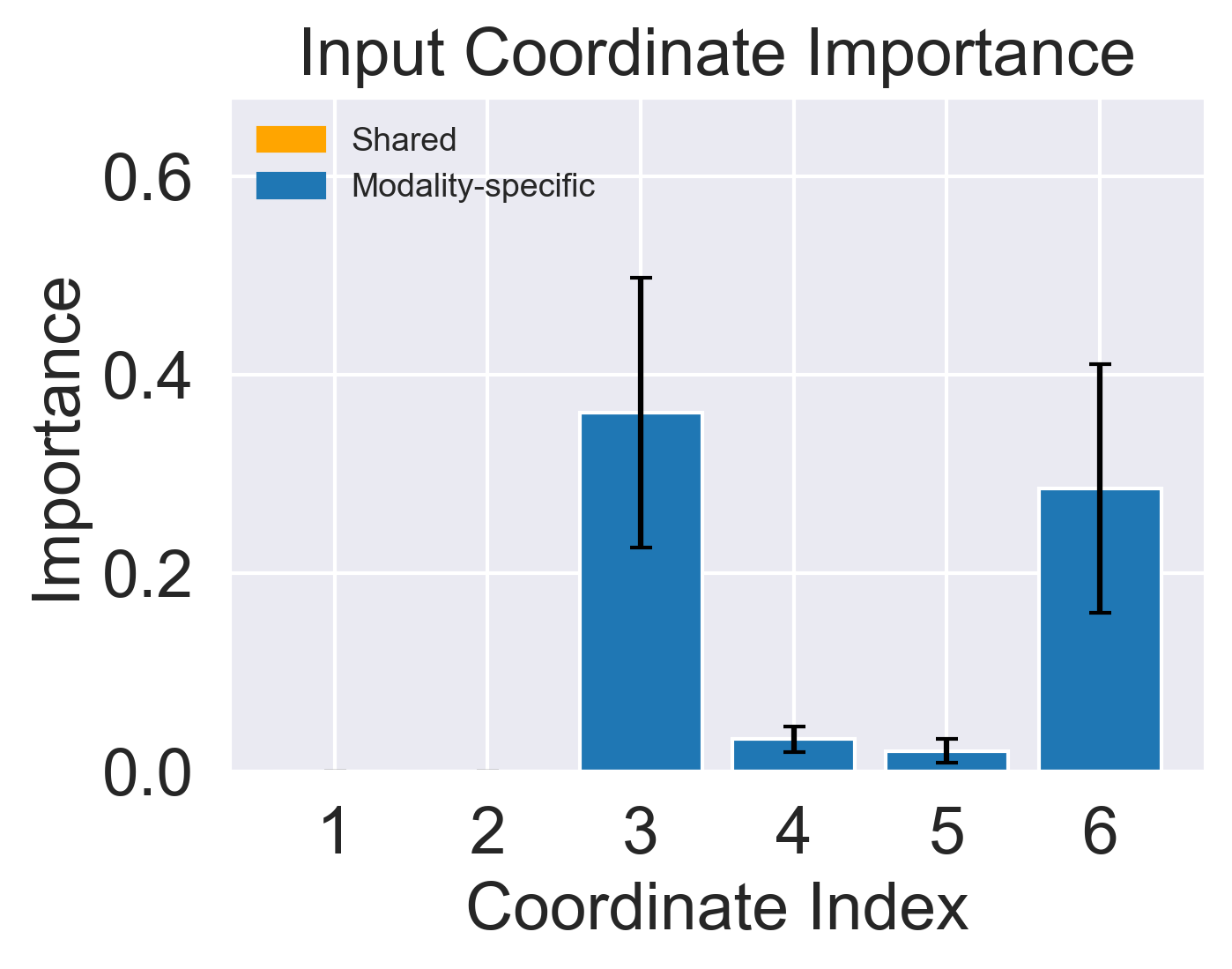}
        \caption{$\lambda=0.01$.}
        \label{fig:rec331}
    \end{subfigure}
    \hfill
    \begin{subfigure}[t]{0.225\linewidth}
        \centering
        \includegraphics[width=\linewidth]{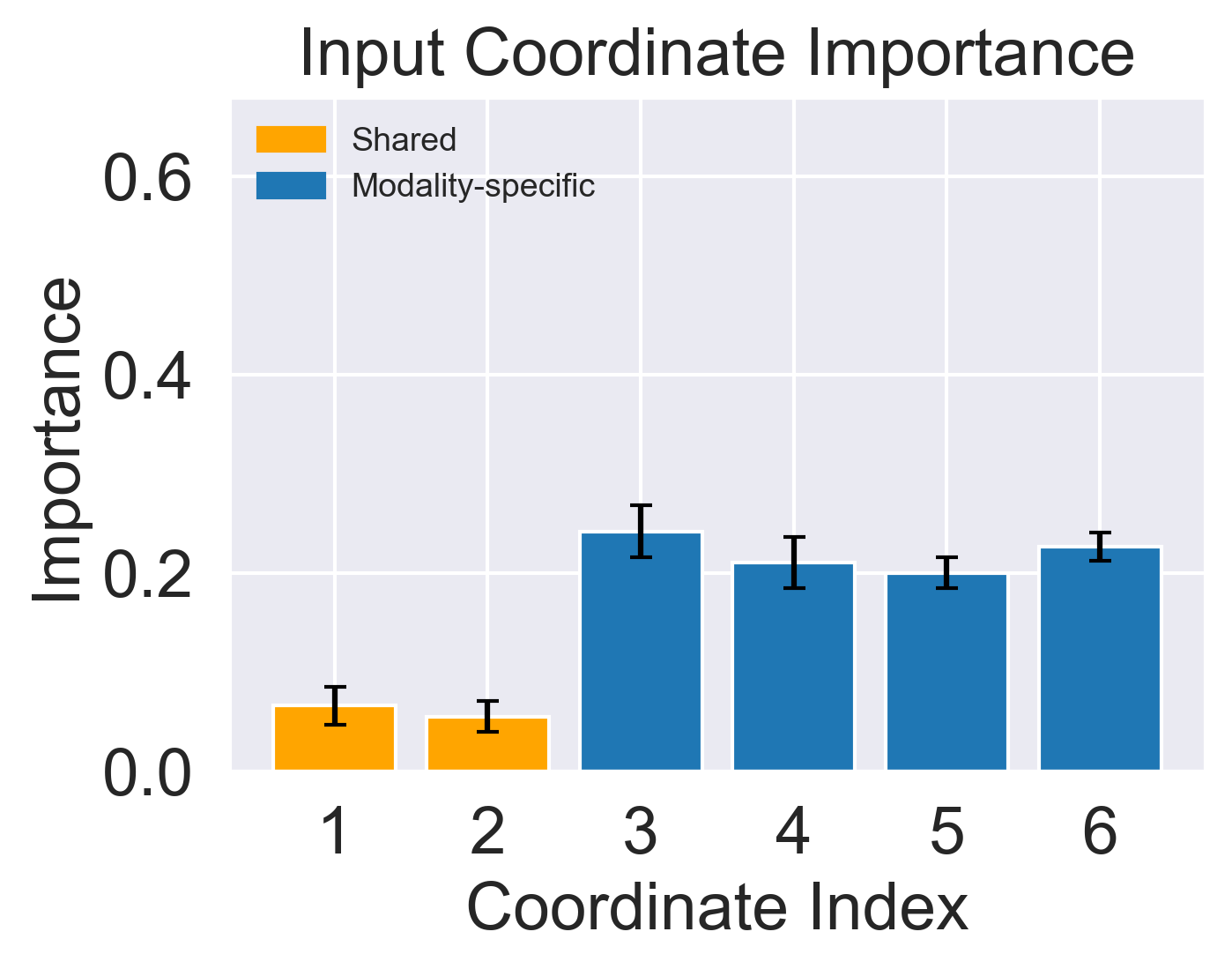}
        \caption{$\lambda=0.1$.}
        \label{fig:rec332}
    \end{subfigure}
    \hfill
    \begin{subfigure}[t]{0.225\linewidth}
        \centering
        \includegraphics[width=\linewidth]{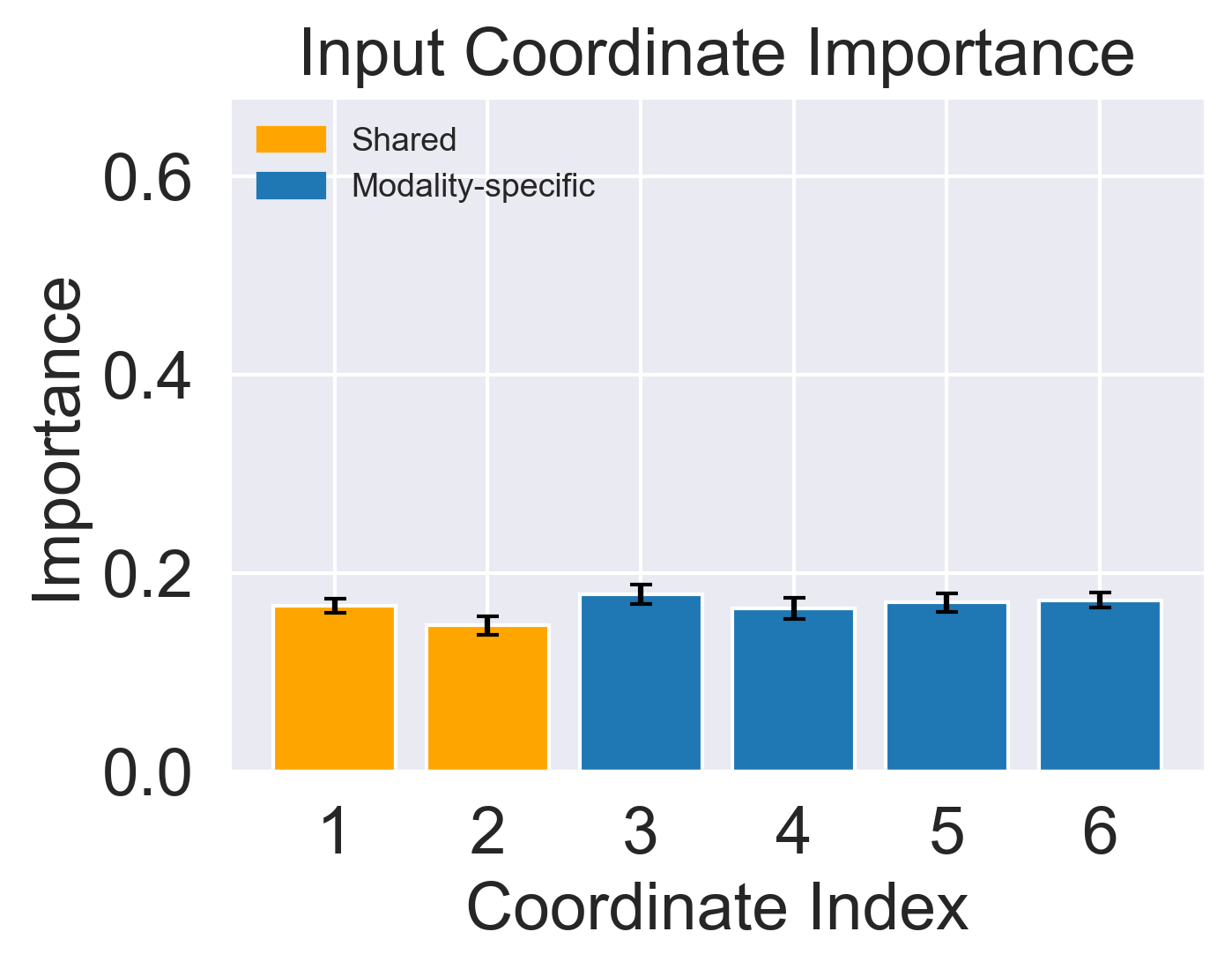}
        \caption{$\lambda=1.0$.}
        \label{fig:rec333}
    \end{subfigure}
    \hfill
    \begin{subfigure}[t]{0.225\linewidth}
        \centering
        \includegraphics[width=\linewidth]{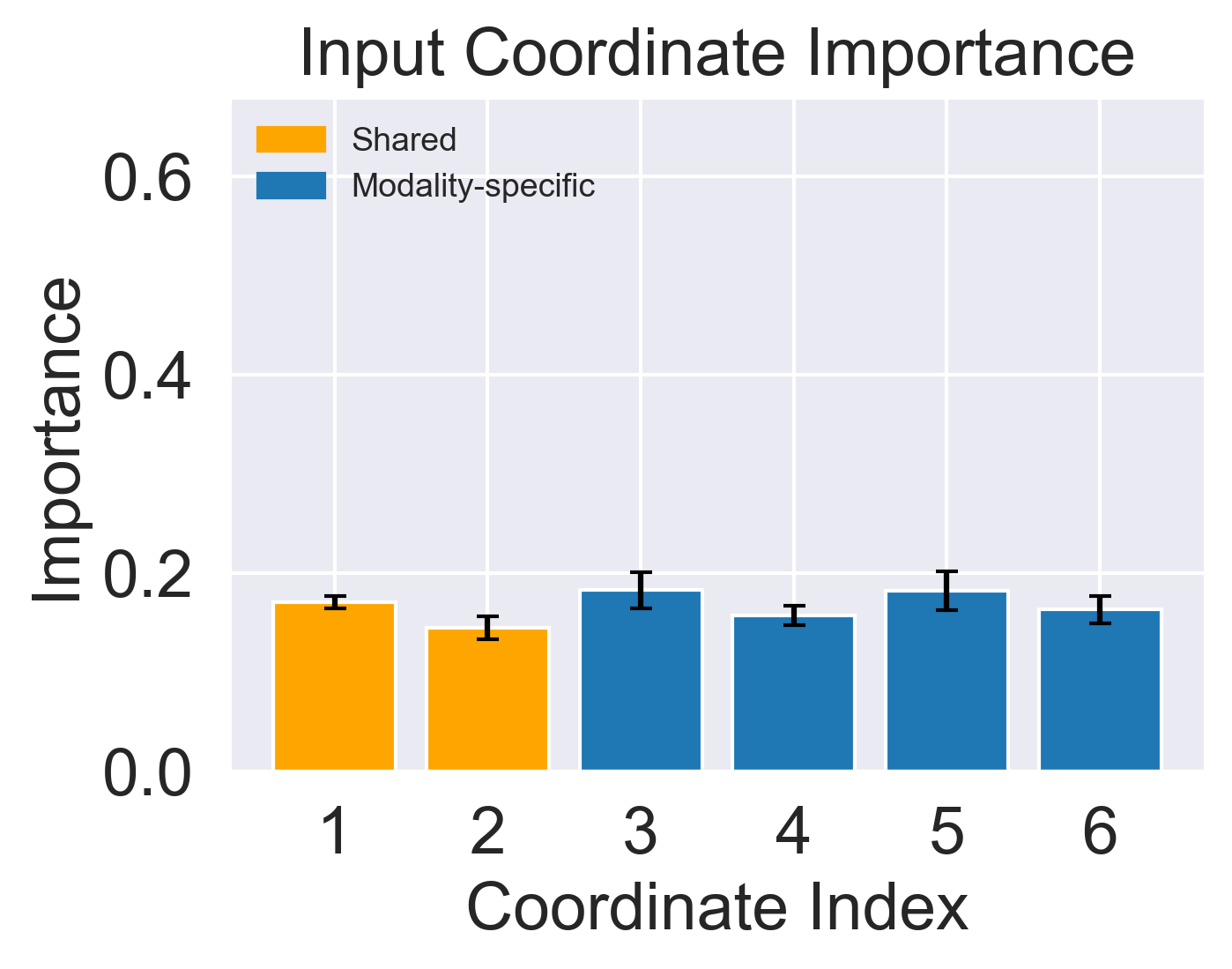}
        \caption{$\lambda=10.0$.}
        \label{fig:rec334}
    \end{subfigure}
    \caption{\yg{\ours~ with varying $\lambda$: feature importance averaged over $10$ simulations (Setting 3).}}
    \label{fig:setting6-recons}
\end{figure}

\begin{figure}[ht]
    \centering
    \begin{subfigure}[t]{0.225\linewidth}
        \centering
        \includegraphics[width=\linewidth]{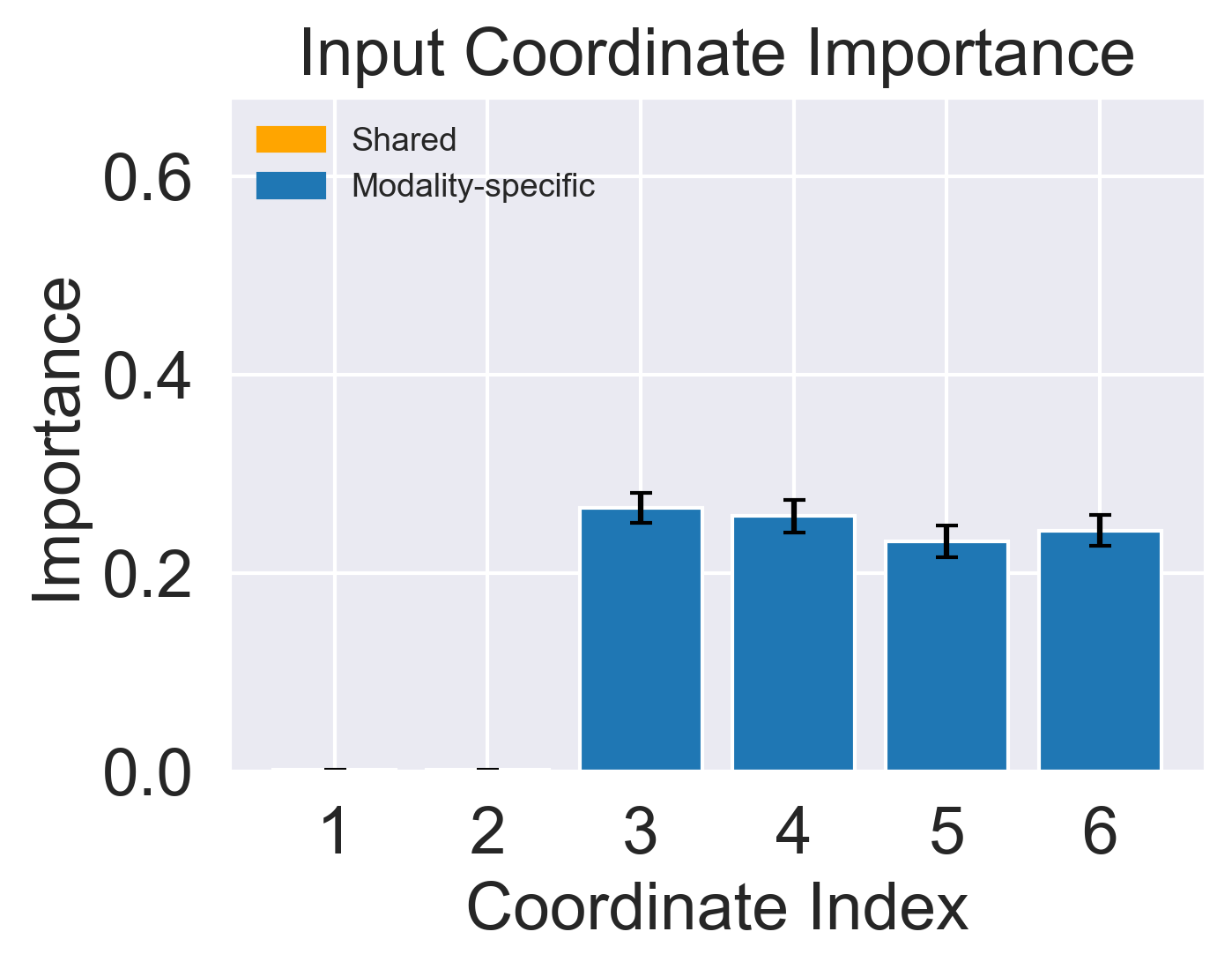}
        \caption{$\lambda=0.01$.}
        \label{fig:fact331}
    \end{subfigure}
    \hfill
    \begin{subfigure}[t]{0.225\linewidth}
        \centering
        \includegraphics[width=\linewidth]{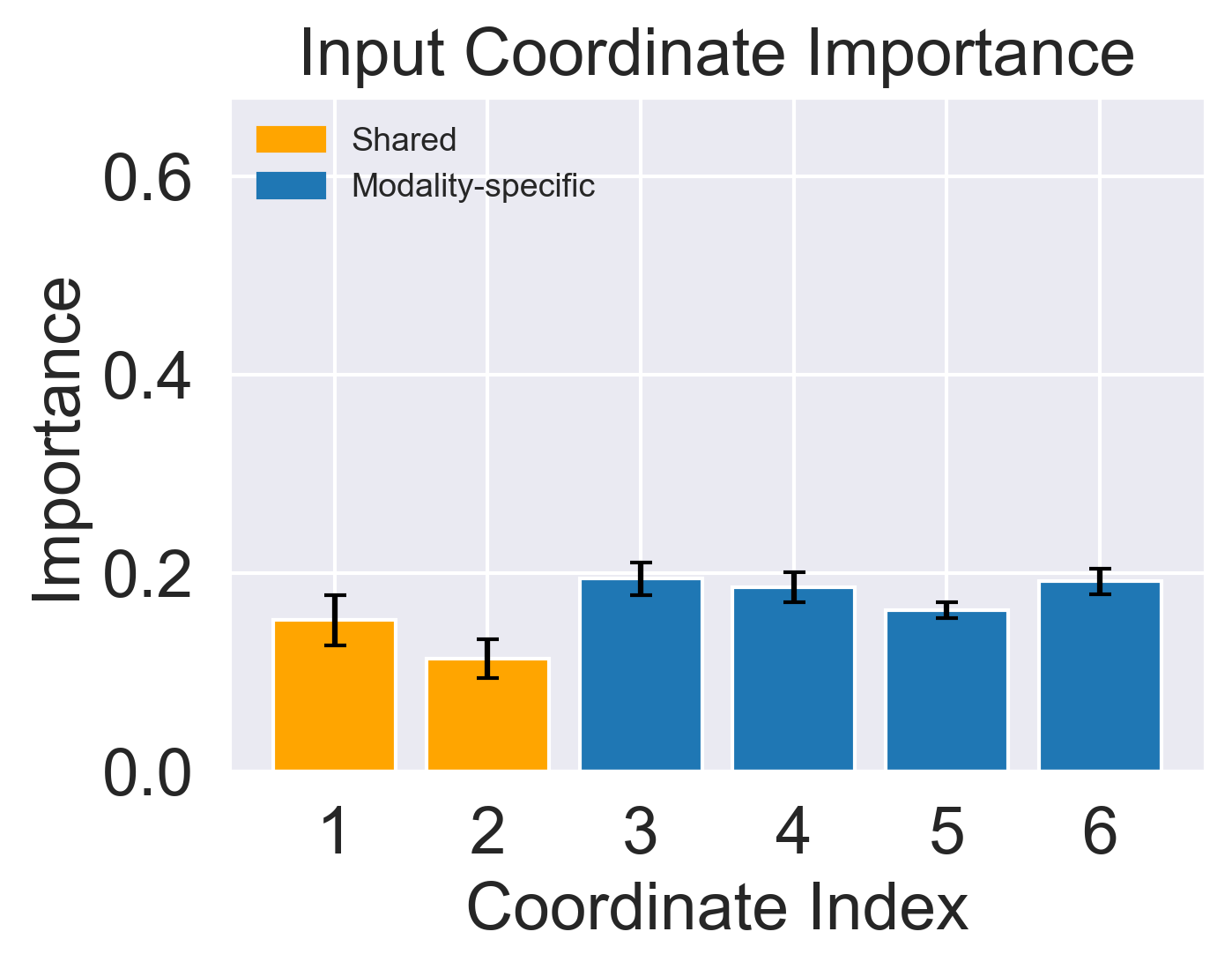}
        \caption{$\lambda=0.1$.}
        \label{fig:fact332}
    \end{subfigure}
    \hfill
    \begin{subfigure}[t]{0.225\linewidth}
        \centering
        \includegraphics[width=\linewidth]{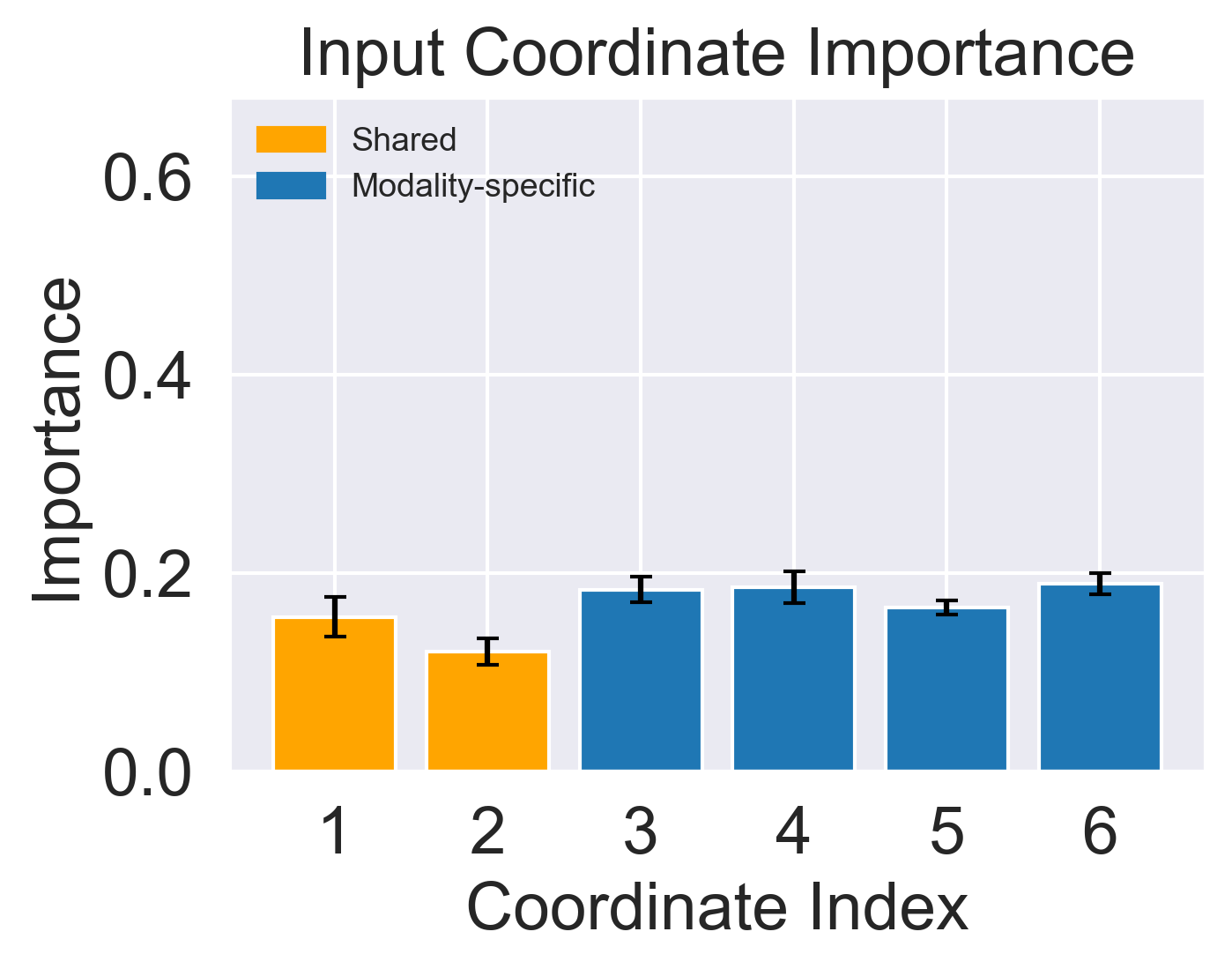}
        \caption{$\lambda=1.0$.}
        \label{fig:fact333}
    \end{subfigure}
    \hfill
    \begin{subfigure}[t]{0.225\linewidth}
        \centering
        \includegraphics[width=\linewidth]{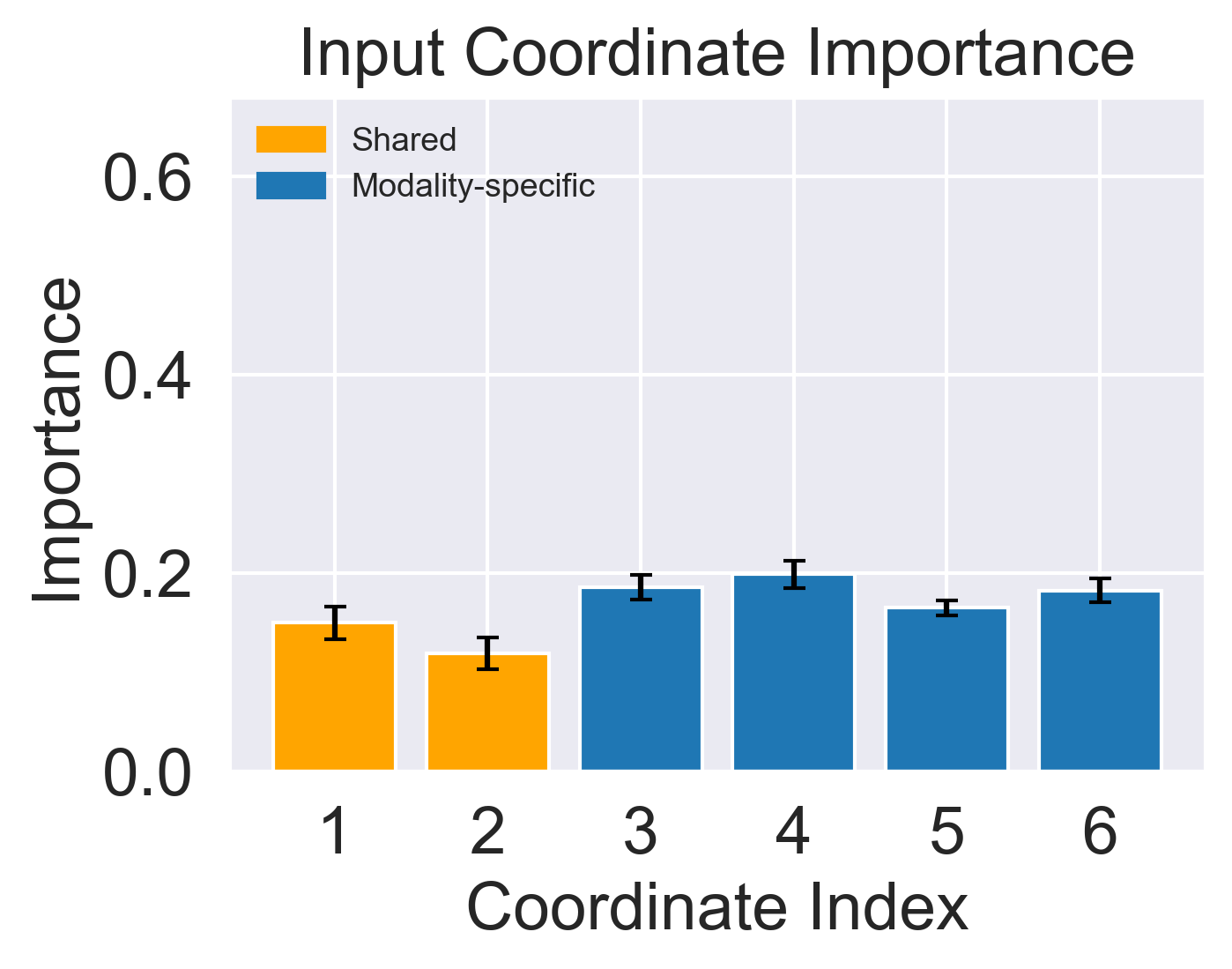}
        \caption{$\lambda=10.0$.}
        \label{fig:fact334}
    \end{subfigure}
    \caption{\yg{Factorized CL with varying $\lambda$: feature importance averaged over $10$ simulations (Setting 3).}}
    \label{fig:setting6-fact}
\end{figure}

\begin{figure}[!h]
    \centering
    \begin{subfigure}[t]{0.225\linewidth}
        \centering
        \includegraphics[width=\linewidth]{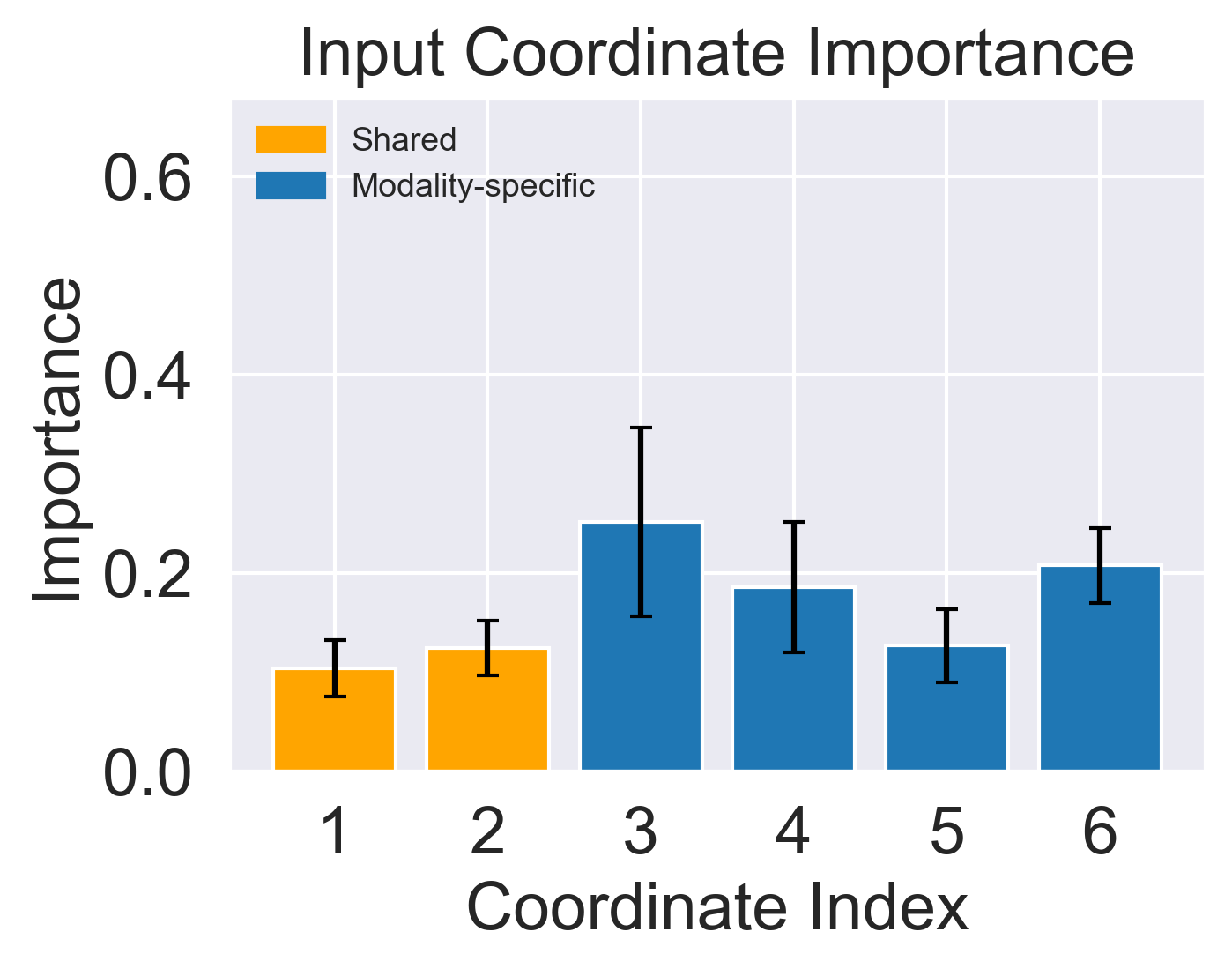}
        \caption{$\lambda=0.01$.}
        \label{fig:disen331}
    \end{subfigure}
    \hfill
    \begin{subfigure}[t]{0.225\linewidth}
        \centering
        \includegraphics[width=\linewidth]{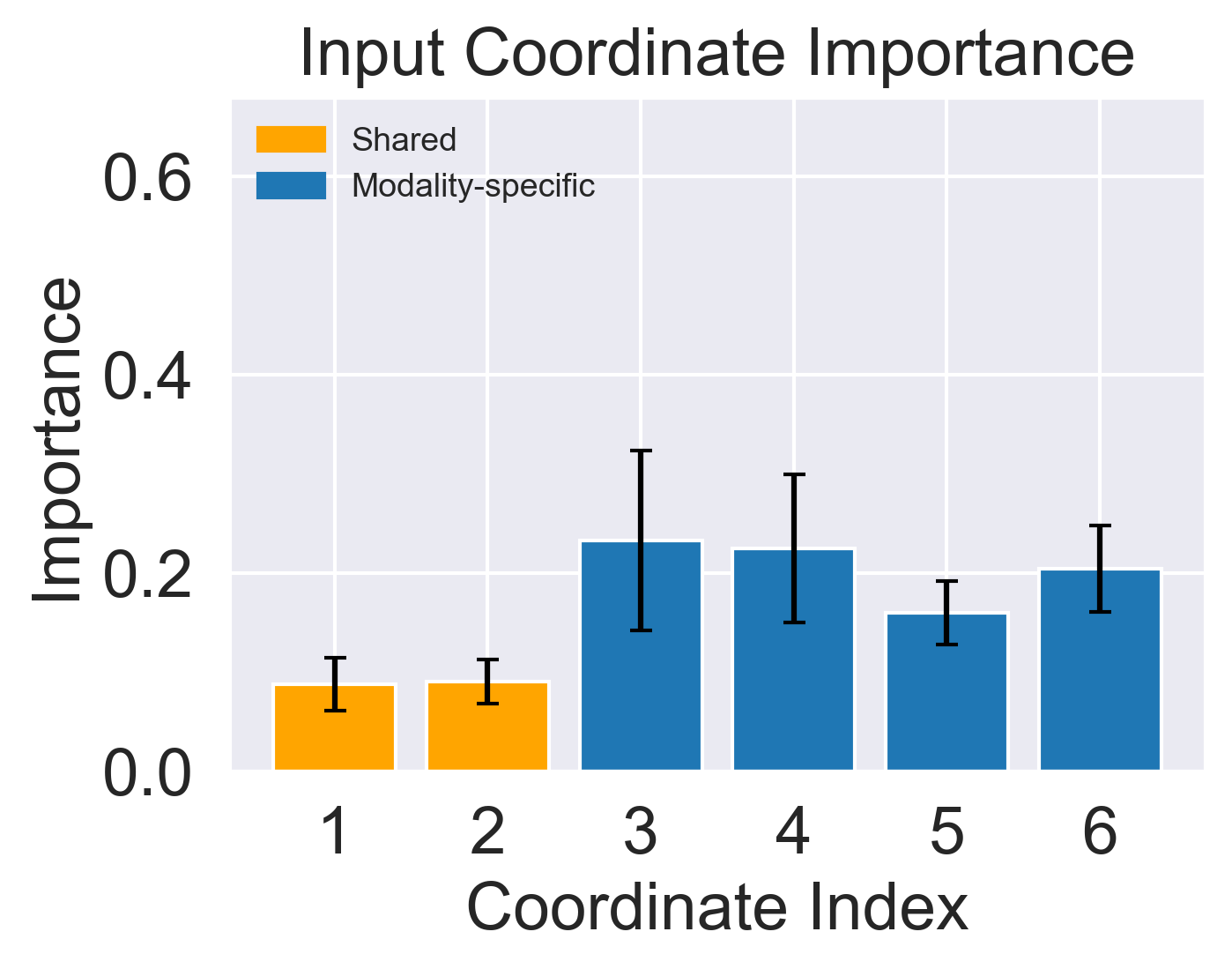}
        \caption{$\lambda=0.1$.}
        \label{fig:disen332}
    \end{subfigure}
    \hfill
    \begin{subfigure}[t]{0.225\linewidth}
        \centering
        \includegraphics[width=\linewidth]{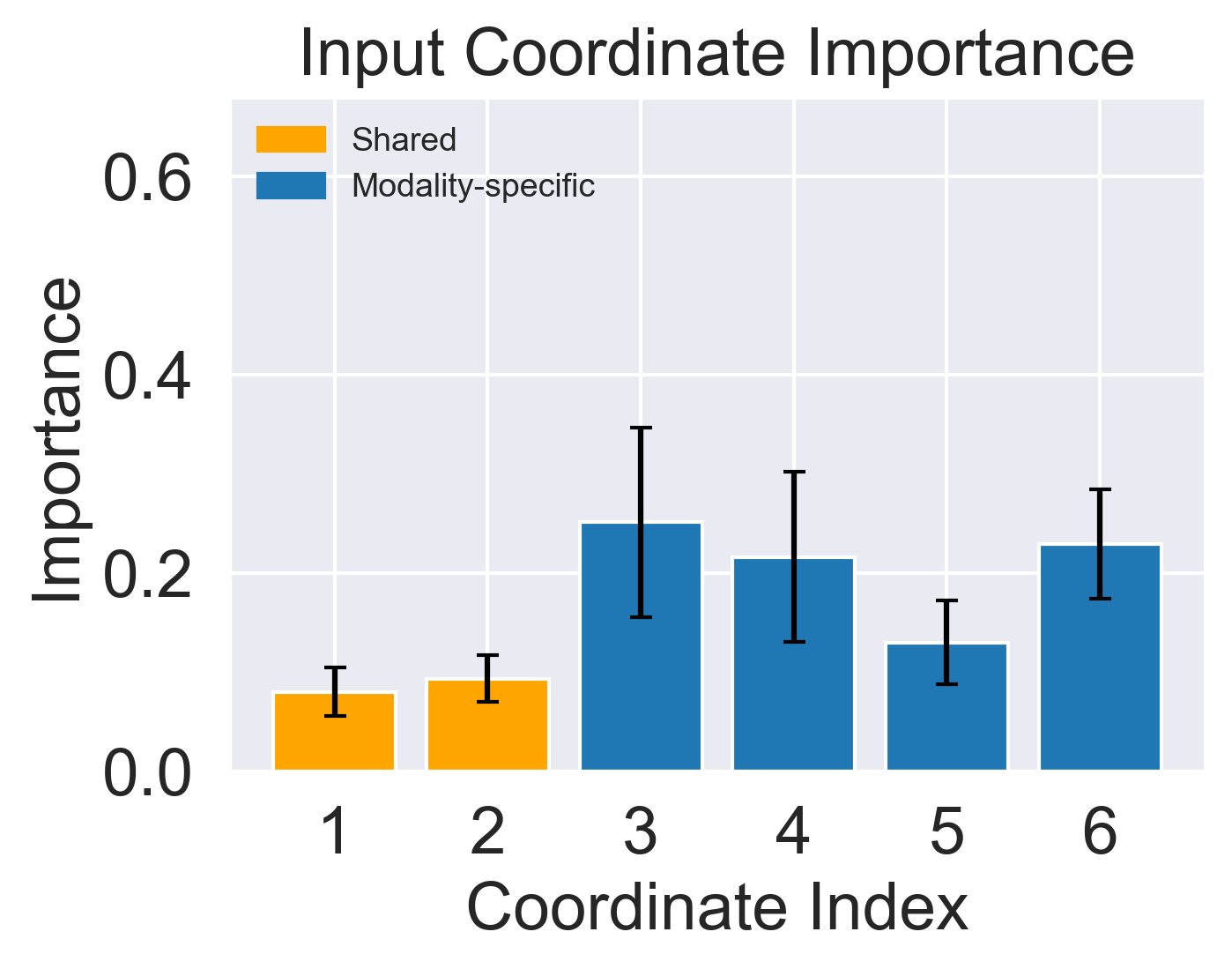}
        \caption{$\lambda=1.0$.}
        \label{fig:disen333}
    \end{subfigure}
    \hfill
    \begin{subfigure}[t]{0.225\linewidth}
        \centering
        \includegraphics[width=\linewidth]{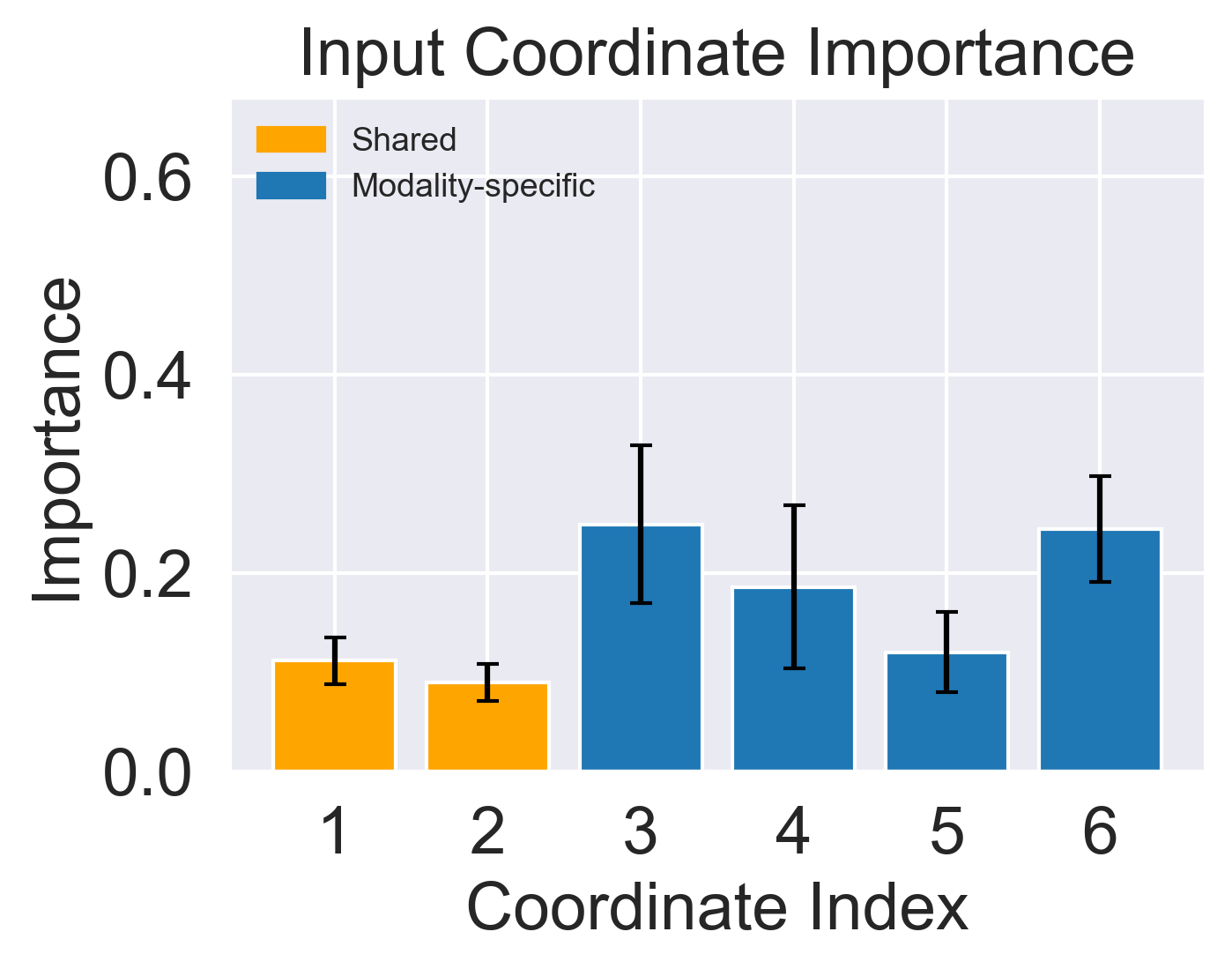}
        \caption{$\lambda=10.0$.}
        \label{fig:disen334}
    \end{subfigure}
    \caption{\yg{\disen~ with varying $\lambda$: feature importance averaged over $10$ simulations (Setting 3).}}
    \label{fig:setting6-disen}
\end{figure}

\begin{figure}[h]
    \centering
    \begin{subfigure}[t]{0.225\linewidth}
        \centering
        \includegraphics[width=\linewidth]{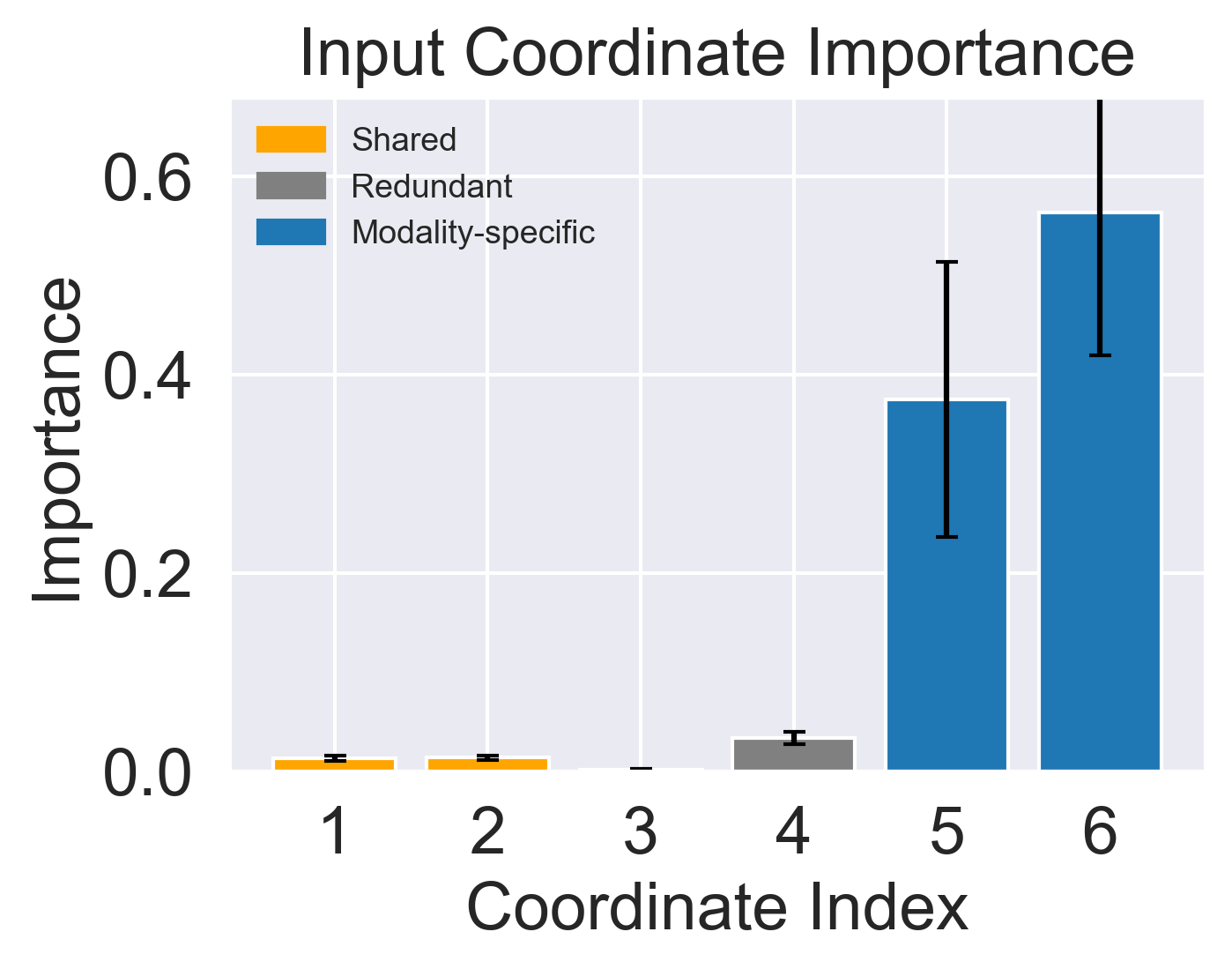}
        \caption{$\lambda=0.01$.}
        \label{fig:rec31}
    \end{subfigure}
    \hfill
    \begin{subfigure}[t]{0.225\linewidth}
        \centering
        \includegraphics[width=\linewidth]{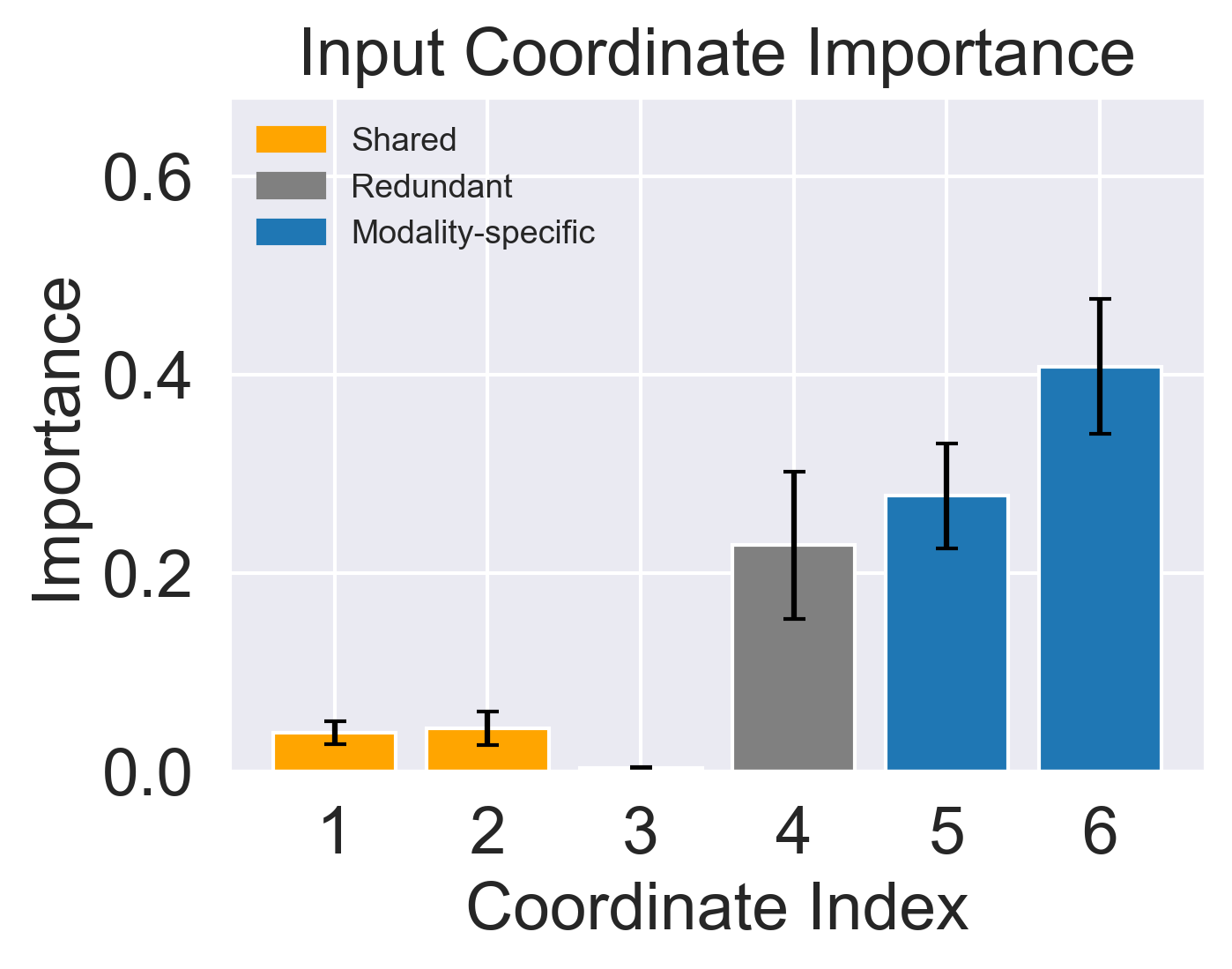}
        \caption{$\lambda=0.1$.}
        \label{fig:rec32}
    \end{subfigure}
    \hfill
    \begin{subfigure}[t]{0.225\linewidth}
        \centering
        \includegraphics[width=\linewidth]{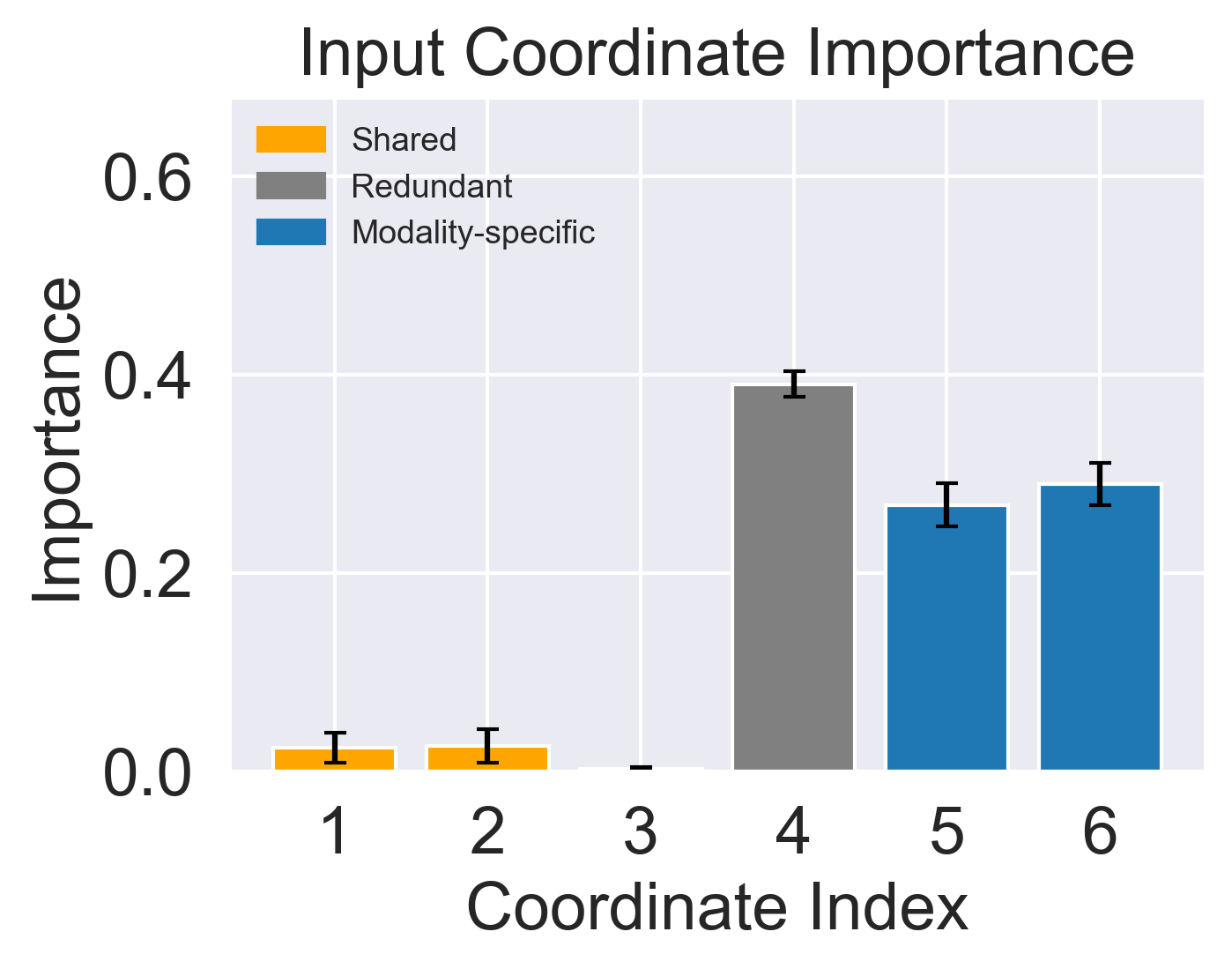}
        \caption{$\lambda=1.0$.}
        \label{fig:rec33}
    \end{subfigure}
    \hfill
    \begin{subfigure}[t]{0.225\linewidth}
        \centering
        \includegraphics[width=\linewidth]{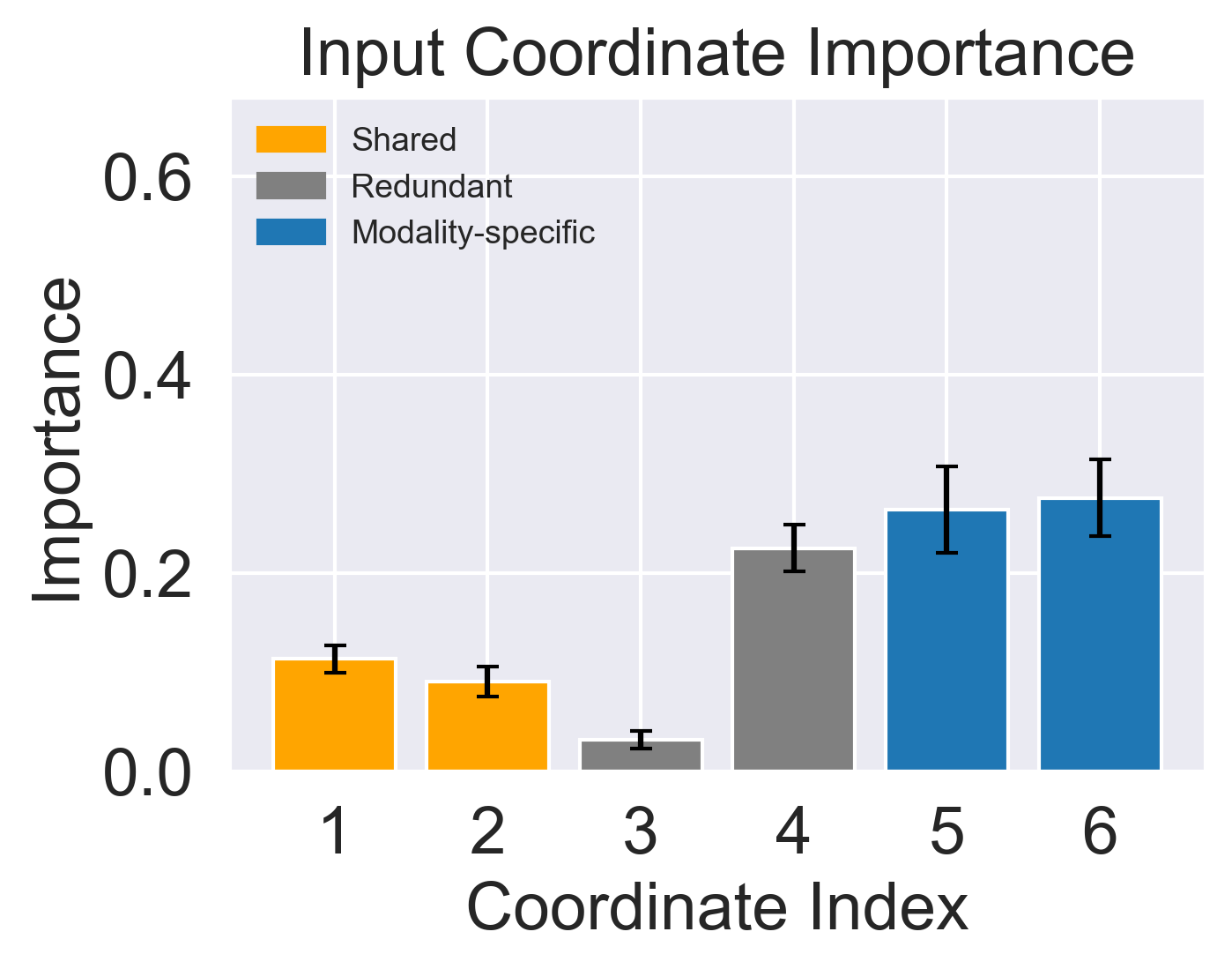}
        \caption{$\lambda=10.0$.}
        \label{fig:rec34}
    \end{subfigure}
    \caption{\yg{\ours~ with varying $\lambda$: feature importance averaged over $10$ simulations (Setting 4).}}
    \label{fig:setting7-recons}
\end{figure}

\begin{figure}[h]
    \centering
    \begin{subfigure}[t]{0.225\linewidth}
        \centering
        \includegraphics[width=\linewidth]{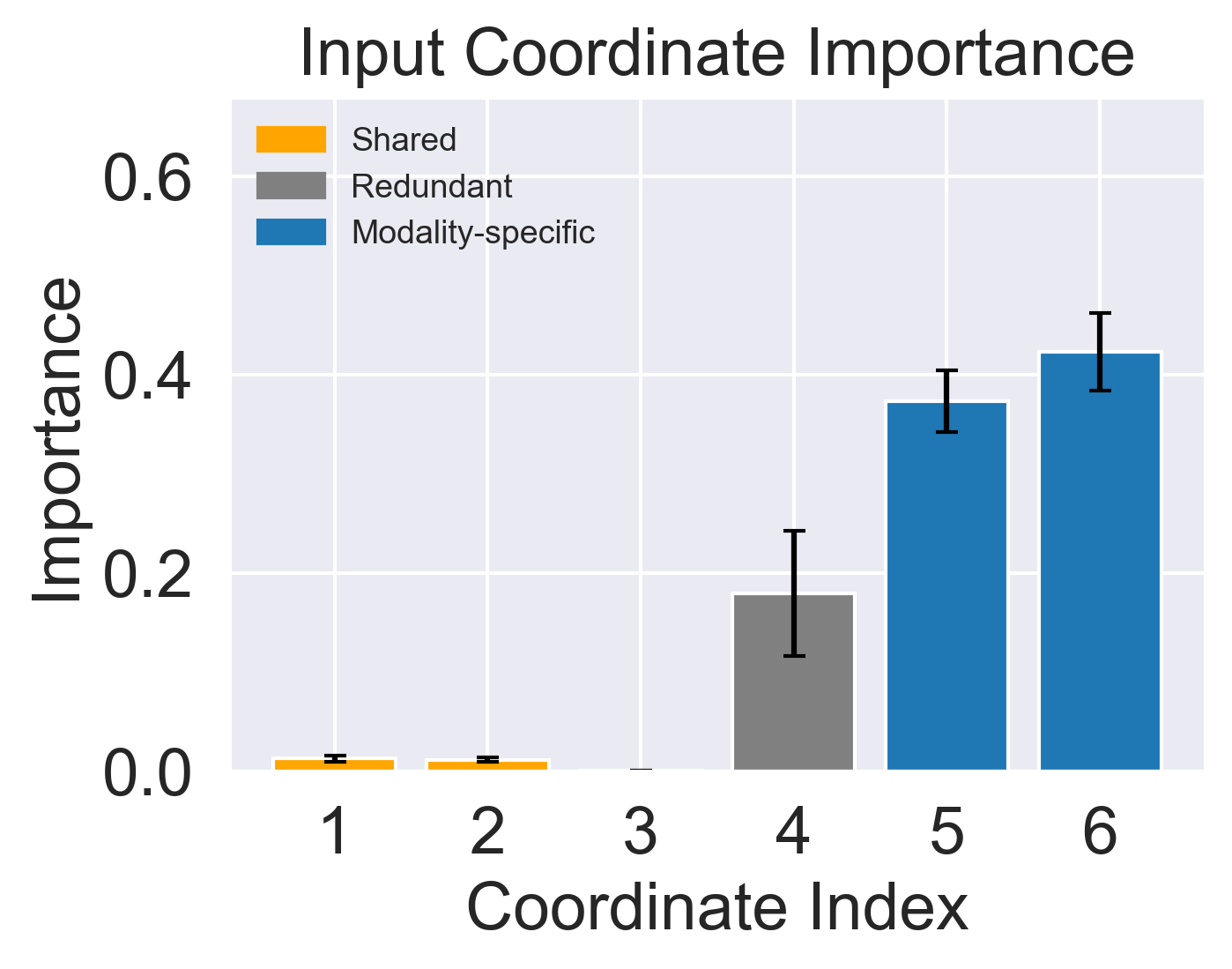}
        \caption{$\lambda=0.01$.}
        \label{fig:fact31}
    \end{subfigure}
    \hfill
    \begin{subfigure}[t]{0.225\linewidth}
        \centering
        \includegraphics[width=\linewidth]{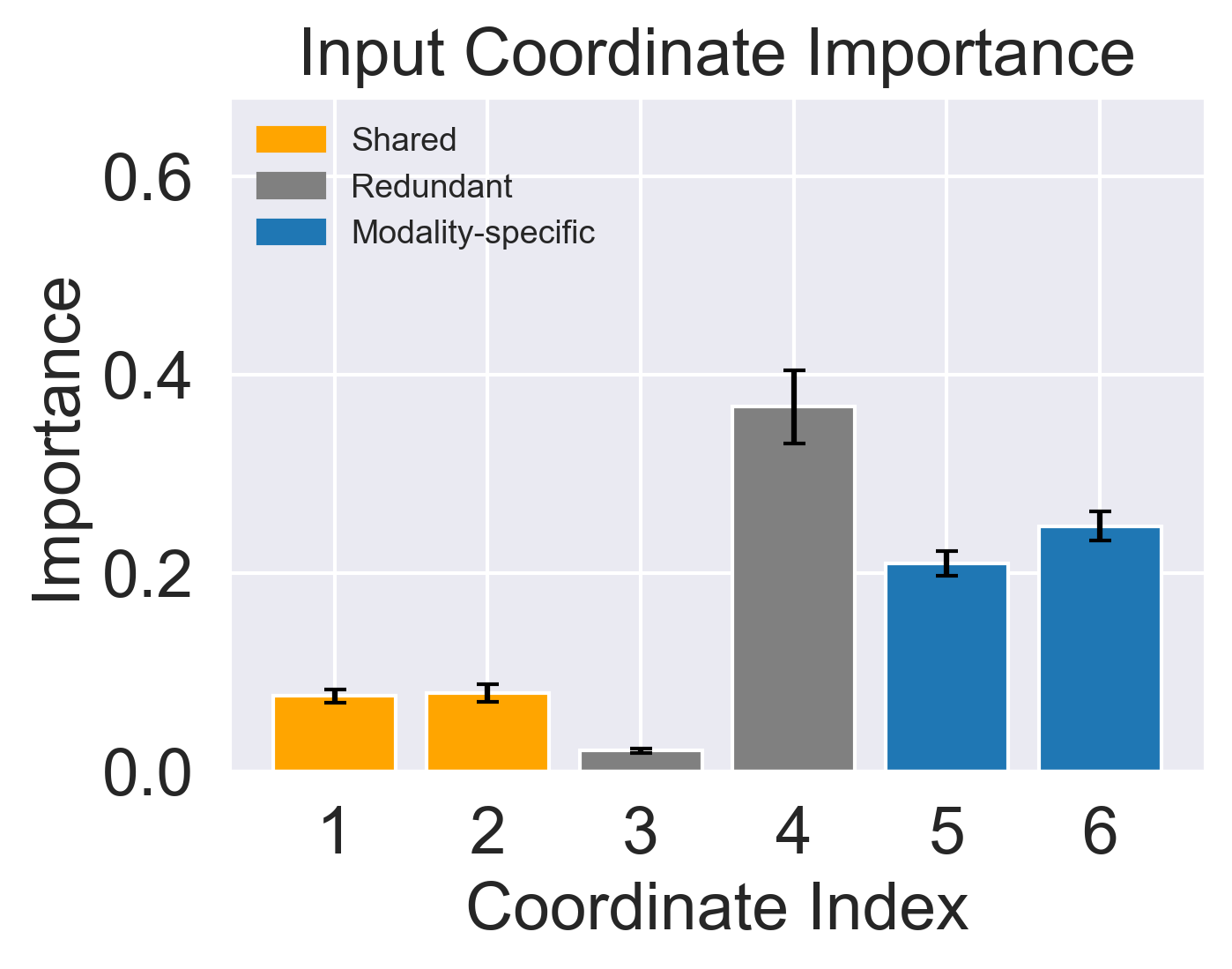}
        \caption{$\lambda=0.1$.}
        \label{fig:fact32}
    \end{subfigure}
    \hfill
    \begin{subfigure}[t]{0.225\linewidth}
        \centering
        \includegraphics[width=\linewidth]{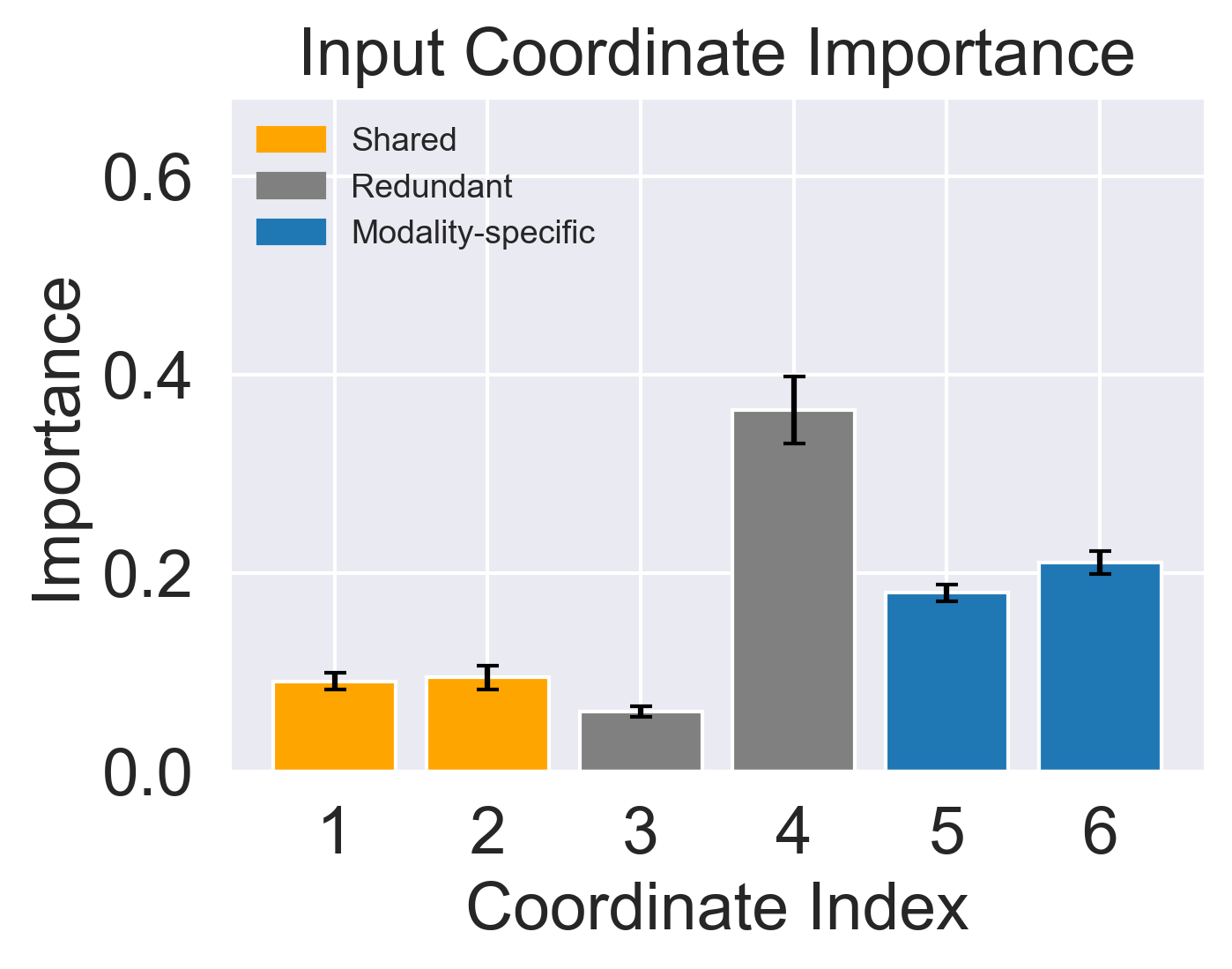}
        \caption{$\lambda=1.0$.}
        \label{fig:fact33}
    \end{subfigure}
    \hfill
    \begin{subfigure}[t]{0.225\linewidth}
        \centering
        \includegraphics[width=\linewidth]{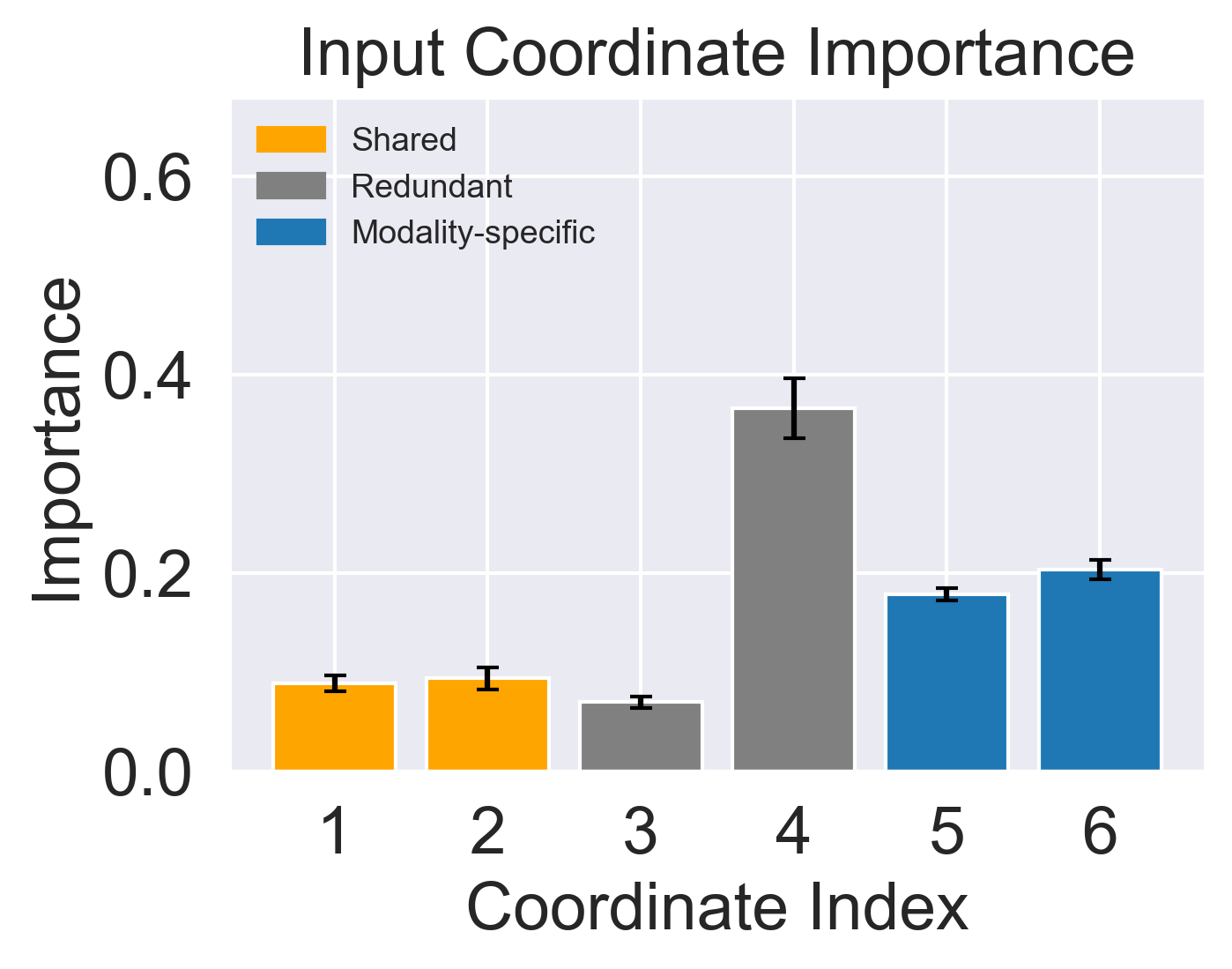}
        \caption{$\lambda=10.0$.}
        \label{fig:fact34}
    \end{subfigure}
    \caption{\yg{Factorized CL with varying $\lambda$: feature importance averaged over $10$ simulations (Setting 4).}}
    \label{fig:setting7-fact}
\end{figure}

\begin{figure}[!h]
    \centering
    \begin{subfigure}[t]{0.225\linewidth}
        \centering
        \includegraphics[width=\linewidth]{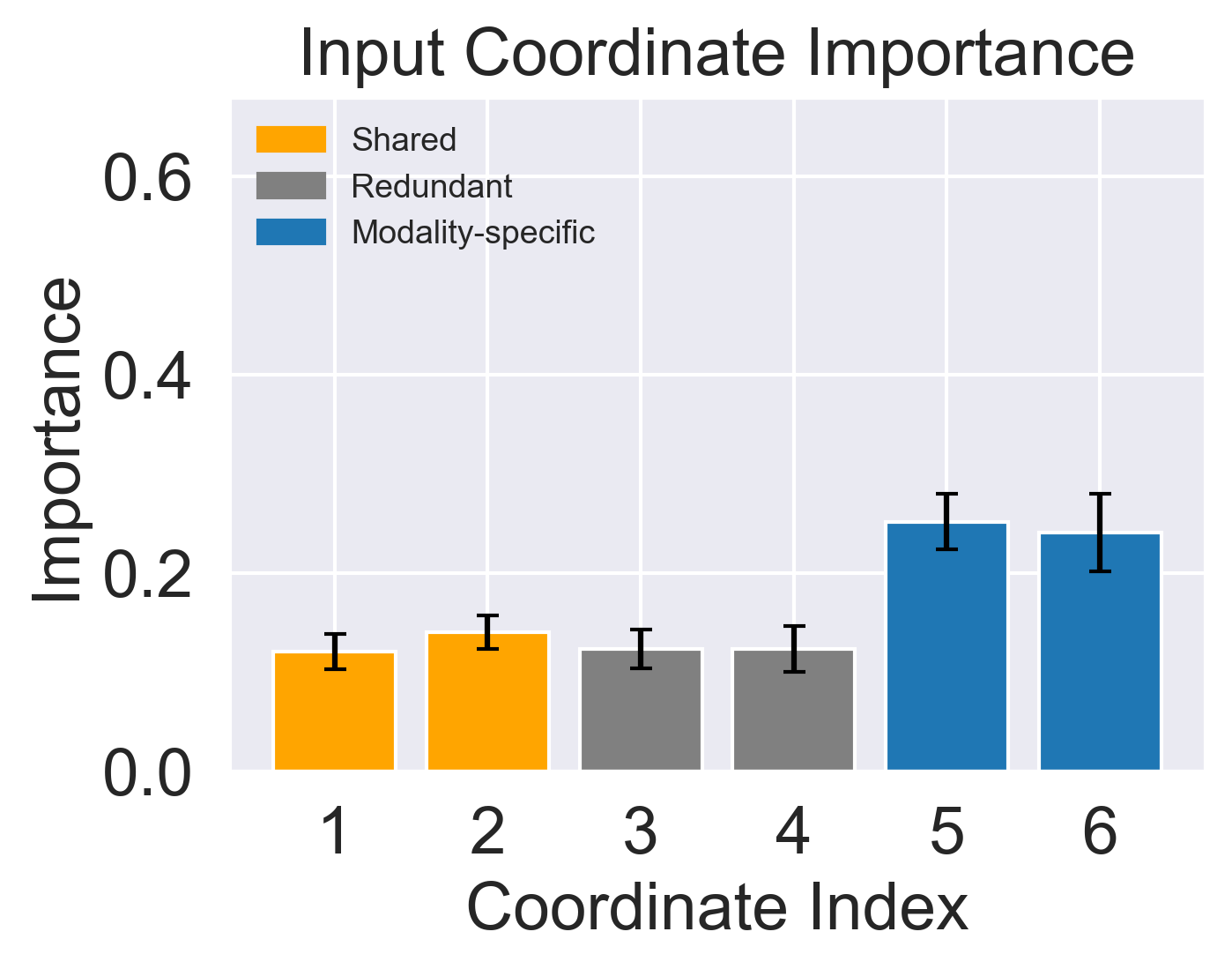}
        \caption{$\lambda=0.01$.}
        \label{fig:disen31}
    \end{subfigure}
    \hfill
    \begin{subfigure}[t]{0.225\linewidth}
        \centering
        \includegraphics[width=\linewidth]{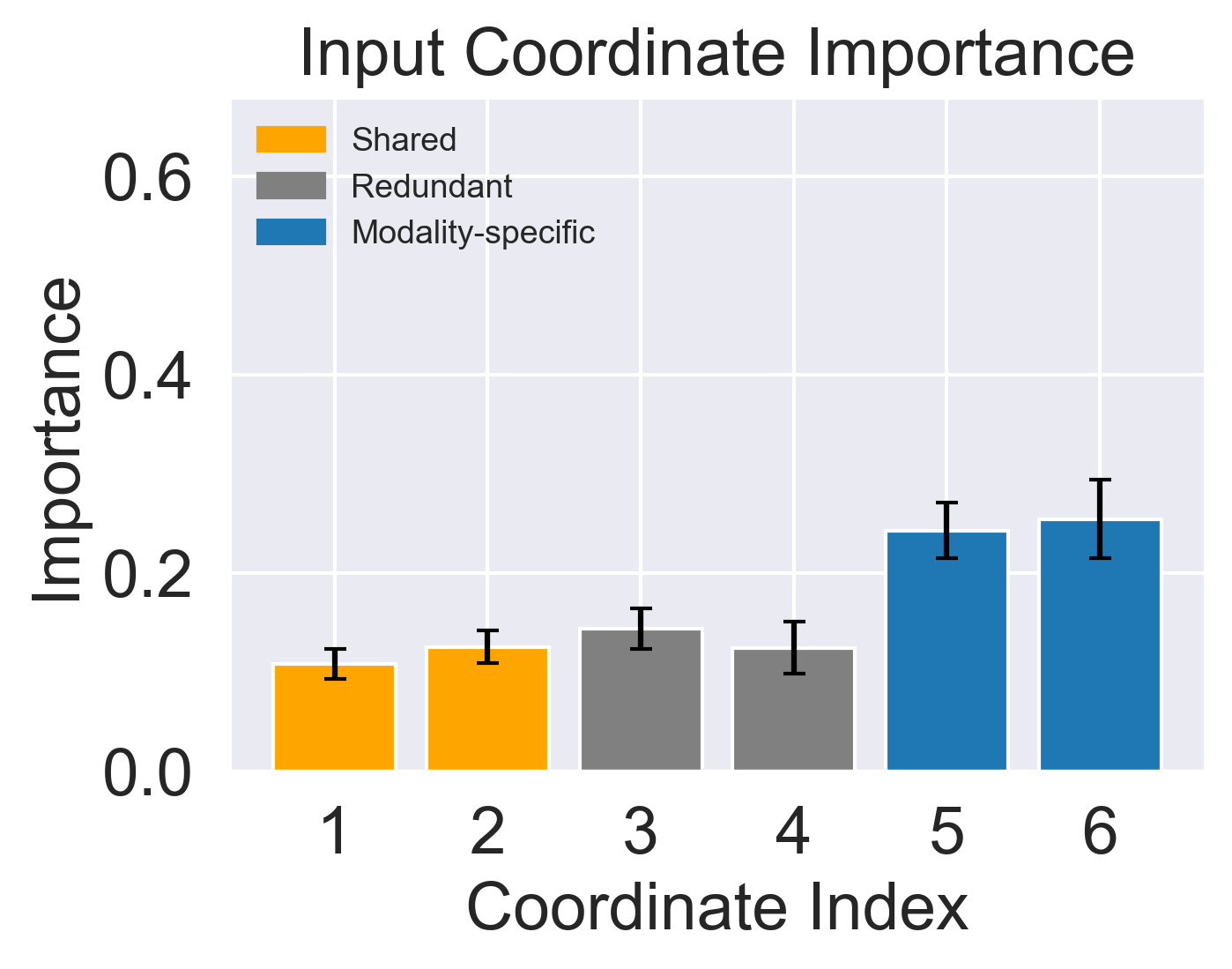}
        \caption{$\lambda=0.1$.}
        \label{fig:disen32}
    \end{subfigure}
    \hfill
    \begin{subfigure}[t]{0.225\linewidth}
        \centering
        \includegraphics[width=\linewidth]{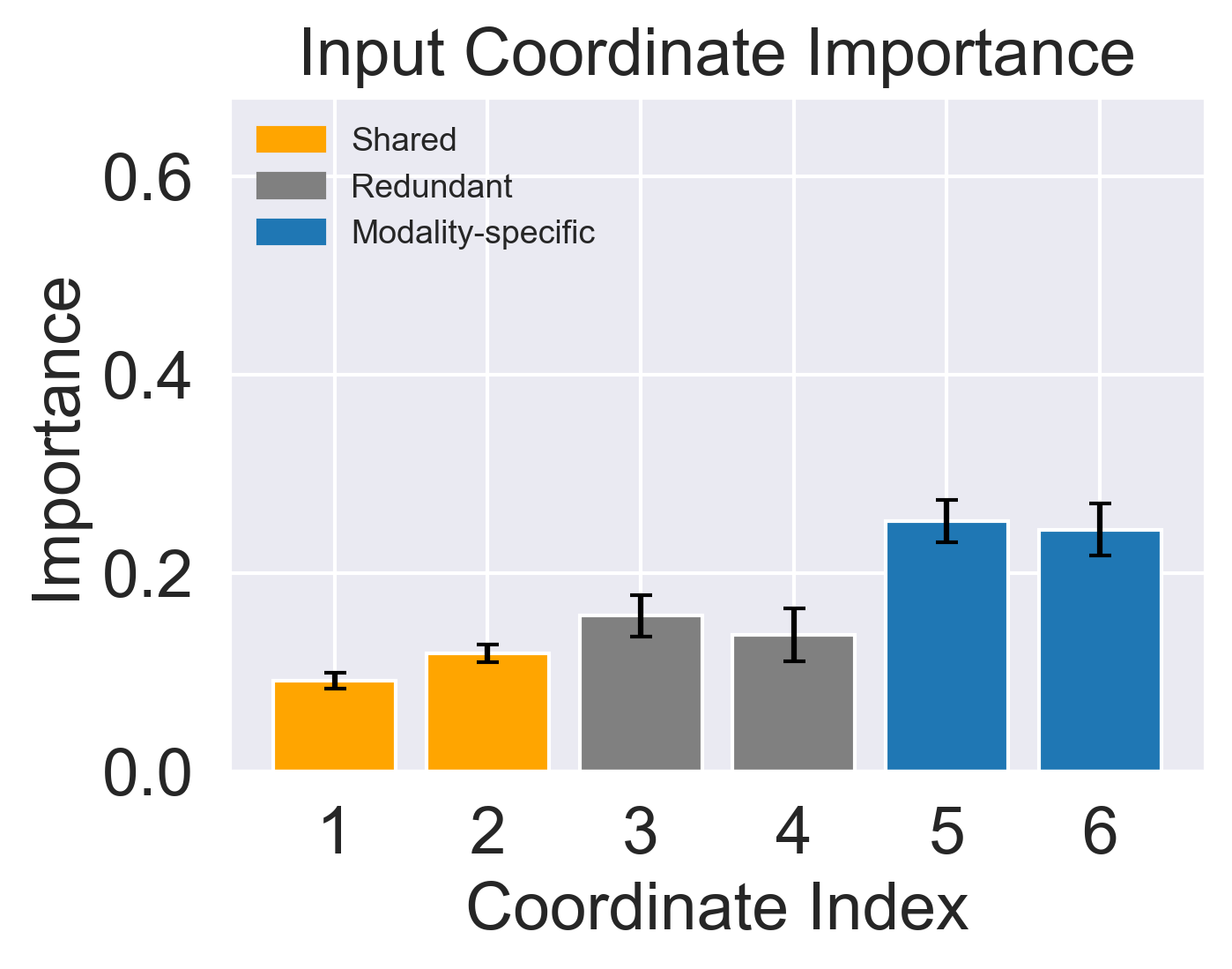}
        \caption{$\lambda=1.0$.}
        \label{fig:disen33}
    \end{subfigure}
    \hfill
    \begin{subfigure}[t]{0.225\linewidth}
        \centering
        \includegraphics[width=\linewidth]{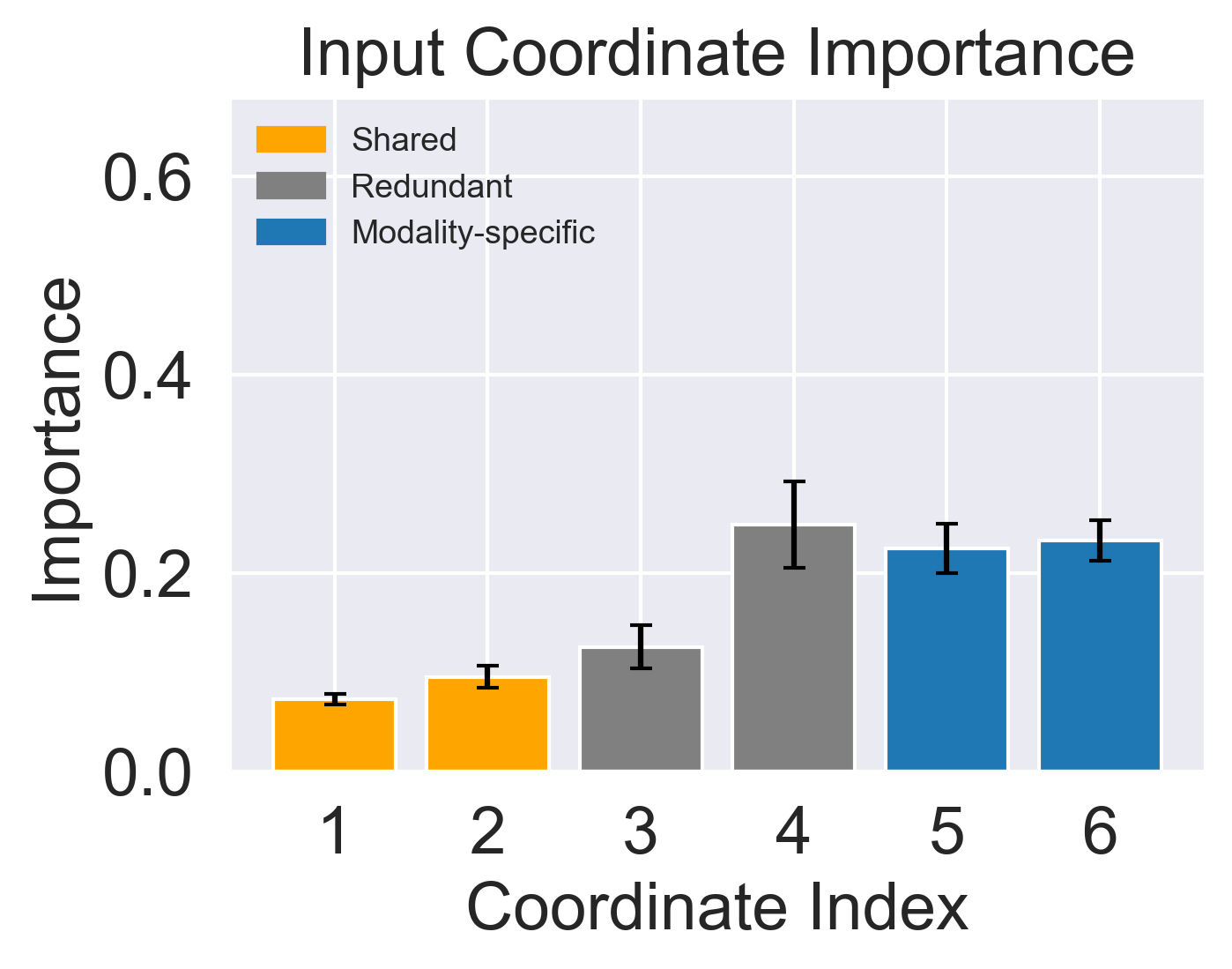}
        \caption{$\lambda=10.0$.}
        \label{fig:disen34}
    \end{subfigure}
    \caption{\yg{\disen\; with varying $\lambda$: feature importance averaged over $10$ simulations (Setting 4).}}
    \label{fig:setting7-disen}
\end{figure}

We also consider ablated variants of Settings 1 and 2 in Section~\ref{sec:simu}. Concretely, we generate modality-specific input coordinates (i.e., $(X_1)_{d_c+1:d}$ in Setting 1 and $(X_1)_{d_c+3:d}$ in Setting 2) independently from a discrete distribution
\begin{align}
    (X_1)_j \sim {\rm Unif}\{-\sqrt{7}/2,-1,-1/2, 1/2,1, \sqrt{7}/2\}.
\end{align}
Results for variants (Setting 3 as a variant for Setting 1, and Setting 4 as a variant for Setting 2) are presented in Figures \ref{fig:setting6-recons}--\ref{fig:setting7-disen}. 
Performances of different approaches are comparable to those under the Gaussian settings.

\subsection{CITE-seq dataset implementation details}\label{sec:app-citeseq}

% \subsubsection{}
We consider the bone marrow CITE-seq dataset from \cite{stuart2019comprehensive} which consists of measurements in two modalities on $30672$ individual cells: transcriptome (RNA) and $25$ cell-surface proteins (ADT). 
We randomly split the dataset into a training set of size $15000$ and a test set of size $15672$.
% We follow the pre-processing in \cite{hao2021integrated},\footnote{\url{https://satijalab.org/seurat/articles/weighted_nearest_neighbor_analysis}} where we conduct scaling and centering based on moments estimated from the training set only.
We follow the preprocessing in \url{https://satijalab.org/seurat/articles/weighted_nearest_neighbor_analysis}, where we normalize ADT data with centering to produce a 24-dimensional input based on moments estimated from the training set only, and normalize the high-dimensional RNA data and extract the first 200 principal components as the inputs for CLIP.

In the experiments presented in Figure~\ref{fig:cite-seq-recons} and Table~\ref{fig:rank-corr-comparison}, architectures of representation maps $g$ and $h$ for three methods are all $5$-layer ReLU networks with width $50$ and output dimension $50$. Throughout experiments with real-world datasets, we adopt a batch size of $128$ and a learning rate of $10^{-4}$ for training, in which we set the number of epochs to $2000$.

\begin{figure}[!h]
    \centering
    \includegraphics[width=0.95\linewidth]{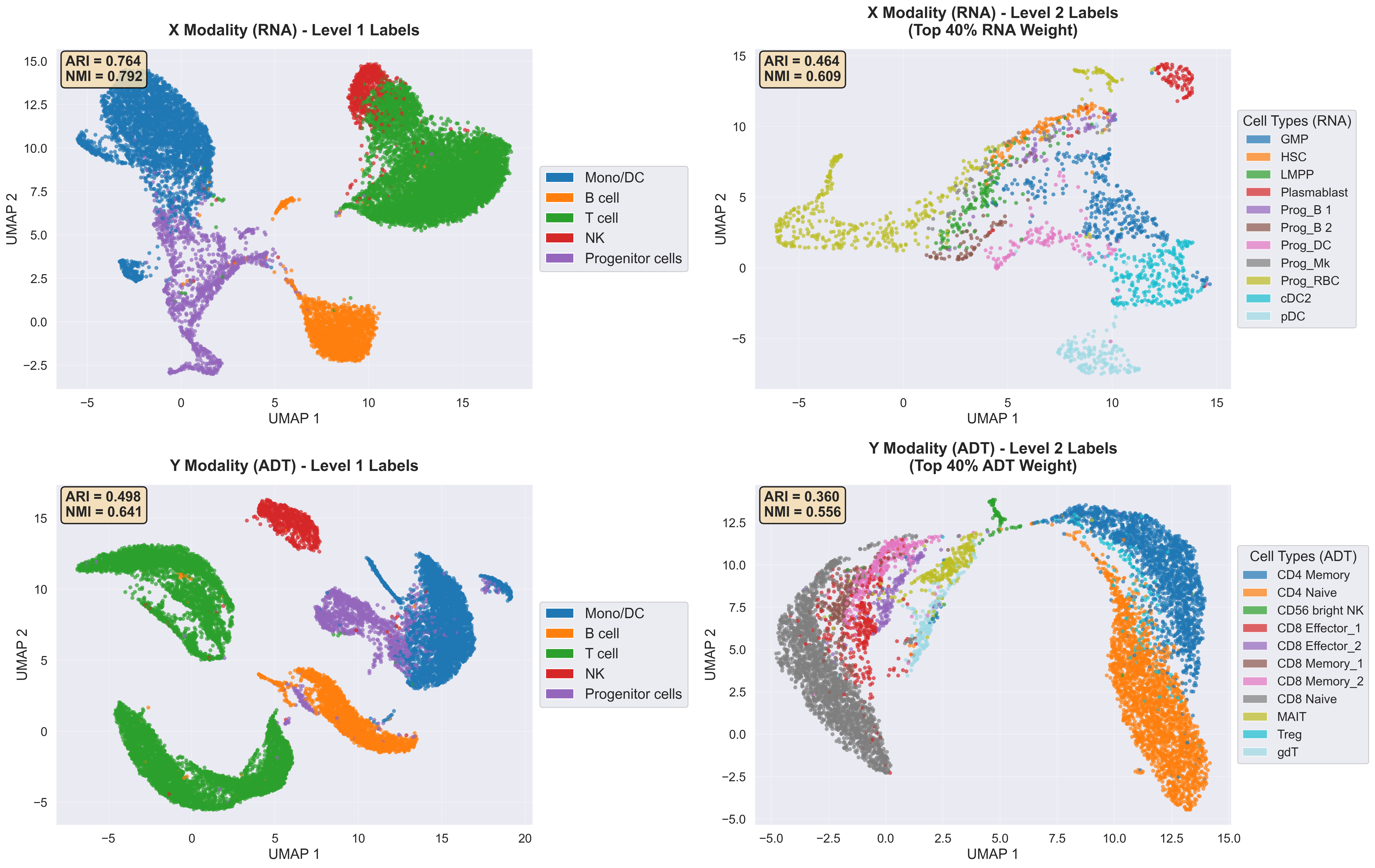}
    \caption{\yg{UMAP of \ours-learned representations $(C_i,Z_i)$\;(colored by cell types).}}
    \label{fig:umap-citeseq-indiseek}
\end{figure}

\begin{figure}[!h]
    \centering
    \includegraphics[width=0.95\linewidth]{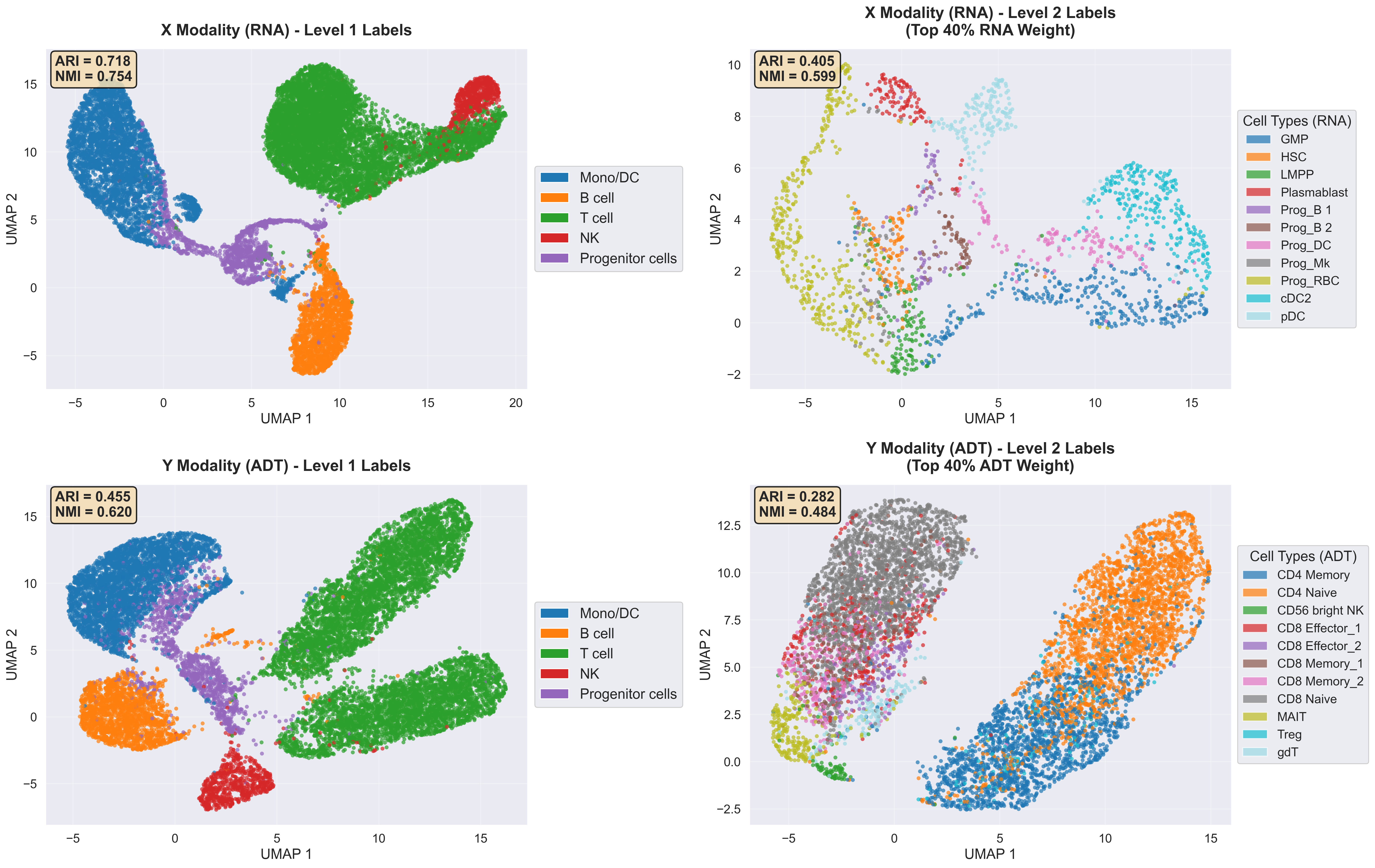}
    \caption{\yg{UMAP of FactorizedCL-learned representations $(C_i,Z_i)$\;(colored by cell types).}}
    \label{fig:umap-citeseq-fact}
\end{figure}

\begin{figure}[!h]
    \centering
    \includegraphics[width=0.95\linewidth]{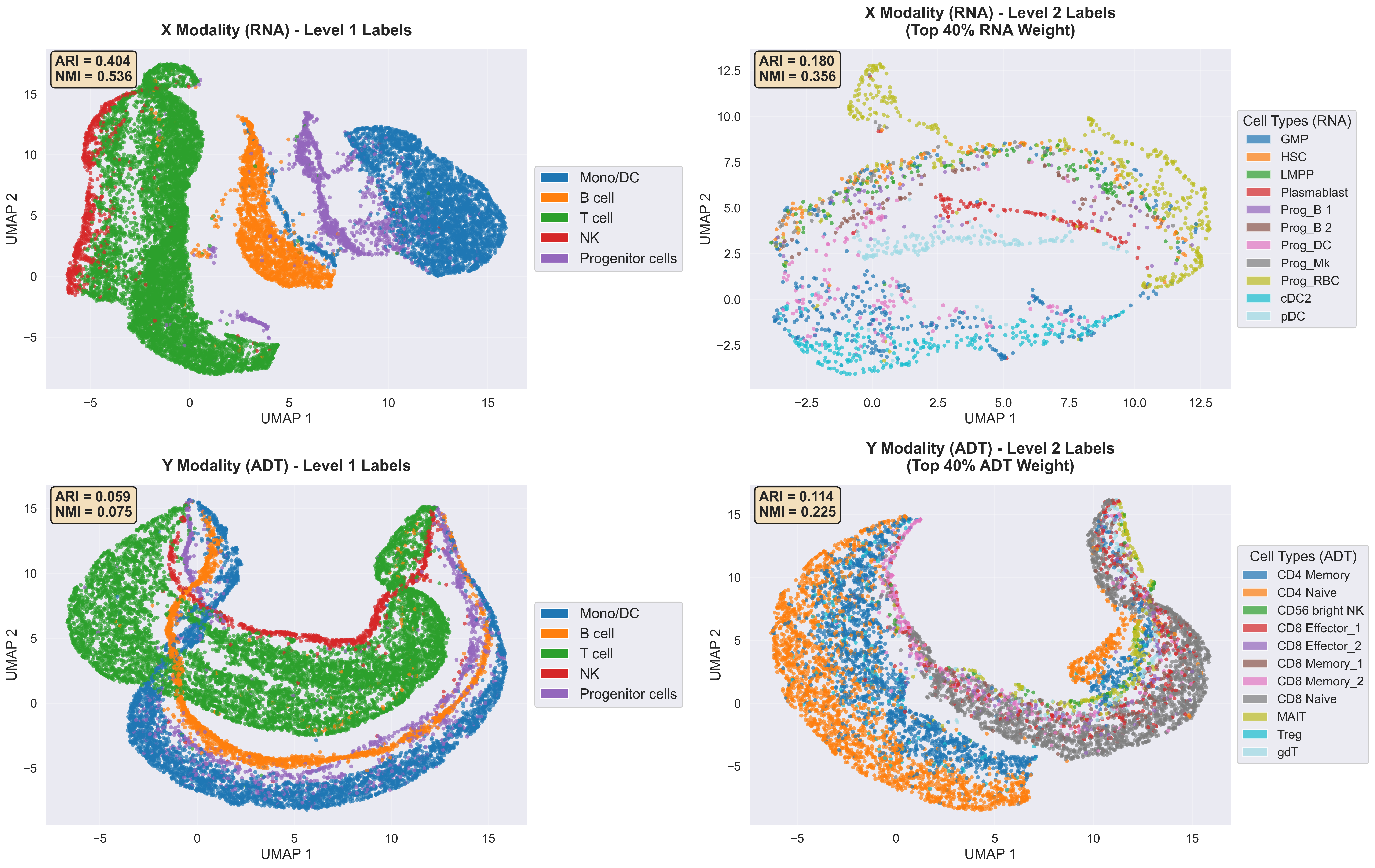}
    \caption{\yg{UMAP of InfoDisen-learned representations $(C_i,Z_i)$\;(colored by cell types).}}
    \label{fig:umap-citeseq-disen}
\end{figure}

\begin{figure}[!h]
    \centering
    \includegraphics[width=\linewidth]{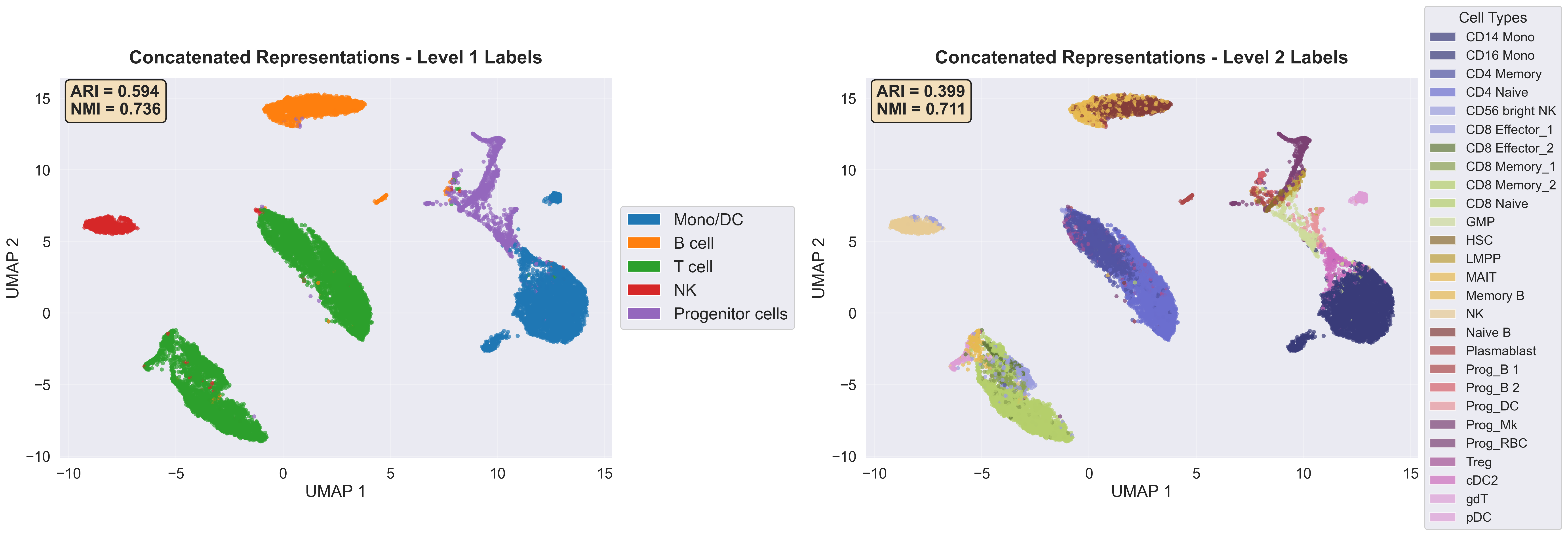}
    \caption{\yg{UMAP of concatenated \ours-learned representations $(C_1,C_2,Z_1,Z_2)$\;(colored by cell types).}}
    \label{fig:umap-citeseq-indiseek-concat}
\end{figure}

\begin{figure}[!h]
    \centering
    \includegraphics[width=\linewidth]{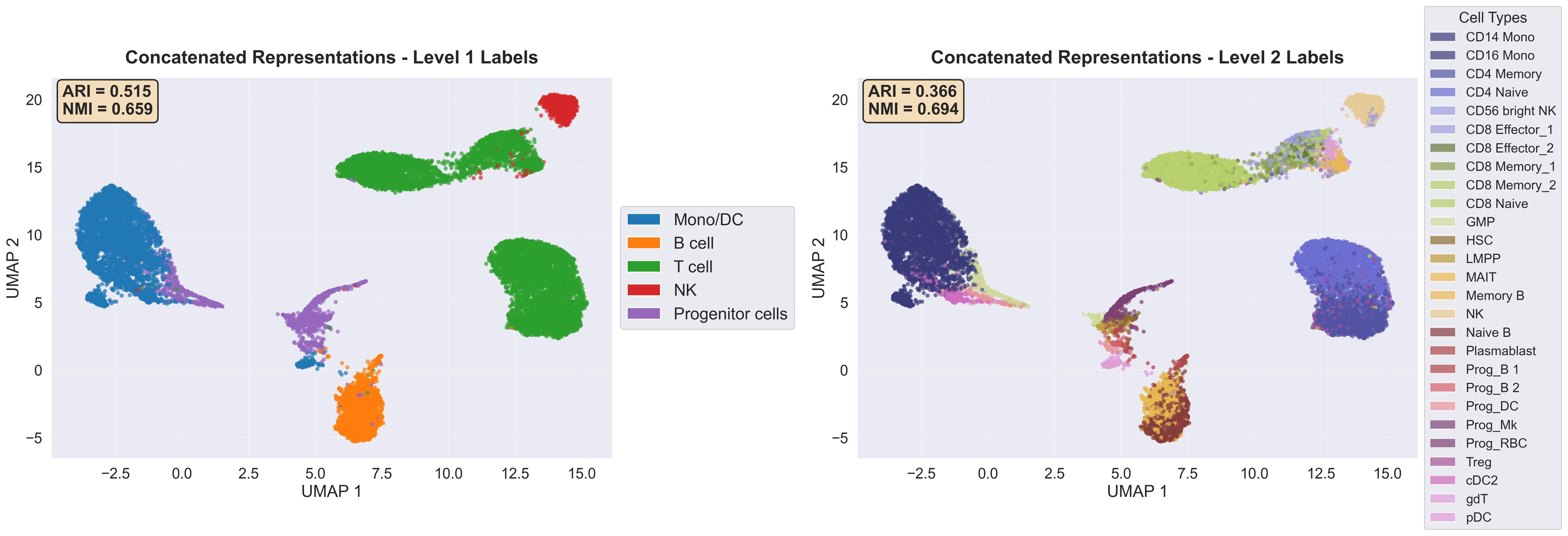}
    \caption{\yg{UMAP of concatenated FactorizedCL-learned representations $(C_1,C_2,Z_1,Z_2)$\;(colored by cell types).}}
    \label{fig:umap-citeseq-fact-concat}
\end{figure}

\begin{figure}[!h]
    \centering
    \includegraphics[width=\linewidth]{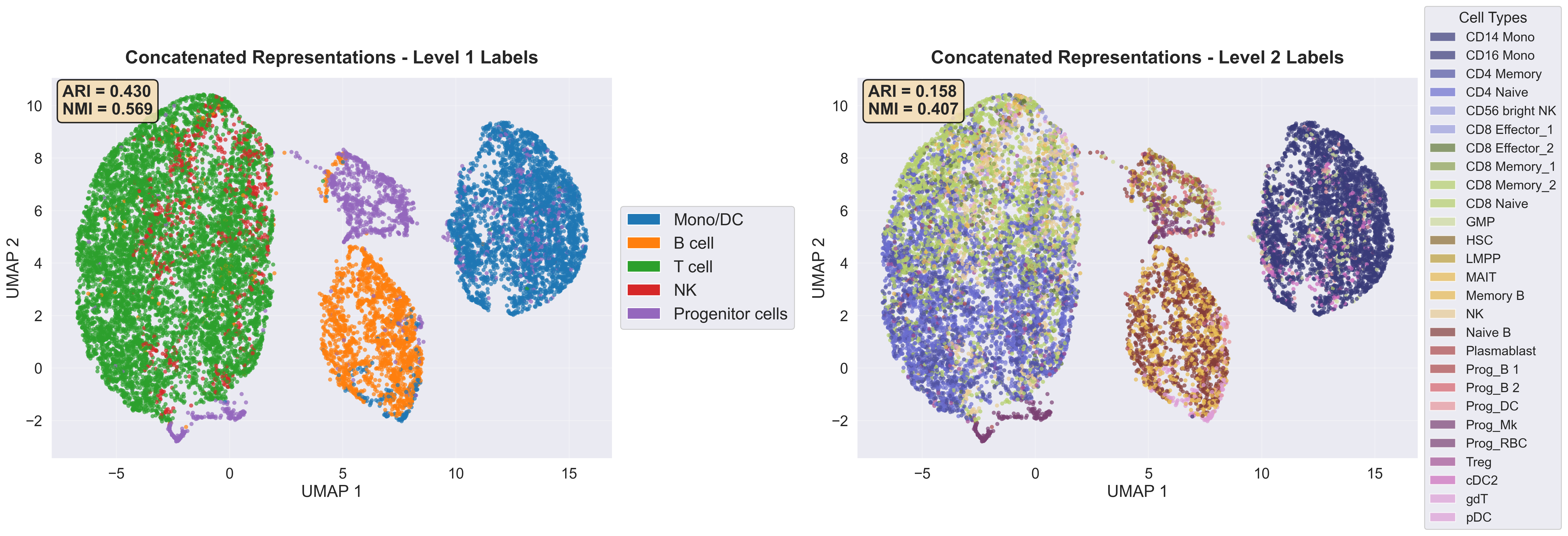}
    \caption{\yg{UMAP of concatenated InfoDisen-learned representations $(C_1,C_2,Z_1,Z_2)$\;(colored by cell types).}}
    \label{fig:umap-citeseq-disen-concat}
\end{figure}

\yg{
In Figure~\ref{fig:umap-citeseq-indiseek}, Figure~\ref{fig:umap-citeseq-fact}, and Figure~\ref{fig:umap-citeseq-disen}, we visualize the learned representations $(C_i,Z_i)$ by all three methods for each modality. 
The top rows correspond to the RNA modality, and the bottom rows to the ADT modality.
In the left panels, the representations are colored by level-1 cell types, and in the right panels, to ease visualization, we plot representations of a subset of cells whose level-2 cell types are among the top-$40\%$ in the average benchmark weights across all level-2 cell types in the corresponding modality.
In other words, the chosen level-2 cell types for each modality are those for which the corresponding modality is more important for their distinction.
If a disentangled learning approach works well, the resulting modality-specific representations of the chosen level-2 cell types are expected to be well separated in the respective UMAPs.
Additionally, in Figure~\ref{fig:umap-citeseq-indiseek-concat}, Figure~\ref{fig:umap-citeseq-fact-concat}, and Figure~\ref{fig:umap-citeseq-disen-concat}, we visualize the concatenated representations $(C_1,C_2,Z_1,Z_2)$ learned by all three methods, in which 
% the coloring of data follows that mentioned before.
representations are colored by level-1 and level-2 cell types in the left and the right panels, respectively. Here, we choose the optimal $\lambda$ implied by Figure~\ref{fig:rank-corr-comparison} for each method, respectively: $0.1$ for FactorizedCL, $1.0$ for \disen, and $10.0$ for \ours.
}

\yg{
To quantify the separation of cell types, we adopt two metrics: (1) the Adjusted Rand Index (ARI)~\citep{hubert1985comparing}, which measures agreement between $k$-means clustering on the learned representations and the ground-truth labels, with values ranging from $-1$ to $1$ where higher values indicate better cluster recovery; and
(2) NMI (normalized mutual information) following 
\citep{luecken2022benchmarking,pedregosa2011scikit}, which quantifies the mutual dependence between $k$-means clustering and the ground-truth labels, with values in $[0, 1]$ where higher scores indicate better alignment between the discovered clusters and true labels.
Figures \ref{fig:umap-citeseq-indiseek}-\ref{fig:umap-citeseq-disen-concat} demonstrate that, when measured by these two metrics, {\ours} outperforms the two SOTA methods in both learned modality-specific representations and the overall learned representations.
}
% (2) the Silhouette Score~\citep{rousseeuw1987silhouettes}, which quantifies cluster compactness and separation using the ground-truth labels, with values in $[-1, 1]$ where higher scores indicate better-defined clusters.}

\begin{figure}[ht]
    \centering
    \includegraphics[width=\linewidth]{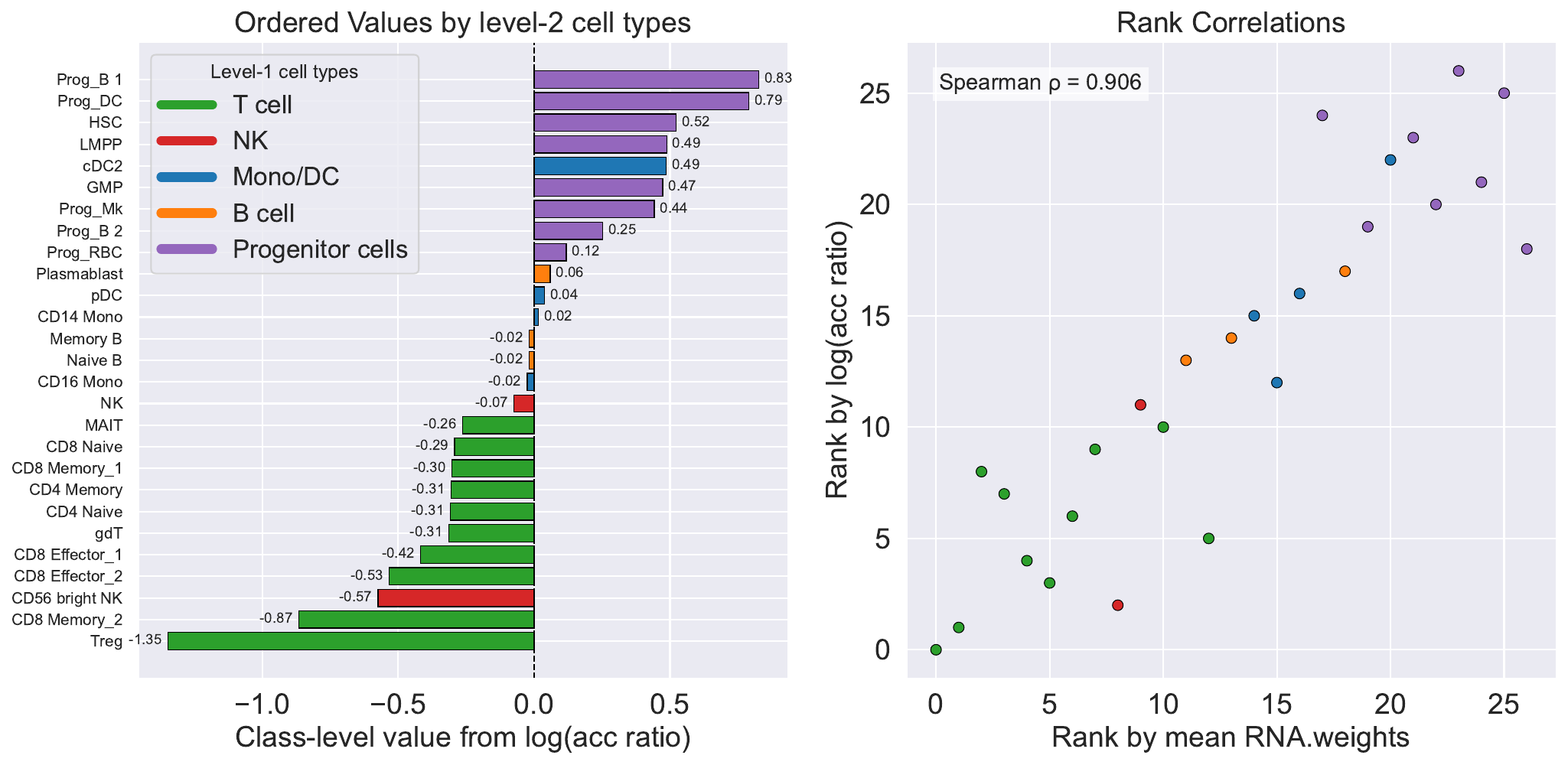}
    \caption{\yg{Performance of \ours\; in CITE-seq dataset ($\lambda=10.0$).}}
    \label{fig:cite-seq-recons-raw}
\end{figure}

\begin{figure}[ht]
    \centering
    \includegraphics[width=0.68\linewidth]{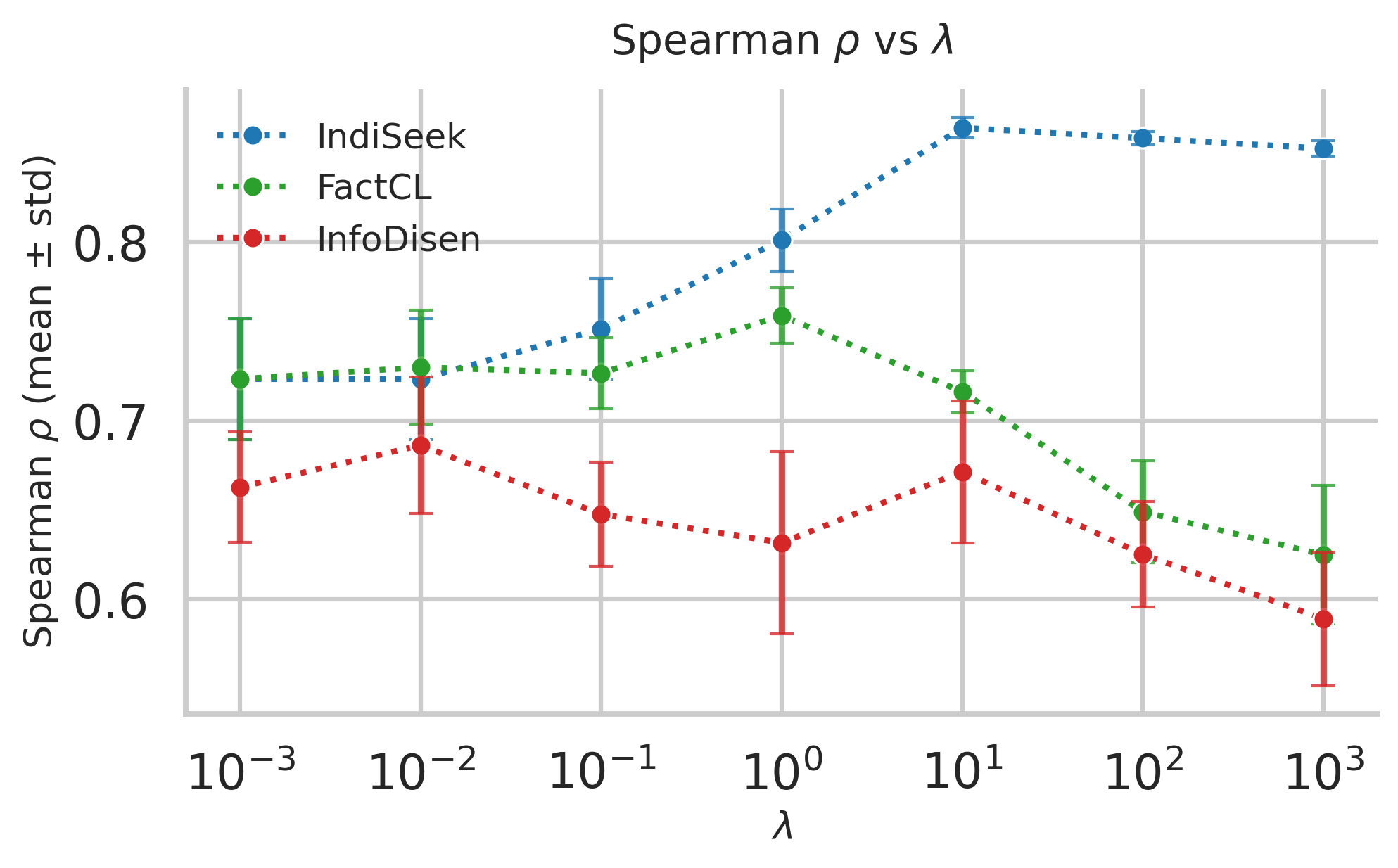}
    \caption{\yg{Comparison of rank correlation metrics across three methods on the CITE-seq dataset.}}
    \label{fig:rank-corr-comparison-raw}
\end{figure}

\yg{
\subsubsection{CITE-seq results without dimension reduction of RNA data}
We also present the results with CITE-seq data following the preprocessing in \url{https://satijalab.org/seurat/articles/weighted_nearest_neighbor_analysis}, where we normalize ADT data with centering to produce a 24-dimensional input based on moments estimated from the training set only, and normalize the extremely high-dimensional RNA data. Here, instead of extracting the first 200 principal components as the inputs for CLIP, we adopt the raw normalized $2000$-dimensional RNA data for training. The results are presented in Figure~\ref{fig:cite-seq-recons-raw} and Figure~\ref{fig:rank-corr-comparison-raw}, respectively.
Compared with Figures~\ref{fig:cite-seq-recons}-\ref{fig:rank-corr-comparison}, results with and without the PCA dimension reduction for the RNA modality are comparable.
}

\subsection{MultiBench datasets}\label{sec:app-multi}

We evaluate the performance of our method on MultiBench datasets \cite{liang2021multibench}, including video datasets (MOSI \citep{zadeh2016mosi}, MOSEI \citep{zadeh2018multimodal}, UR-FUNNY \citep{hasan2019ur}, MUStARD \citep{castro2019towards}) with modalities of text, image, and audio, and the MIMIC dataset \citep{johnson2016mimic}.
For video datasets, following \cite{liang2023factorized,wang2024information}, we focus solely on vision and text modalities, excluding audio data in these examples.
More concretely, 
\begin{itemize}
\item \textbf{MOSI} \cite{zadeh2016mosi}: $2199$ YouTube clips with vision and text modalities. Task: binary sentiment classification.
\item \textbf{MOSEI} \citep{zadeh2018multimodal}: $23000$ monologue videos with vision and text modalities. Task: binary sentiment classification.
\item \textbf{UR-FUNNY} \citep{hasan2019ur}: TED talk videos with vision and text modalities. Task: binary humor detection.
\item \textbf{MUStARD} \citep{castro2019towards}: $690$ TV show clips with vision and text modalities. Task: binary sarcasm detection.
\item \textbf{MIMIC} \citep{johnson2016mimic}: over $36000$ ICU patient records with time-series vitals and tabular data. Task: Following \cite{wang2024information}, we aim at predicting ICD-9 group $7$ membership.
\end{itemize}

We follow the processing pipeline in \cite{liang2023factorized} and adopt exactly the same data splitting and feature pre-extraction. We use a smaller Transformer architecture with $2$ heads, $2$ layers, and the intermediate layers with a width $128$.

\begin{table}[h]
\centering
\caption{\yg{MOSI results with varying $\lambda$ (averaged over 10 seeds, standard errors in parentheses). All values are percentages.}}
\label{tab:mosi}
\begin{tabular}{lcccccccc}
\toprule
$\lambda$ & 0.01 & 0.1 & 1.0 & 10.0 & 100.0 & 1000.0 & max over $\lambda$ \\
\midrule
IndiSeek & 67.46$_{\scriptstyle(0.74)}$ & 69.48$_{\scriptstyle(0.86)}$ & 70.03$_{\scriptstyle(1.39)}$ & 66.43$_{\scriptstyle(0.34)}$ & 65.95$_{\scriptstyle(0.45)}$ & 65.90$_{\scriptstyle(0.89)}$ & 70.03$_{\scriptstyle(1.39)}$ \\
FactorizedCL & 66.76$_{\scriptstyle(0.58)}$ & 65.07$_{\scriptstyle(0.92)}$ & 66.12$_{\scriptstyle(0.67)}$ & 64.26$_{\scriptstyle(1.07)}$ & 65.07$_{\scriptstyle(0.58)}$ & 65.12$_{\scriptstyle(0.89)}$ & 67.11$_{\scriptstyle(0.34)}$ \\
InfoDisen & 65.57$_{\scriptstyle(0.84)}$ & 65.31$_{\scriptstyle(1.26)}$ & 65.83$_{\scriptstyle(1.00)}$ & 65.22$_{\scriptstyle(1.08)}$ & 64.84$_{\scriptstyle(1.26)}$ & 67.52$_{\scriptstyle(0.62)}$ & 67.52$_{\scriptstyle(0.62)}$ \\
CLIP & 67.61$_{\scriptstyle(0.66)}$ & 67.61$_{\scriptstyle(0.66)}$ & 67.61$_{\scriptstyle(0.66)}$ & 67.61$_{\scriptstyle(0.66)}$ & 67.61$_{\scriptstyle(0.66)}$ & 67.61$_{\scriptstyle(0.66)}$ & 67.61$_{\scriptstyle(0.66)}$ \\
\bottomrule
\end{tabular}
\end{table}

\begin{table}[h]
\centering
\caption{\yg{MOSEI results with varying $\lambda$ (averaged over 10 seeds, standard errors in parentheses). All values are percentages.}}
\label{tab:mosei}
\begin{tabular}{lcccccccc}
\toprule
$\lambda$ & 0.01 & 0.1 & 1.0 & 10.0 & 100.0 & 1000.0 & max over $\lambda$ \\
\midrule
IndiSeek & 74.70$_{\scriptstyle(0.06)}$ & 74.71$_{\scriptstyle(0.05)}$ & 75.47$_{\scriptstyle(0.13)}$ & 75.29$_{\scriptstyle(0.09)}$ & 75.19$_{\scriptstyle(0.11)}$ & 75.03$_{\scriptstyle(0.11)}$ & 75.47$_{\scriptstyle(0.13)}$ \\
FactorizedCL & 74.74$_{\scriptstyle(0.04)}$ & 74.74$_{\scriptstyle(0.05)}$ & 74.65$_{\scriptstyle(0.04)}$ & 74.69$_{\scriptstyle(0.07)}$ & 74.65$_{\scriptstyle(0.04)}$ & 74.60$_{\scriptstyle(0.06)}$ & 74.74$_{\scriptstyle(0.04)}$ \\
InfoDisen & 74.65$_{\scriptstyle(0.06)}$ & 74.61$_{\scriptstyle(0.03)}$ & 74.58$_{\scriptstyle(0.04)}$ & 74.70$_{\scriptstyle(0.03)}$ & 74.73$_{\scriptstyle(0.08)}$ & 74.56$_{\scriptstyle(0.04)}$ & 74.73$_{\scriptstyle(0.08)}$ \\
CLIP & 74.70$_{\scriptstyle(0.05)}$ & 74.70$_{\scriptstyle(0.05)}$ & 74.70$_{\scriptstyle(0.05)}$ & 74.70$_{\scriptstyle(0.05)}$ & 74.70$_{\scriptstyle(0.05)}$ & 74.70$_{\scriptstyle(0.05)}$ & 74.70$_{\scriptstyle(0.05)}$ \\
\bottomrule
\end{tabular}
\end{table}

\begin{table}[h]
\centering
\caption{\yg{UR-FUNNY results with varying $\lambda$ (averaged over 10 seeds, standard errors in parentheses). All values are percentages.}}
\label{tab:humor}
\begin{tabular}{lcccccccc}
\toprule
$\lambda$ & 0.01 & 0.1 & 1.0 & 10.0 & 100.0 & 1000.0 & max over $\lambda$ \\
\midrule
IndiSeek & 58.85$_{\scriptstyle(1.08)}$ & 62.40$_{\scriptstyle(0.82)}$ & 63.08$_{\scriptstyle(0.44)}$ & 63.79$_{\scriptstyle(0.39)}$ & 63.12$_{\scriptstyle(0.55)}$ & 63.71$_{\scriptstyle(0.62)}$ & 63.79$_{\scriptstyle(0.39)}$ \\
FactorizedCL & 58.36$_{\scriptstyle(0.25)}$ & 57.81$_{\scriptstyle(0.41)}$ & 58.22$_{\scriptstyle(0.38)}$ & 57.43$_{\scriptstyle(0.64)}$ & 57.13$_{\scriptstyle(0.24)}$ & 57.02$_{\scriptstyle(0.32)}$ & 58.36$_{\scriptstyle(0.25)}$ \\
InfoDisen & 56.09$_{\scriptstyle(0.33)}$ & 56.14$_{\scriptstyle(0.98)}$ & 56.28$_{\scriptstyle(0.76)}$ & 57.52$_{\scriptstyle(0.57)}$ & 58.08$_{\scriptstyle(0.50)}$ & 57.73$_{\scriptstyle(0.70)}$ & 58.08$_{\scriptstyle(0.50)}$ \\
CLIP & 58.32$_{\scriptstyle(0.68)}$ & 58.32$_{\scriptstyle(0.68)}$ & 58.32$_{\scriptstyle(0.68)}$ & 58.32$_{\scriptstyle(0.68)}$ & 58.32$_{\scriptstyle(0.68)}$ & 58.32$_{\scriptstyle(0.68)}$ & 58.32$_{\scriptstyle(0.68)}$ \\
\bottomrule
\end{tabular}
\end{table}

\begin{table}[h]
\centering
\caption{\yg{MUStARD results with varying $\lambda$ (averaged over 10 seeds, standard errors in parentheses). All values are percentages.}}
\label{tab:sarcasm}
\begin{tabular}{lcccccccc}
\toprule
$\lambda$ & 0.01 & 0.1 & 1.0 & 10.0 & 100.0 & 1000.0 & max over $\lambda$ \\
\midrule
IndiSeek & 55.51$_{\scriptstyle(1.28)}$ & 57.46$_{\scriptstyle(1.04)}$ & 55.72$_{\scriptstyle(1.95)}$ & 55.07$_{\scriptstyle(1.06)}$ & 53.48$_{\scriptstyle(1.42)}$ & 52.25$_{\scriptstyle(1.17)}$ & 57.46$_{\scriptstyle(1.04)}$ \\
FactorizedCL & 53.91$_{\scriptstyle(0.57)}$ & 54.71$_{\scriptstyle(1.21)}$ & 54.86$_{\scriptstyle(1.10)}$ & 54.64$_{\scriptstyle(0.70)}$ & 56.45$_{\scriptstyle(1.16)}$ & 55.07$_{\scriptstyle(0.91)}$ & 56.45$_{\scriptstyle(1.16)}$ \\
InfoDisen & 54.13$_{\scriptstyle(0.83)}$ & 53.41$_{\scriptstyle(1.00)}$ & 54.42$_{\scriptstyle(0.97)}$ & 54.57$_{\scriptstyle(0.94)}$ & 56.16$_{\scriptstyle(0.92)}$ & 55.58$_{\scriptstyle(1.27)}$ & 56.16$_{\scriptstyle(0.92)}$ \\
CLIP & 55.36$_{\scriptstyle(1.12)}$ & 55.36$_{\scriptstyle(1.12)}$ & 55.36$_{\scriptstyle(1.12)}$ & 55.36$_{\scriptstyle(1.12)}$ & 55.36$_{\scriptstyle(1.12)}$ & 55.36$_{\scriptstyle(1.12)}$ & 55.36$_{\scriptstyle(1.12)}$ \\
\bottomrule
\end{tabular}
\end{table}

\begin{table}[h]
\centering
\caption{\yg{MIMIC results with varying $\lambda$ (averaged over 10 seeds, standard errors in parentheses). All values are percentages.}}
\label{tab:mimic}
\begin{tabular}{lcccccccc}
\toprule
$\lambda$ & 0.01 & 0.1 & 1.0 & 10.0 & 100.0 & 1000.0 & max over $\lambda$ \\
\midrule
IndiSeek & 65.52$_{\scriptstyle(0.22)}$ & 65.79$_{\scriptstyle(0.17)}$ & 65.82$_{\scriptstyle(0.24)}$ & 65.89$_{\scriptstyle(0.24)}$ & 65.96$_{\scriptstyle(0.28)}$ & 65.99$_{\scriptstyle(0.31)}$ & 65.99$_{\scriptstyle(0.31)}$ \\
FactorizedCL & 65.34$_{\scriptstyle(0.29)}$ & 65.11$_{\scriptstyle(0.31)}$ & 65.69$_{\scriptstyle(0.11)}$ & 65.33$_{\scriptstyle(0.30)}$ & 64.92$_{\scriptstyle(0.35)}$ & 65.24$_{\scriptstyle(0.39)}$ & 65.69$_{\scriptstyle(0.11)}$ \\
InfoDisen & 64.89$_{\scriptstyle(0.30)}$ & 64.80$_{\scriptstyle(0.31)}$ & 65.10$_{\scriptstyle(0.25)}$ & 64.96$_{\scriptstyle(0.38)}$ & 65.47$_{\scriptstyle(0.23)}$ & 65.32$_{\scriptstyle(0.24)}$ & 65.47$_{\scriptstyle(0.23)}$ \\
CLIP & 64.56$_{\scriptstyle(0.29)}$ & 64.56$_{\scriptstyle(0.29)}$ & 64.56$_{\scriptstyle(0.29)}$ & 64.56$_{\scriptstyle(0.29)}$ & 64.56$_{\scriptstyle(0.29)}$ & 64.56$_{\scriptstyle(0.29)}$ & 64.56$_{\scriptstyle(0.29)}$ \\
\bottomrule
\end{tabular}
\end{table}

\begin{table}[ht]
\centering
\caption{\yg{Comparison of training time on multimodal datasets (averaged over 10 seeds, standard errors in parentheses, max average over $\lambda$). All values are in seconds ($\times 10^2$).}}
\label{tab:time_summary}
\begin{tabular}{lccccc}
\toprule
\textbf{Method} & \textbf{MOSI} & \textbf{MOSEI} & \textbf{UR-FUNNY} & \textbf{MUStARD} & \textbf{MIMIC} \\
\midrule
IndiSeek         & 7.11$_{\scriptstyle(0.09)}$ & 81.43$_{\scriptstyle(0.83)}$ & 40.75$_{\scriptstyle(0.29)}$ & 2.63$_{\scriptstyle(0.04)}$ & 132.74$_{\scriptstyle(1.76)}$ \\
FactorizedCL     & 7.35$_{\scriptstyle(0.09)}$ & 83.27$_{\scriptstyle(0.87)}$ & 41.97$_{\scriptstyle(0.31)}$ & 2.72$_{\scriptstyle(0.04)}$ & 135.93$_{\scriptstyle(1.84)}$ \\
InfoDisen        & 7.18$_{\scriptstyle(0.09)}$ & 81.32$_{\scriptstyle(0.85)}$ & 40.70$_{\scriptstyle(0.30)}$ & 2.65$_{\scriptstyle(0.04)}$  & 132.60$_{\scriptstyle(1.98)}$ \\
\bottomrule
\end{tabular}
\end{table}

\subsubsection{Experimental results with varying $\lambda$}\label{sec:app-vary-lam}
In Section~\ref{sec:multibench}, we present the accuracies averaged over $10$ seeds with the optimal $\lambda$ for each method. In the following tables, we present more detailed results on the accuracy for each method with different choices of $\lambda$: Table~\ref{tab:mosei} for MOSEI, Table~\ref{tab:mosi} for MOSI, Table~\ref{tab:sarcasm} for MUStARD, Table~\ref{tab:humor} for UR-FUNNY, and Table~\ref{tab:mimic} for MIMIC.

\subsubsection{Training time comparison}

We compare the computational efficiency of different methods by measuring the training time required to compute modality-specific features. Specifically, we record the time for training the disentangled representations for both modalities (text and vision) after the initial CLIP feature extraction. All methods are trained with consistent configurations: we use a batch size of 128, a maximum of 2000 epochs with early stopping, Adam optimizer with learning rate $10^{-4}$ and weight decay $10^{-4}$, and Transformer-based encoders to map raw sequences to a $50$-dimensional latent space. The training is performed on a single GPU, and we measure the wall-clock time for the complete training of both modality-specific encoders.

Table~\ref{tab:time_summary} presents the training time comparison across datasets, averaged over 10 random seeds. The results show that all three methods have comparable computational costs, with differences primarily stemming from dataset sizes and sequence lengths. IndiSeek demonstrates competitive efficiency while achieving superior accuracy (as shown in Table~\ref{tab:accuracy_summary}), making it a practical choice for multimodal learning tasks.

\subsubsection{Ablation studies and comparison with CoMM}

To understand the contribution of individual components in \ours, we conduct an ablation study on the MOSI dataset. Specifically, we compare \ours method against IndiSeek0, a variant that removes the modality-switching mechanism and uses a fixed pairing between raw inputs and CLIP embeddings. We also include a comparison with CoMM~\citep{dufumier2024align}.
Table~\ref{tab:mosi_ablation} presents the results of this analysis. 
CoMM achieves competitive performance at 69.56\%, showing that collaborative learning approaches can be effective for multimodal representation learning.

\begin{table}[h]
\centering
\caption{\yg{MOSI ablation study results with varying $\lambda$ (averaged over 10 seeds, standard errors in parentheses). All values are percentages.}}
\label{tab:mosi_ablation}
\begin{tabular}{lcccccccc}
\toprule
$\lambda$ & 0.01 & 0.1 & 1.0 & 10.0 & 100.0 & 1000.0 & max over $\lambda$ \\
\midrule
IndiSeek & 67.11$_{\scriptstyle(0.85)}$ & 69.61$_{\scriptstyle(0.55)}$ & 70.09$_{\scriptstyle(0.80)}$ & 66.63$_{\scriptstyle(0.56)}$ & 66.43$_{\scriptstyle(0.50)}$ & 66.06$_{\scriptstyle(0.72)}$ & 70.09$_{\scriptstyle(0.80)}$ \\
IndiSeek0 & 66.24$_{\scriptstyle(0.43)}$ & 69.21$_{\scriptstyle(0.48)}$ & 69.59$_{\scriptstyle(0.27)}$ & 66.57$_{\scriptstyle(0.49)}$ & 65.73$_{\scriptstyle(0.35)}$ & 65.07$_{\scriptstyle(0.21)}$ & 69.59$_{\scriptstyle(0.27)}$ \\
CoMM & 68.34$_{\scriptstyle(0.79)}$ & 68.69$_{\scriptstyle(0.65)}$ & 69.56$_{\scriptstyle(0.47)}$ & 67.14$_{\scriptstyle(0.37)}$ & 66.79$_{\scriptstyle(0.59)}$ & 66.60$_{\scriptstyle(0.44)}$ & 69.56$_{\scriptstyle(0.47)}$ \\
FactorizedCL & 66.76$_{\scriptstyle(0.41)}$ & 65.07$_{\scriptstyle(0.65)}$ & 66.12$_{\scriptstyle(0.47)}$ & 64.26$_{\scriptstyle(0.76)}$ & 65.07$_{\scriptstyle(0.41)}$ & 65.12$_{\scriptstyle(0.63)}$ & 67.11$_{\scriptstyle(0.24)}$ \\
InfoDisen & 65.57$_{\scriptstyle(0.59)}$ & 65.31$_{\scriptstyle(0.89)}$ & 65.83$_{\scriptstyle(0.71)}$ & 65.22$_{\scriptstyle(0.76)}$ & 64.84$_{\scriptstyle(0.89)}$ & 67.52$_{\scriptstyle(0.44)}$ & 67.52$_{\scriptstyle(0.44)}$ \\
CLIP & 67.14$_{\scriptstyle(0.87)}$ & 67.14$_{\scriptstyle(0.87)}$ & 67.14$_{\scriptstyle(0.87)}$ & 67.14$_{\scriptstyle(0.87)}$ & 67.14$_{\scriptstyle(0.87)}$ & 67.14$_{\scriptstyle(0.87)}$ & 67.14$_{\scriptstyle(0.87)}$ \\
\bottomrule
\end{tabular}
\end{table}

\subsubsection{Multi-task performance on MIMIC dataset}
In this section, we evaluate \ours~on all $20$ tasks of predicting ICD-9 groups, and results are summarized in the following table.

\begin{table}[ht]
\centering
\caption{\yg{Multi-task performance on MIMIC dataset (averaged over 10 seeds, standard errors in parentheses, max average over $\lambda$). All values are percentages.}}
\label{tab:multi-all}
\begin{tabular}{ccccc}
\toprule
Task ID & IndiSeek & FactorizedCL & InfoDisen & CLIP \\
\midrule
0   & 77.73$_{\scriptstyle(0.06)}$ & 77.76$_{\scriptstyle(0.10)}$ & 77.34$_{\scriptstyle(0.04)}$ & 77.07$_{\scriptstyle(0.18)}$ \\
1   & 91.31$_{\scriptstyle(0.03)}$ & 91.30$_{\scriptstyle(0.02)}$ & 91.30$_{\scriptstyle(0.02)}$ & 91.12$_{\scriptstyle(0.07)}$ \\
2   & 70.76$_{\scriptstyle(0.19)}$ & 70.67$_{\scriptstyle(0.11)}$ & 70.62$_{\scriptstyle(0.24)}$ & 70.22$_{\scriptstyle(0.09)}$ \\
3   & 67.04$_{\scriptstyle(0.10)}$ & 67.01$_{\scriptstyle(0.08)}$ & 66.78$_{\scriptstyle(0.12)}$ & 66.89$_{\scriptstyle(0.11)}$ \\
4   & 71.04$_{\scriptstyle(0.22)}$ & 70.86$_{\scriptstyle(0.32)}$ & 70.87$_{\scriptstyle(0.20)}$ & 70.81$_{\scriptstyle(0.07)}$ \\
5   & 72.22$_{\scriptstyle(0.08)}$ & 72.03$_{\scriptstyle(0.05)}$ & 72.11$_{\scriptstyle(0.09)}$ & 71.97$_{\scriptstyle(0.09)}$ \\
6   & 86.00$_{\scriptstyle(0.06)}$ & 86.17$_{\scriptstyle(0.10)}$ & 85.94$_{\scriptstyle(0.11)}$ & 85.98$_{\scriptstyle(0.07)}$ \\
7   & 66.04$_{\scriptstyle(0.18)}$ & 65.81$_{\scriptstyle(0.35)}$ & 65.58$_{\scriptstyle(0.34)}$ & 64.50$_{\scriptstyle(0.42)}$ \\
8   & 65.66$_{\scriptstyle(0.07)}$ & 65.84$_{\scriptstyle(0.05)}$ & 65.50$_{\scriptstyle(0.19)}$ & 65.04$_{\scriptstyle(0.12)}$ \\
9   & 73.09$_{\scriptstyle(0.17)}$ & 73.16$_{\scriptstyle(0.09)}$ & 72.75$_{\scriptstyle(0.19)}$ & 72.44$_{\scriptstyle(0.19)}$ \\
10  & 99.60$_{\scriptstyle(0.00)}$ & 99.60$_{\scriptstyle(0.01)}$ & 99.60$_{\scriptstyle(0.00)}$ & 99.60$_{\scriptstyle(0.00)}$ \\
11  & 89.14$_{\scriptstyle(0.00)}$ & 89.14$_{\scriptstyle(0.00)}$ & 89.14$_{\scriptstyle(0.00)}$ & 89.14$_{\scriptstyle(0.00)}$ \\
12  & 80.93$_{\scriptstyle(0.00)}$ & 80.94$_{\scriptstyle(0.02)}$ & 80.93$_{\scriptstyle(0.01)}$ & 80.93$_{\scriptstyle(0.00)}$ \\
13  & 96.69$_{\scriptstyle(0.00)}$ & 96.69$_{\scriptstyle(0.02)}$ & 96.69$_{\scriptstyle(0.01)}$ & 96.69$_{\scriptstyle(0.00)}$ \\
14  & 69.43$_{\scriptstyle(0.05)}$ & 69.21$_{\scriptstyle(0.17)}$ & 69.13$_{\scriptstyle(0.13)}$ & 68.99$_{\scriptstyle(0.02)}$ \\
15  & 90.98$_{\scriptstyle(0.00)}$ & 90.98$_{\scriptstyle(0.00)}$ & 90.98$_{\scriptstyle(0.00)}$ & 90.98$_{\scriptstyle(0.00)}$ \\
16  & 96.84$_{\scriptstyle(0.00)}$ & 96.84$_{\scriptstyle(0.00)}$ & 96.84$_{\scriptstyle(0.00)}$ & 96.84$_{\scriptstyle(0.00)}$ \\
17  & 62.58$_{\scriptstyle(0.17)}$ & 61.83$_{\scriptstyle(0.20)}$ & 61.72$_{\scriptstyle(0.14)}$ & 61.50$_{\scriptstyle(0.14)}$ \\
18  & 60.99$_{\scriptstyle(0.19)}$ & 60.36$_{\scriptstyle(0.14)}$ & 60.40$_{\scriptstyle(0.34)}$ & 60.71$_{\scriptstyle(0.09)}$ \\
19  & 69.71$_{\scriptstyle(0.11)}$ & 69.30$_{\scriptstyle(0.12)}$ & 69.46$_{\scriptstyle(0.14)}$ & 69.25$_{\scriptstyle(0.16)}$ \\
\midrule
Average & 77.89 & 77.82 & 77.74 & 77.55 \\
\bottomrule
\end{tabular}
\end{table}

In Figure~\ref{tab:multi-all}, \ours~leads the performance in most tasks. Moreover, in many real-world scenarios, downstream tasks are unknown in advance, or learned representations are expected to handle hundreds of tasks simultaneously. In this case, the task-relevant optimality of augmentations could be barely met in \cite{liang2023factorized}, which highlights the flexibility and practicality of task-agnostic disentangled learning.

% \begin{figure}[ht]
%     \centering
%     % --- first row ---
%     \begin{subfigure}[b]{0.48\linewidth}
%         \centering
%         \includegraphics[width=\linewidth]{figs/plot_mosi_transformer_concat.png}
%         \caption{MOSI}
%         \label{fig:lambda_mosi}
%     \end{subfigure}%
%     \hfill
%     \begin{subfigure}[b]{0.48\linewidth}
%         \centering
%         \includegraphics[width=\linewidth]{figs/plot_mosei_transformer_concat.png}
%         \caption{MOSEI}
%         \label{fig:lambda_mosei}
%     \end{subfigure}

%     % --- second row ---
%     \begin{subfigure}[b]{0.48\linewidth}
%         \centering
%         \includegraphics[width=\linewidth]{figs/plot_humor_transformer_concat.png}
%         \caption{UR-FUNNY}
%         \label{fig:lambda_humor}
%     \end{subfigure}%
%     \hfill
%     \begin{subfigure}[b]{0.48\linewidth}
%         \centering
%         \includegraphics[width=\linewidth]{figs/plot_sarcasm_transformer_concat.png}
%         \caption{MUStARD}
%         \label{fig:lambda_sarcasm}
%     \end{subfigure}

%     \caption{Comparison of results across four video datasets with varying $\lambda$.}
%     \label{fig:lambda_all}
% \end{figure}

% \begin{figure}[!h]
%     \centering
%     \includegraphics[width=0.5\linewidth]{figs/plot_mimic_transformer_concat.png}
%     \caption{Comparison of results on MIMIC dataset.}
%     \label{fig:mimic}
% \end{figure}

%% file: main.bib
@article{shi2025towards,
	author = {Shi, Long and Ye, Yunshan and Wang, Wenjie and Lei, Tao and Zhao, Yu and Kou, Gang and Chen, Badong},
	journal = {arXiv preprint arXiv:2509.02084},
	title = {Towards Comprehensive Information-theoretic Multi-view Learning},
	year = {2025}}

@inproceedings{wang2024decoupling,
	author = {Wang, Yi and Albrecht, Conrad M and Braham, Nassim Ait Ali and Liu, Chenying and Xiong, Zhitong and Zhu, Xiao Xiang},
	booktitle = {European Conference on Computer Vision},
	organization = {Springer},
	pages = {286--303},
	title = {Decoupling common and unique representations for multimodal self-supervised learning},
	year = {2024}}

@article{kotliar2025reproducible,
	author = {Kotliar, Dylan and Curtis, Michelle and Agnew, Ryan and Weinand, Kathryn and Nathan, Aparna and Baglaenko, Yuriy and Slowikowski, Kamil and Zhao, Yu and Sabeti, Pardis C and Rao, Deepak A and others},
	journal = {Nature Methods},
	pages = {1--17},
	publisher = {Nature Publishing Group US New York},
	title = {Reproducible single-cell annotation of programs underlying T cell subsets, activation states and functions},
	year = {2025}}

@article{szabo2019single,
	author = {Szabo, Peter A and Levitin, Hanna Mendes and Miron, Michelle and Snyder, Mark E and Senda, Takashi and Yuan, Jinzhou and Cheng, Yim Ling and Bush, Erin C and Dogra, Pranay and Thapa, Puspa and others},
	journal = {Nature communications},
	number = {1},
	pages = {4706},
	publisher = {Nature Publishing Group UK London},
	title = {Single-cell transcriptomics of human T cells reveals tissue and activation signatures in health and disease},
	volume = {10},
	year = {2019}}

@article{zheng2021pan,
	author = {Zheng, Liangtao and Qin, Shishang and Si, Wen and Wang, Anqiang and Xing, Baocai and Gao, Ranran and Ren, Xianwen and Wang, Li and Wu, Xiaojiang and Zhang, Ji and others},
	journal = {Science},
	number = {6574},
	pages = {abe6474},
	publisher = {American Association for the Advancement of Science},
	title = {Pan-cancer single-cell landscape of tumor-infiltrating T cells},
	volume = {374},
	year = {2021}}

@article{fischer2020conditional,
	author = {Fischer, Ian},
	journal = {Entropy},
	number = {9},
	pages = {999},
	publisher = {MDPI},
	title = {The conditional entropy bottleneck},
	volume = {22},
	year = {2020}}

@article{oko2025statistical,
	author = {Oko, Kazusato and Lin, Licong and Cai, Yuhang and Mei, Song},
	journal = {arXiv preprint arXiv:2501.04641},
	title = {A Statistical Theory of Contrastive Pre-training and Multimodal Generative AI},
	year = {2025}}

@article{oord2018representation,
	author = {Oord, Aaron van den and Li, Yazhe and Vinyals, Oriol},
	journal = {arXiv preprint arXiv:1807.03748},
	title = {Representation learning with contrastive predictive coding},
	year = {2018}}

@article{teichmann2020method,
	author = {Teichmann, Sarah and Efremova, Mirjana},
	journal = {Nat. Methods},
	number = {1},
	pages = {2020},
	title = {Method of the Year 2019: single-cell multimodal omics},
	volume = {17},
	year = {2020}}

@article{hao2021integrated,
	author = {Hao, Yuhan and Hao, Stephanie and Andersen-Nissen, Erica and Mauck, William M and Zheng, Shiwei and Butler, Andrew and Lee, Maddie J and Wilk, Aaron J and Darby, Charlotte and Zager, Michael and others},
	journal = {Cell},
	number = {13},
	pages = {3573--3587},
	publisher = {Elsevier},
	title = {Integrated analysis of multimodal single-cell data},
	volume = {184},
	year = {2021}}

@article{baltruvsaitis2018multimodal,
	author = {Baltru{\v{s}}aitis, Tadas and Ahuja, Chaitanya and Morency, Louis-Philippe},
	journal = {IEEE transactions on pattern analysis and machine intelligence},
	number = {2},
	pages = {423--443},
	publisher = {IEEE},
	title = {Multimodal machine learning: A survey and taxonomy},
	volume = {41},
	year = {2018}}

@inproceedings{sridharan2008information,
	author = {Sridharan, Karthik and Kakade, Sham M},
	booktitle = {COLT},
	number = {114},
	pages = {403--414},
	title = {An information theoretic framework for multi-view learning},
	year = {2008}}

@article{hubert1985comparing,
  title={Comparing partitions},
  author={Hubert, Lawrence and Arabie, Phipps},
  journal={Journal of Classification},
  volume={2},
  number={1},
  pages={193--218},
  year={1985},
  publisher={Springer}
}

@inproceedings{radford2021learning,
	author = {Radford, Alec and Kim, Jong Wook and Hallacy, Chris and Ramesh, Aditya and Goh, Gabriel and Agarwal, Sandhini and Sastry, Girish and Askell, Amanda and Mishkin, Pamela and Clark, Jack and others},
	booktitle = {International conference on machine learning},
	organization = {PMLR},
	pages = {8748--8763},
	title = {Learning transferable visual models from natural language supervision},
	year = {2021}}

@article{stoeckius2017simultaneous,
	author = {Stoeckius, Marlon and Hafemeister, Christoph and Stephenson, William and Houck-Loomis, Brian and Chattopadhyay, Pratip K and Swerdlow, Harold and Satija, Rahul and Smibert, Peter},
	journal = {Nature methods},
	number = {9},
	pages = {865--868},
	publisher = {Nature Publishing Group},
	title = {Simultaneous epitope and transcriptome measurement in single cells},
	volume = {14},
	year = {2017}}

@article{stuart2019comprehensive,
	author = {Stuart, Tim and Butler, Andrew and Hoffman, Paul and Hafemeister, Christoph and Papalexi, Efthymia and Mauck, William M and Hao, Yuhan and Stoeckius, Marlon and Smibert, Peter and Satija, Rahul},
	journal = {cell},
	number = {7},
	pages = {1888--1902},
	publisher = {Elsevier},
	title = {Comprehensive integration of single-cell data},
	volume = {177},
	year = {2019}}

@article{lin2025statistical,
	author = {Lin, Licong and Mei, Song},
	journal = {arXiv preprint arXiv:2503.17538},
	title = {A Statistical Theory of Contrastive Learning via Approximate Sufficient Statistics},
	year = {2025}}

@book{cover1999elements,
	author = {Cover, Thomas M},
	publisher = {John Wiley \& Sons},
	title = {Elements of information theory},
	year = {1999}}

@article{wang2024information,
	author = {Wang, Chenyu and Gupta, Sharut and Zhang, Xinyi and Tonekaboni, Sana and Jegelka, Stefanie and Jaakkola, Tommi and Uhler, Caroline},
	journal = {arXiv preprint arXiv:2410.23996},
	title = {An Information Criterion for Controlled Disentanglement of Multimodal Data},
	year = {2024}}

@article{dufumier2024align,
	author = {Dufumier, Benoit and Castillo-Navarro, Javiera and Tuia, Devis and Thiran, Jean-Philippe},
	journal = {arXiv preprint arXiv:2409.07402},
	title = {What to align in multimodal contrastive learning?},
	year = {2024}}

@article{liang2023factorized,
	author = {Liang, Paul Pu and Deng, Zihao and Ma, Martin Q and Zou, James Y and Morency, Louis-Philippe and Salakhutdinov, Ruslan},
	journal = {Advances in Neural Information Processing Systems},
	pages = {32971--32998},
	title = {Factorized contrastive learning: Going beyond multi-view redundancy},
	volume = {36},
	year = {2023}}

@article{luecken2022benchmarking,
  title={Benchmarking atlas-level data integration in single-cell genomics},
  author={Luecken, Malte D and B{\"u}ttner, Maren and Chaichoompu, Kridsadakorn and Danese, Anna and Interlandi, Marta and M{\"u}ller, Michaela F and Strobl, Daniel C and Zappia, Luke and Dugas, Martin and Colom{\'e}-Tatch{\'e}, Maria and others},
  journal={Nature methods},
  volume={19},
  number={1},
  pages={41--50},
  year={2022},
  publisher={Nature Publishing Group US New York}
}

@article{pedregosa2011scikit,
  title={Scikit-learn: Machine learning in Python},
  author={Pedregosa, Fabian and Varoquaux, Ga{\"e}l and Gramfort, Alexandre and Michel, Vincent and Thirion, Bertrand and Grisel, Olivier and Blondel, Mathieu and Prettenhofer, Peter and Weiss, Ron and Dubourg, Vincent and others},
  journal={the Journal of machine Learning research},
  volume={12},
  pages={2825--2830},
  year={2011},
  publisher={JMLR. org}
}

@article{liang2023factorizedgithub,
	author = {Liang, Paul Pu and Deng, Zihao and Ma, Martin Q and Zou, James Y and Morency, Louis-Philippe and Salakhutdinov, Ruslan},
	journal = {github},
	title = {FactorCL: Factorized contrastive learning: Going beyond multi-view redundancy \url{https://github.com/pliang279/FactorCL/blob/main/multibench_model.py}},
	year = {2023}}

@inproceedings{cheng2020club,
	author = {Cheng, Pengyu and Hao, Weituo and Dai, Shuyang and Liu, Jiachang and Gan, Zhe and Carin, Lawrence},
	booktitle = {International conference on machine learning},
	organization = {PMLR},
	pages = {1779--1788},
	title = {Club: A contrastive log-ratio upper bound of mutual information},
	year = {2020}}

@article{gui2025multi,
	author = {Gui, Yu and Ma, Cong and Ma, Zongming},
	journal = {Advances in Neural Information Processing Systems},
	title = {Multi-modal contrastive learning adapts to intrinsic dimensions of shared latent variables},
	year = {2025}}

@article{robnik2008explaining,
	author = {Robnik-{\v{S}}ikonja, Marko and Kononenko, Igor},
	journal = {IEEE Transactions on Knowledge and Data Engineering},
	number = {5},
	pages = {589--600},
	publisher = {IEEE},
	title = {Explaining classifications for individual instances},
	volume = {20},
	year = {2008}}

@inproceedings{zeiler2014visualizing,
	author = {Zeiler, Matthew D and Fergus, Rob},
	booktitle = {European conference on computer vision},
	organization = {Springer},
	pages = {818--833},
	title = {Visualizing and understanding convolutional networks},
	year = {2014}}

@article{li2016understanding,
	author = {Li, Jiwei and Monroe, Will and Jurafsky, Dan},
	journal = {arXiv preprint arXiv:1612.08220},
	title = {Understanding neural networks through representation erasure},
	year = {2016}}

@article{spearman1961proof,
	author = {Spearman, Charles},
	publisher = {Appleton-Century-Crofts},
	title = {The proof and measurement of association between two things.},
	year = {1961}}

@article{liang2021multibench,
	author = {Liang, Paul Pu and Lyu, Yiwei and Fan, Xiang and Wu, Zetian and Cheng, Yun and Wu, Jason and Chen, Leslie and Wu, Peter and Lee, Michelle A and Zhu, Yuke and others},
	journal = {Advances in neural information processing systems},
	number = {DB1},
	pages = {1},
	title = {Multibench: Multiscale benchmarks for multimodal representation learning},
	volume = {2021},
	year = {2021}}

@article{caron2025multimodal,
	author = {Caron, Daniel P and Specht, William L and Chen, David and Wells, Steven B and Szabo, Peter A and Jensen, Isaac J and Farber, Donna L and Sims, Peter A},
	journal = {Cell Reports Methods},
	number = {1},
	publisher = {Elsevier},
	title = {Multimodal hierarchical classification of CITE-seq data delineates immune cell states across lineages and tissues},
	volume = {5},
	year = {2025}}

@inproceedings{tosh2021contrastive,
	author = {Tosh, Christopher and Krishnamurthy, Akshay and Hsu, Daniel},
	booktitle = {Algorithmic Learning Theory},
	organization = {PMLR},
	pages = {1179--1206},
	title = {Contrastive learning, multi-view redundancy, and linear models},
	year = {2021}}

@article{liu2023focal,
	author = {Liu, Shengzhong and Kimura, Tomoyoshi and Liu, Dongxin and Wang, Ruijie and Li, Jinyang and Diggavi, Suhas and Srivastava, Mani and Abdelzaher, Tarek},
	journal = {Advances in Neural Information Processing Systems},
	pages = {47309--47338},
	title = {Focal: Contrastive learning for multimodal time-series sensing signals in factorized orthogonal latent space},
	volume = {36},
	year = {2023}}

@inproceedings{jia2021scaling,
	author = {Jia, Chao and Yang, Yinfei and Xia, Ye and Chen, Yi-Ting and Parekh, Zarana and Pham, Hieu and Le, Quoc and Sung, Yun-Hsuan and Li, Zhen and Duerig, Tom},
	booktitle = {International conference on machine learning},
	organization = {PMLR},
	pages = {4904--4916},
	title = {Scaling up visual and vision-language representation learning with noisy text supervision},
	year = {2021}}

@article{akbari2021vatt,
	author = {Akbari, Hassan and Yuan, Liangzhe and Qian, Rui and Chuang, Wei-Hong and Chang, Shih-Fu and Cui, Yin and Gong, Boqing},
	journal = {Advances in neural information processing systems},
	pages = {24206--24221},
	title = {Vatt: Transformers for multimodal self-supervised learning from raw video, audio and text},
	volume = {34},
	year = {2021}}

@article{liang2024foundations,
	author = {Liang, Paul Pu and Zadeh, Amir and Morency, Louis-Philippe},
	journal = {ACM Computing Surveys},
	number = {10},
	pages = {1--42},
	publisher = {ACM New York, NY},
	title = {Foundations \& trends in multimodal machine learning: Principles, challenges, and open questions},
	volume = {56},
	year = {2024}}

@article{johnson2016mimic,
	author = {Johnson, Alistair EW and Pollard, Tom J and Shen, Lu and Lehman, Li-wei H and Feng, Mengling and Ghassemi, Mohammad and Moody, Benjamin and Szolovits, Peter and Anthony Celi, Leo and Mark, Roger G},
	journal = {Scientific data},
	number = {1},
	pages = {1--9},
	publisher = {Nature Publishing Group},
	title = {MIMIC-III, a freely accessible critical care database},
	volume = {3},
	year = {2016}}

@inproceedings{zadeh2018multimodal,
	author = {Zadeh, AmirAli Bagher and Liang, Paul Pu and Poria, Soujanya and Cambria, Erik and Morency, Louis-Philippe},
	booktitle = {Proceedings of the 56th Annual Meeting of the Association for Computational Linguistics (Volume 1: Long Papers)},
	pages = {2236--2246},
	title = {Multimodal language analysis in the wild: Cmu-mosei dataset and interpretable dynamic fusion graph},
	year = {2018}}

@article{hasan2019ur,
	author = {Hasan, Md Kamrul and Rahman, Wasifur and Zadeh, Amir and Zhong, Jianyuan and Tanveer, Md Iftekhar and Morency, Louis-Philippe and others},
	journal = {arXiv preprint arXiv:1904.06618},
	title = {UR-FUNNY: A multimodal language dataset for understanding humor},
	year = {2019}}

@article{castro2019towards,
	author = {Castro, Santiago and Hazarika, Devamanyu and P{\'e}rez-Rosas, Ver{\'o}nica and Zimmermann, Roger and Mihalcea, Rada and Poria, Soujanya},
	journal = {arXiv preprint arXiv:1906.01815},
	title = {Towards multimodal sarcasm detection (an \_obviously\_ perfect paper)},
	year = {2019}}

@article{zadeh2016mosi,
	author = {Zadeh, Amir and Zellers, Rowan and Pincus, Eli and Morency, Louis-Philippe},
	journal = {arXiv preprint arXiv:1606.06259},
	title = {Mosi: multimodal corpus of sentiment intensity and subjectivity analysis in online opinion videos},
	year = {2016}}
